\documentclass{Araya}

\title{Prior preferences in active inference agents: soft, hard, and goal shaping}

\author%
{\href{https://orcid.org/0000-0001-8790-2565}{\includegraphics[scale=0.06]{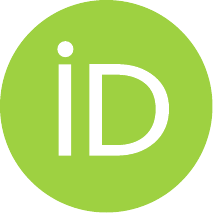}\hspace{1mm}Filippo Torresan}~$^{1, 2}$, \href{https://orcid.org/0000-0002-0186-2687}{\includegraphics[scale=0.06]{orcid.pdf}\hspace{1mm}Ryota Kanai}~$^{1}$, \href{https://orcid.org/0000-0002-6086-4711}{\includegraphics[scale=0.06]{orcid.pdf}\hspace{1mm}Manuel Baltieri}~$^{1,2,}$\footnote[2]{Correspondence e-mail: \texttt{manuel\_baltieri@araya.org}}\\
\vspace{1em} 
\normalfont{\small $^{1}$ Araya Inc., Tokyo, Japan}\\
\normalfont{\small $^{2}$ School of Engineering and Informatics, University of Sussex, Brighton, UK}\\
}

\begin{document}

\maketitle
\thispagestyle{firstpagestyle} 

\begin{abstract}
  Active inference proposes expected free energy as an objective for planning and decision-making to adequately balance exploitative and explorative drives in learning agents.
  The exploitative drive, or what an agent wants to achieve, is formalised as the Kullback-Leibler divergence between a variational probability distribution, updated at each inference step, and a preference probability distribution that indicates what states or observations are more likely for the agent, hence determining the agent's goal in a certain environment.
  In the literature, the questions of how the preference distribution should be specified and of how a certain specification impacts inference and learning in an active inference agent have been given hardly any attention.
  In this work, we consider four possible ways of defining the preference distribution, either providing the agents with hard or soft goals and either involving or not goal shaping (i.e., intermediate goals).
  We compare the performances of four agents, each given one of the possible preference distributions, in a grid world navigation task.
  Our results show that goal shaping enables the best performance overall (i.e., it promotes exploitation) while sacrificing learning about the environment's transition dynamics (i.e., it hampers exploration).
\end{abstract}

\keywords{active inference, Bayesian inference, POMDP, variational free energy, expected free energy, prior preferences}

\section{Introduction}\label{sec:intro}

Active inference has become an influential computational framework used to account for several aspects of cognition and adaptive behaviour in cognitive science and computational neuroscience \parencite{Friston2005a,Friston2009c,Friston2017a,Parr2022b,Pezzulo2024a}.
The fundamental idea of active inference is that adaptive agents are continuously engaged in a process of predicting upcoming sensory observations and inferring the best course of action to minimize prediction error.
This kind of perception-action loop is described at different spatio-temporal levels as a form of variational Bayesian inference, on the hidden states of the environment, that relies on a (hierarchical) generative model to minimize (variational) free energy, a proxy for prediction error~\parencite{Lee2003a,Friston2008a,Friston2017a}.
In this Bayesian framework, the minimisation of free energy and expected free energy enables an agent to infer its current state (perception), to infer the best sequence of actions (policies) to reach preferred states or observation (planning/goal-directed decision making), and to progressively learn the transition dynamics and state-observation mappings in the environment~\parencite{Clark2013b,Clark2015a,Pezzulo2015a,Catal2020a,Kaplan2018a,Bruineberg2017a}.

In contrast to reinforcement learning~\parencite{Sutton2018a}, the active inference framework tries to dispense with the notion of reward, so it assumes that agents are endowed with goals in the form of prior preferences to be achieved by performing free energy minimisation.
In various works deploying active inference agents in more traditional reinforcement learning environments, it is common to use a preference distribution over observations implicitly defined by regarding the reward signal in the environment as the desired observation for the agent~\parencite{Tschantz2020d,Tschantz2020c,Fountas2020a,Sajid2021a}.
However, despite the technical and theoretical fundamentals of active inference have been reviewed extensively in the literature~\parencite{Buckley2017a,DaCosta2020b,Gottwald2020a,Heins2022a,Lanillos2021a,Mazzaglia2022a,Nehrer2025a}, the issue of how to specify the preference distribution and its impact on inference and learning in an active inference agent have been largely overlooked.

In this work, our goal is to offer a thorough analysis of how different specifications of the preference distribution over states affect perception, decision-making, and learning in an active inference agent that has to solve a navigation task in a simple grid world.
In particular, we consider preference distributions that vary along two dimensions, i.e., providing the agent (1) with soft goals vs.\ hard goals and (2) with goal shaping or not.
The former determines how strongly the agent wants to reach a certain goal, whereas the latter specifies whether to give an agent a series of intermediate goals to reach the ultimate one. 

In~\cref{sec:aif-discrete-state-spaces} we briefly review the fundamental aspects of the active inference framework, with a focus on expected free energy and on how preference distributions play a key role in it.
With one experiment, we analyze inference and learning in four active inference agents, characterised by one of the possible preference distributions (based on the considered dimensions of variations), in a simple grid-world environment (\cref{sec:results}).
We will conclude with a discussion of how agent's performance and learning is affected by each of the considered preference distributions as well as with a few more general considerations about the notions of reward and goal-directedness in active inference (\cref{sec:discussion}).

\section{Active Inference in discrete state spaces}\label{sec:aif-discrete-state-spaces}

In the discrete state-space formulation of active inference, an agent's adaptive behaviour is modelled as a process of variational Bayesian inference given a generative model of the environment.
At each time step, the agent relies on its updated Bayesian beliefs to execute an action from a policy, \(\pi \in \Pi\) (a sequence of actions from a set of allowed sequences), so as to access one or more preferred states or observations (see~\cref{tab:summary-notation} for a summary of the notation used hereafter).
By acting this way, the generative model will partially reflect over time (through learning) the emission and transition maps that jointly characterise the environment's generative process, i.e., how observations are generated from states and how actions affect the transition from one state to another, respectively.
In the next few sections, we provide a brief overview of the main components and steps that characterize this active inference framework.
For a more detailed overview of the framework, see e.g.~\parencite{DaCosta2020b,Torresan2025a}.

\subsection{The Generative process and the generative model}\label{ssec:genmodel-genprocess}

Both the generative process and model are specified as discrete-time partially observable Markov decision processes (POMDPs).
Formally, we can define the generative process as follows:

\begin{definition}[POMDP in active inference, the generative process]\label{def:pomdp}
A POMDP is a six-element tuple, \((\statespace, \obsspace, \actionspace, \Transition, \Emission, \ntime)\), where:

    \begin{itemize}
        \item \(\statespace\) is a finite set of states,
        \item \(\obsspace\) is a finite set of observations,
        \item \(\actionspace\) is a finite set of admissible actions,
        \item \(\statevar_{i}, \obsvar_{i}, \actionvar_{i}\), with \(i \in [1, \ntime]\), are time-indexed random variables defined over the respective spaces, where the time index \(\ntime\) represents a terminal time step,
        \item \(\Transition: \statespace \times \actionspace \rightarrow \Delta(\statespace)\) is a transition function that maps state-action pairs to a probability distribution in the set \(\Delta(\statespace)\) of probability distribution defined over \(\statespace\)
        \item \(\Emission: \statespace  \rightarrow \Delta(\obsspace)\) is an emission function that maps a state to a probability distribution in the set \(\Delta(\obsspace)\) of probability distribution defined over \(\obsspace\)~\footnote{We note that standard definitions of POMDPs \parencites[Ch.~16]{Russell2021a}[Ch.~34]{Murphy2023a}[Ch.~17]{Sutton2018a} include also a notion of \emph{reward} for an agent, here we don't however include them since active inference normally specifies targets for an agent by means of a prior probability distribution over goal states or observations (see~\cref{ssec:aif-preferences}). Formally, however, this can be easily accommodated in the above definition by stating that our observations \(\obsspace\) include both observations $\mathcal{Y}$ and rewards $\mathcal{R}$ of standard POMDP definitions: $\obsspace = \mathcal{Y} \times \mathcal{R}$. Active inference works involving high-dimensional state spaces have adopted this approach in practice (see, e.g.,~\parencite{Tschantz2020c,Tschantz2020d,Fountas2020a}).}.
    \end{itemize}
\end{definition}

The generative model is specified as a joint probability distribution that can be generated by a POMDP in the sense of \cref{def:pomdp}.
We then define the generative model as follows:

\begin{definition}[Generative model in active inference]\label{def:aif-genmodel}

The generative model \(\genmodel\) of an active inference agent is a joint probability distribution over a sequence of state and observation random variables, a policy random variable for sequences of actions, and parameters stored in matrix \(\obsmap\) (for the emission map) and tensor \(\transmap\) (for the transition map), that is, a joint that factors as:

\begin{equation}\label{eq:aif-genmodel-fact}
    \begin{split}
        P(\seqv{\obsvar}{1}{\ntime}, \seqv{\statevar}{1}{\ntime}, \policy, \obsmap, \transmap) = \prob{\policy} \prob{\obsmap} \prob{\transmap} \prob{\statevar_{1}} \prod_{t=2}^{\ntime} \tprobaif \prod_{t=1}^{\ntime} \eprobaif.
    \end{split}
\end{equation}

The matrix \(\obsmap \in \mathbb{R}^{n \times m}\) stores the categorical probability distribution \(P(\obsvar_{t}|\state^{j}; \obsparams_{j})\) as the \(j\)th column, specifying the probabilities of the observations produced by state \(\statevar_{t} = \state^{j}\), for all state values \(s^{1}, \dotsc, s^{m}\) (those probabilities are stored by the parameter vector \(\obsparams_{j}\) that \emph{coincides} with the \(j\)th column of \(\obsmap\)).
The tensor \(\transmap \in \mathbb{R}^{\card{\actionspace} \times m \times m}\) stores the action-dependent categorical distribution \(P(\statevar_{t}|\state^{j}_{t-1}, x; \stateparams_{j})\) as the \(j\)th column of \(\transmap^{x}\), where \(x\) indicates the action under consideration, specifying the probabilities of the next state values given the previous state \(\statevar_{t-1} = \state^{j}\), for all state values \(s^{1}, \dotsc, s^{m}\) and for each action (again, those probabilities are stored by the parameter vector \(\stateparams_{j}\) that \emph{coincides} with the \(j\)th column of \(\transmap^{x}\)).
Each column of \(\obsmap\) and \(\transmap^{x}\) can be seen as an output of an approximation (learned by the active inference agent) of the emission map \(\Emission\) and the transition map \(\Transition\), so the generative model has indeed the same structure of a POMDP as defined in \cref{def:pomdp}.
\end{definition}

The goal of an active inference agent is then, in an intuitive sense, to enforce a \emph{synchronization} between the generative model it parameterises and the generative process of the environment it interacts with, see~\cite{Baltieri2025b} for a more in depth discussion of this reading of active inference.
More in detail, we can imagine that when an agent starts interacting with an environment, before its goal is achieved, the generative model has yet to capture the emission and transition maps of the generative process, i.e., the ground-truth POMDP representing the environment.
However, the agent can acquire such knowledge through experience.
At each time step, the agent performs inference on the most likely states corresponding to an observation, which also means revising its probabilistic beliefs about past and future consequences of performing different sequences of actions, then plans and decides what action to perform next.
At regular intervals, information about an experienced trajectory in the environment (i.e., a collection of observations plus probabilistic beliefs about the most likely states) is used to update the generative model's emission and transition map.
Next, we will provide a few more technical details on this procedure.

\subsection{Variational Bayesian inference for POMDPs}\label{ssec:approx-bayes}

An active inference agent learns to perform actions that will lead to its desired observations and/or states in the environment.
Observations received from the environment are evidence or feedback that can indicate to the agent whether the generative model captures the environmental dynamics well enough to yield accurate predictions and goal-conducive actions.
Such observations are used to infer the (1) most likely hidden states generating an observation at each time step, the (2) most likely policy given some preferred states or observations, and the (3) most likely parameters of the generative model to make more accurate predictions in the environment.

Given the agent's generative model (see~\cref{def:aif-genmodel}), this process of inference can be implemented by Bayes' rule, which in this setting corresponds to the following:
\begin{equation}\label{eq:aif-bayes-pomdp}
    \condprob{\seqv{\statevar}{1}{\ntime}, \policy, \obsmap, \transmap}{\seqv{\obsvar}{1}{\ntime}} = \frac{\condprob{\seqv{\obsvar}{1}{\ntime}}{\seqv{\statevar}{1}{\ntime}, \policy, \obsmap, \transmap} \prob{\seqv{\statevar}{1}{\ntime}, \policy, \obsmap, \transmap}}{\prob{\seqv{\obsvar}{1}{\ntime}}},
\end{equation}
where the generative model \(\genmodel\) appears in the numerator, factorised as the product between the likelihood and the prior probability distributions.
The goal here is to find a posterior joint distribution (the left-hand side) over the state, policy, and transition and emission maps' parameters random variables.
However, finding an analytic solution to \cref{eq:aif-bayes-pomdp} is often intractable, so active inference proposes to implement an approximate Bayesian inference scheme based on the minimisation of variational free energy.
This quantity is defined in relation to the available generative model, so it can be written as follows:
\begin{equation}\label{eq:aif-fe-pomdp}
    \fe \bigl[ \varprob{\statevar_{1:\ntime}, \policy, \obsmap, \transmap} \bigr] \coloneqq \mathbb{E}_{Q} \Bigl[\log \varprob{\statevar_{1:\ntime}, \policy, \obsmap, \transmap} - \log \prob{ \obsvar_{1:\ntime}, \statevar_{1:\ntime}, \policy, \obsmap, \transmap} \Bigr],
\end{equation}
where \(\varprob{\statevar_{1:\ntime}, \policy, \obsmap, \transmap}\) is known as the \emph{variational posterior}, a probability distribution introduced to approximate the posterior distribution, \(\condprob{\seqv{\statevar}{1}{\ntime}, \policy, \obsmap, \transmap}{\seqv{\obsvar}{1}{\ntime}}\), in \cref{eq:aif-bayes-pomdp} (the outcome of Bayesian inference).
To minimize the free energy defined in \cref{eq:aif-fe-pomdp}, we make some assumptions about the variational posterior so that the optimisation procedure described above becomes more tractable.
In discrete-time active inference, it is thus common to adopt a \emph{mean-field} approximation \parencite{DaCosta2020b}, meaning that the variational posterior is factorised as follows:

\begin{equation}\label{eq:variational-approx}
    \varprob{\statevar_{1:\ntime}, \policy, \obsmap, \transmap} = \varprob{\obsmap} \varprob{\transmap} \varprob{\policy} \prod_{t=1}^{\ntime} \varprob{\statevar_{t}|\policy}.
\end{equation}

By substituting this expression in \cref{eq:aif-fe-pomdp} for the variational posterior, and by considering the factorization of the generative model, we can rewrite the free energy as follows (\emph{cf}.,~\cite{DaCosta2020b}):
\begin{equation}\label{eq:fe-unpacked}
    \begin{split}
        \fe \bigl[\varprob{\statevar_{1:\ntime}, \policy, \obsmap, \transmap} \bigr] = & \infdiv[\Big]{\varprob{\obsmap}}{\prob{\obsmap}} + \infdiv[\Big]{\varprob{\transmap}}{\prob{\transmap}} + \infdiv[\Big]{\varprob{\policy}}{\prob{\policy}} \\
        & + \mathbb{E}_{\varprob{\policy_{k}}} \Biggl[
        \sum_{t=1}^{\ntime} \mathbb{E}_{\varprob{\statevar_{t}|\policy_{k}}} \Bigl[\log \varprob{\statevar_{t}|\policy_{k}} \Bigr] - \sum_{t=1}^{\tau} \mathbb{E}_{\varprob{\statevar_{t}|\policy_{k}} \varprob{\obsmap}} \Bigl[ \log \condprob{\obs_{t}}{\statevar_{t}, \obsmap} \Bigr] \\
        & - \mathbb{E}_{\varprob{\statevar_{1}|\policy_{k}}} \Bigl[ \log \prob{\statevar_{1}} \Bigr] - \sum_{t=2}^{\ntime} \mathbb{E}_{\varprob{\statevar_{t}|\policy_{k}} \varprob{\statevar_{t-1}|\policy_{k}}} \Bigl[\log \condprob{\statevar_{t}}{\statevar_{t-1}, \policy_{k}} \Bigr] \Biggr],
    \end{split}
\end{equation}
where we have singled out the KL divergences between the posterior probability distributions from the variational approximation and the prior probability distributions from the generative model (first three terms), and grouped together all the terms involving one of the variational posteriors \(\varprob{\statevar_{1:\ntime} | \policy_{k}}\), \(k \in [1, \numpolicies]\) where $\numpolicies$ is an integer indicating the maximum number of policies, inside the expectation \(\mathbb{E}_{\varprob{\policy_{k}}}[\dots]\) (last term), which computes an average with respect to all policies.

When the expression in~\cref{eq:fe-unpacked} is optimised with respect to the policy-conditioned variational distributions, \(\varprob{\statevar_{t}|\policy_{k}}\), \(\forall k \in [1, \numpolicies]\), we can simply focus on the argument of \(\mathbb{E}_{\varprob{\policy_{k}}}[\dots]\) to compute the associated gradient (since that is the only term that contributes to the gradient and ignoring the expectation does not change the solution of \(\nabla_{\stateparams_{t}} \fe[\varprob{\statevar_{t}|\policy_{k}}] = 0\)).
That argument defines a policy-conditioned free energy:

\begin{equation}\label{eq:policy-cond-fe-annotated}
    \begin{split}
        \fe_{\policy_{k}} \bigl[ \varprob{\statevar_{1:T} | \policy_{k}} \bigr] \coloneq
        & \sum_{t=1}^{\ntime} \mathbb{E}_{\varprob{\statevar_{t}|\policy_{k}}} \Bigl[ \underbrace{\log \varprob{\statevar_{t}|\policy_{k}}}_{\textbf{state log-probabilities}} \Bigr] - \sum_{t=1}^{\tau} \mathbb{E}_{\varprob{\statevar_{t}|\policy_{k}} \varprob{\obsmap}} \Bigl[ \underbrace{\log \condprob{\obs_{t}}{\statevar_{t}, \obsmap}}_{\textbf{observation log-likelihoods}} \Bigr] - \\
        & - \mathbb{E}_{\varprob{\statevar_{1}|\policy_{k}}} \Bigl[ \underbrace{\log \prob{\statevar_{1}}}_{\textbf{state log-probabilities}} \Bigr] - \sum_{t=2}^{\ntime} \mathbb{E}_{\varprob{\statevar_{t}|\policy_{k}} \varprob{\statevar_{t-1}|\policy_{k}}} \Bigl[ \underbrace{\log \condprob{\statevar_{t}}{\statevar_{t-1}, \policy_{k}}}_{\textbf{transition log-likelihoods}}  \Bigr].
    \end{split}
\end{equation}

The update rules for \(\varprob{\statevar_{t}|\policy_{k}}\), \(\forall k \in [1, \numpolicies]\), derived by taking the corresponding gradient of the expression in \cref{eq:policy-cond-fe-annotated}, define an optimisation/inference scheme called \emph{variational message passing} which makes use of past, present and future information to update, in this case, variational probability distributions at different time points along a trajectory (see appendices in~\parencite{DaCosta2020b,Torresan2025a} for details on the update equations).
Following standard treatments in the literature of stochastic processes and (Bayesian) estimation, it is an example of smoothing, to be contrasted with inference (which uses present information only) and filtering (which relies on past and present information), and  prediction (which uses the past only)~\parencite{Jazwinski1970a, Sarkka2013a}.

\subsection{Expected free energy}\label{ssec:efe}

Similarly, the derivation of the updated probability distribution over policies, \(\varprob{\policy}\), involves taking the gradient of the free energy in~\cref{eq:fe-unpacked} but with respect to the parameters of \(\varprob{\policy}\) this time.
While we again refer to~\parencite{DaCosta2020b,Torresan2025a} for the full details of the derivation, we briefly overview below the update equation so to introduce the notion of expected free energy next.

By indicating with \(\policyparams_{Q}^{\intercal}\) and \(\policyparams_{P}^{\intercal}\) the row vectors of parameters of the variational distribution \(\varprob{\policy}\) and the prior distribution \(\prob{\policy}\), respectively, and with \(\boldsymbol{\fe}_{\policy}^{\intercal}\) the row vector of policy-conditioned free energies (one for each policy, i.e., for each value the policy random variable can take, see~\cref{eq:policy-cond-fe-annotated}), the updated parameters for \(\varprob{\policy}\) are computed as follows:
\begin{equation}\label{eq:fe-policy-update-norm}
    \policyparams_{Q}^{\intercal} = \softmax{\ln \policyparams_{P}^{\intercal} - \boldsymbol{\fe}_{\policy}^{\intercal}},
\end{equation}
where the softmax function, \(\softmax{\cdot}\), is used to obtain a vector \(\policyparams_{Q}^{\intercal}\) of normalised probabilities from the \emph{unnormalised} probabilities of the vector \(\ln \policyparams_{Q}^{\intercal}\), after setting the gradient \(\nabla_{\policyparams} \fe[\varprob{\policy}]\) to zero and rearranging (see again appendices in~\parencite{DaCosta2020b,Torresan2025a}).

One of the key moves of active inference is to specify the prior parameters \(\policyparams_{P}^{\intercal} \) in terms of the vector \(\boldsymbol{\totefe}^{\intercal}\) of total expected free energies \(\totefe\) (defined below), one for each policy under consideration, i.e., \(\policyparams_{P}^{\intercal} \coloneq \softmax{-\boldsymbol{\totefe}^{\intercal}}\), in order to arrive at the folowing update rule~\footnote{Note that \(\ln \softmax{-\boldsymbol{\totefe}}\), obtained by substituting the definition for \(\policyparams_{P}^{\intercal} \) in~\cref{eq:fe-policy-update-norm}, simplifies to \(-\boldsymbol{\totefe}\) because the natural logarithm is the inverse of raising to the power of \(e\) and because we can ignore the normalizing constant, which would be anyway absorbed into the second softmax used to obtain \(\policyparams_{Q}^{\intercal}\).}:
\begin{equation}\label{eq:fe-policy-update-efe-fe}
    \policyparams_{Q}^{\intercal} = \softmax{-\boldsymbol{\totefe}^{\intercal} - \boldsymbol{\fe}_{\policy}^{\intercal}},
\end{equation}
where each component of \(\boldsymbol{\totefe}\) is the total expected free energy \(\totefe(\policy)\) for a certain policy \(\policy\).
Specifically, for a given policy \(\policy_{k}\), \(\totefe(\policy_{k})\) is defined as the sum of expected free energies at future time steps up to the policy horizon, \(\polhorizon\), i.e.:
\begin{equation}
    \totefe(\policy_{k}) = \sum_{t=\tau + 1}^{\polhorizon} \efe_{t}(\policy_{k}),
\end{equation}
where the expected free energy at time \(t\) for the same policy is specified as follows:
\begin{equation}
    \label{eq:efe-unpacked}
    \begin{split}
        \efe_{t}(\policy_{k}) \coloneq
        & \underbrace{\mathbb{E}_{\varprob{\statevar_{t}|\policy_{k}}} \Bigl[\entropy \bigl[\condprob{\obsvar_{t}}{\statevar_{t}} \bigr] \Bigr]}_{\textsc{Ambiguity}} - \underbrace{\mathbb{E}_{\condprob{\obsvar_{t}}{\statevar_{t}} \varprob{\statevar_{t}|\policy_{k}}} \Bigl[\kldiv \bigl[\varprob{\obsmap|\obs_{t}, \state_{t}}| \varprob{\obsmap} \bigr]\Bigr]}_{\textsc{A-Novelty}} \\
        & + \underbrace{\kldiv \bigl[ \varprob{\statevar_{t}|\policy_{k}} | \pprob{\statevar_{t}} \bigr]}_{\textsc{Risk}} - \underbrace{\mathbb{E}_{\condprob{\obsvar_{t}}{\statevar_{t}} \varprob{\statevar_{t}|\policy_{k}}} \Bigl[\kldiv \bigl[\varprob{\transmap|\obs_{t}, \state_{t}}|\varprob{\transmap}\bigr]\Bigr]}_{\textsc{B-novelty}}.
    \end{split}
\end{equation}

In the above expression, the risk term quantifies the divergence between the variational state distribution and a target distribution (here defined over state random variables, but see below), the ambiguity terms quantifies the uncertainty related to the observation map, and the two novelty terms are expected information gains for the parameters of the observation and the transition maps, thus indicating parts of the generative model that are still inaccurate.

Therefore, according to~\cref{eq:fe-policy-update-efe-fe}, policy probabilities are updated at teach time step based on a combination of negative expected free energies and free energies associated with each policy.
Concretely, this means that a policy will become more probable to the extent that it minimises free energy and expected free energy.
A policy that minimises free energy represents a sequence of action that is most likely associated with the observations collected so far, i.e., there is a good chance that the policy's actions produced a sequence of states emitting those observations.
A policy that minimises expected free energy tries to makes sure that the agent reaches its goals (i.e., by having a low risk, equivalently: a high \emph{instrumental} or \emph{extrinsic} value) and that the agent explores sufficiently enough the environment to acquire relevant information about emission and transition maps (i.e., by being associated with low ambiguity and high novelty terms, equivalently: a high \emph{epistemic} or \emph{intrinsic} value).

An informative and correct probability distribution over policies \(\varprob{\policy}\) is then a crucial component of the decision-making step of an active inference agent, involving the selection of what action to perform next.
There are different ways to specify an action-selection procedure or decision rule, e.g., an agent could pick the action the most probable policy suggests at time \(t\).
Another option would be to pick the action associated with the highest sum of probability mass coming from each policy that suggests that action at \(t\).
The latter, which is used in the following experiments, is known as \emph{Bayesian model average}~\parencite{DaCosta2020b} and can be formally stated as follows:
\begin{equation}\label{eq:bayes-m-avg-policy}
    \action_{t} = \argmax_{\action \in \actionspace} \Biggl(\sum_{\policy_{k} \in \policyspace} \delta_{\action, c^{\policy_{k}}_{t}} \varprob{\policy_{k}} \Biggr),
\end{equation}
where the Kronecker delta $\delta_{\action, c^{\policy_{k}}_{t}}$ compares a potential action \(\action \in \actionspace\) and the action \(c^{\policy_{k}}_{t}\) that a policy \(\policy_{k}\) dictates at time step \(t\), giving $1$ if they are equal and $0$ otherwise.
This decision rule thus picks the action with the highest marginal probability, \(\prob{\action} = \sum_{\policy_{k} \in \policyspace} \prob{\action, \pi_{k}}\), at time step \(t\).

A crucial piece of information included in the expression for the expected free energy is the prior probability distribution \(\pprob{\cdot}\), which can defined over observation random variables \(\obsvar_{t}\), in the partially observable case, or over state random variables \(\statevar_{t}\), in the fully observable case (e.g., when \(\obsvar_{t} = \statevar_{t}\)).
Henceforth, we will focus on the latter, because we are interested in agents that need to learn an environment's transition model as opposed to the observation map (i.e., the \(\transmap\) matrices instead of the \(\obsmap\) matrix), and refer to \(\pprob{\cdot}\) as the \textbf{preference distribution}~\footnote{A similar argument can be made for partially observable scenarios, i.e., when agents have to learn the observation model \(\obsmap\), with the risk in~\cref{eq:efe-unpacked} written in terms of variational and preference distributions over observation random variables. In particular, the variational distribution over observations is obtained from the agent's current observation model by marginalizing over states random variable, i.e., \(\varprob{\obsvar_{t}|\policy_{k}} = \sum_{\state_{t}} \condprob{\obsvar_{t}}{\statevar_{t}} \varprob{\statevar_{t}|\policy_{k}}\) (see appendix in~\parencite{DaCosta2020b}).}.
In the next section, we will delve into some details about the form of this distribution and on what it accomplishes.

\subsection{Preference distributions in active inference agents}\label{ssec:aif-preferences}

In the discrete state-space setting, the agent's preferences are represented by a categorical distribution, indicated by \(\pprob{\statevar}\) in the fully observable case, that effectively encodes an agent's goals in terms of particular instantiations of the random variable \(\statevar\).
That is, goal-directedness in active inference is cast as the concentration of probability mass on some states more than others, for the possible \(m\) realisations, \(\state^{1}, \dots, \state^{m}\), of \(\statevar\), as indicated by the parameters' vector of \(\pprob{\statevar}\).

The specified preference distribution constrains policy and action selection because the computation of expected free energy depends on it.
Specifically, the risk component of expected free energy at time \(t\) is the KL divergence between the variational distribution \(\varprob{\statevar_{t}|\policy_{k}}\) and the preference distribution \(\pprob{\statevar_{t}}\) (see~\cref{eq:efe-unpacked}).
This KL divergence encapsulates the contrast between how things are and how they should be for the agent: the variational distributions at different time steps indicate to the agent what the most probable states afforded by a policy are (they encode the agent's current beliefs about the consequence of a policy, determined by the learned generative model) whereas the fixed preference distribution indicates what the most probable state \emph{should} be (regardless of any policy).
Selecting a policy that minimises risk means to increase the chance that executing its actions will result in variational beliefs that align with the preference distribution (what the agent wants).
To the extent that the variational beliefs reflect the environment's dynamics, i.e., where the agents most probably will be located (and not just where it believes it will), the minimisation of risk will lead the agent to achieve its goal(s).
In other words, risk quantifies the instrumental value of the policy being evaluated, i.e., whether the agents believes it will lead to the preferred states.
If risk is low for a certain policy, then this policy will have a chance of being selected and bring the agent to one of its preferred states.

The preference distribution specifies a certain goal (e.g., a particular state), potentially connected with the solution of a task, and establishes how strongly an agent wants to achieve it. This distribution needs to be specified in advance since in general active inference agents are not able to learn what they should do (but see~\parencite{Sajid2021a,Shin2022a}).
This can be done in different ways, depending on the application and on what the agent is supposed to do.
There are at least two degrees of freedom in the specification of the agent's preferences: (1) how strongly an agent prefers to achieve a certain goal, and (2) whether the preference distribution specifies different goals at different time steps (potentially indicating to the agent a trajectory through state-space).

\renewcommand{\arraystretch}{3}
\begin{table}[!ht]
    \caption{Types of preferences considered in this work.}
    \centering
    \begin{tabular}{c@{\hspace{.8cm}}>{\centering\arraybackslash}p{4cm}|>{\centering\arraybackslash}p{4cm}}
        & \multicolumn{2}{c}{\textbf{Preference strength}} \\[1em]
        \multirow{2}{*}{\raisebox{-1.0\height}{\rotatebox{90}{\textbf{Time-dependence}}}} & Soft preferences with goal shaping & Hard preferences with goal shaping \\
        \cline{2-3}
        & Soft preferences without goal shaping & Hard preferences without goal shaping \\
    \end{tabular}
    \label{tab:pref-types}
\end{table}

The first dimension relates to how much the probability mass of \(\pprob{\statevar_{t}}\) is concentrated on some states as opposed to others.
In particular, the agent could have \textbf{hard goals} or preferences, meaning that most of the probability mass is concentrated on a single state, $\statevar_{t} = \state^{j}$, while for all states the probabilities are close to 0, thus making $\pprob{\statevar_{t}}$ an approximate delta distribution.
Alternatively, the agent could have \textbf{soft goals} or preferences, in the sense that the probability mass is more evenly distributed among at least two or more states.

The second dimension captures the possible time-dependence of preferences: whether the agent has a different preference distribution for each time step, providing the agent with a distinct goal for each time step in a given time interval.
To understand this, recall that the expected free energy defined by~\cref{eq:efe-unpacked} is computed for each future action that a policy suggests therefore  each \(\pprob{\statevar_{t}}\) for \(t \in [1, \ntime]\) could assign probability mass to states differently.
Again, there are at least two possibilities.
With preference or \textbf{goal shaping}, for every future time step \(t\) we specify a different preference distribution capturing what goal(s) the agent ought to strive towards at each of those time steps.
This means that we have effectively a collection of preference distributions defining a desired trajectory in state-space for the agent to follow, which can be viewed as a sequence of sub-goals or intermediate goals the agent needs to obtain before reaching their main goal (e.g., at the last time step).

In contrast, \emph{without} preference or goal shaping, there is only a single, fixed preference distribution that is used to compute the risk term at each time step.
This distribution could specify hard or soft goals and represents the “final” distribution over states to which the agent tends.
In other words, the agent is given one or more goals to obtain as soon as possible, but no indication of what intermediate goals to pursue to get there.

In the active inference literature, the specifications of hard goals and soft goals have often been used in grid world simulations (see, e.g., \parencite{Friston2016a,Friston2017c,Friston2017b,Sales2019a} for hard goals, and~\parencite{Friston2021b} for soft goals).
The presence or absence of goal shaping implicitly appears in active inference works that involve common reinforcement learning environments, where the agent's preferences are encoded as states or observations that yield high rewards, and depends on whether the environment's reward function was designed to provide a dense vs.\ sparse reward signal~\parencite{Tschantz2020b,Tschantz2020d,Tschantz2020c,Catal2019a,Millidge2020e}.

In the experiment described in the next section, we consider active inference agents in a grid-world navigation task, each with one of the four possible preference distributions (determined by the dimensions described above), and examine how the specification of preference distribution affects their ability to reach their final goal, i.e., being in a particular maze's tile, that corresponds with the final solution of the task.
In particular, agents with soft goals and goal shaping have different preference distributions \(\pprob{\statevar_{t}}, \dots, \pprob{\statevar_{\ntime}}\), with the probability mass distributed among states, based on the Manhattan distance of each state from a state representing an intermediate goal (tile position), at \(t\), or from the state corresponding to the final goal, at \(\ntime\).
Similarly, agents with hard goals and goal shaping have different preference distributions, but this time each one approximates a Dirac delta.
Without goal shaping, agents with soft goals have a single preference distribution, \(\pprob{\statevar}\), used to compute the risk in the expected free energy at each time step in the episode, again with a more even distribution of probability mass (as explained above).
Similarly, agents with hard goals have a single preference distribution that assigns most of the probability mass to the final goal.

\section{Results}\label{sec:results}

We trained action-unaware active inference agents, i.e., agents that have no access to previously executed actions (see~\parencite{Torresan2025a} for details), with different preferences in a \(3 \times 3\) grid world.
All agents start in the top left corner and their goal is to reach the bottom right corner, with no obstacles. 
We specified preference distributions in four different ways, based on~\cref{tab:pref-types}.

The problem is simplified to be a fully observable MDP (with \(\obsmap\) diagonal and known to the agent), with deterministic but unknown state transitions \(\transmap\).
We trained 10 agents of each kind for 200 episodes of 5 steps each, i.e., \(\ntime = 5\), with a policy horizon \(\polhorizon = 4\), giving us at most 256 policies to evaluate, to allow for convergence.
Of the 256 policies, 6 are task-solving, since they allow an agent to reach the bottom right corner of the grid via a combination of two \(\downarrow\) and two \(\rightarrow\) actions, in any order.
The remaining 250 are task-failing since they don't lead to the final goal.

\begin{figure}[!ht]
  \centering
  \begin{subfigure}{0.45\textwidth}\label{fig:gridw9-agents-goal-softgs}
    \centering
    \includegraphics[width=\textwidth]{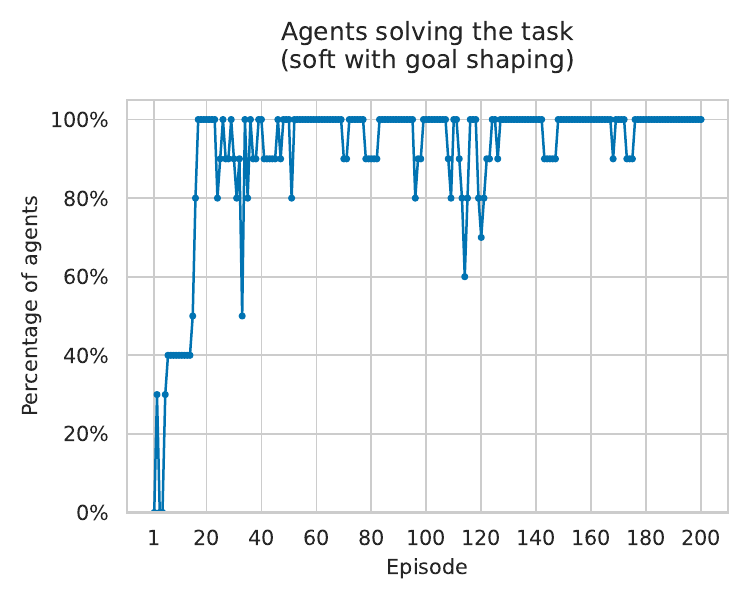}
  \end{subfigure}
  \begin{subfigure}{0.45\textwidth}\label{fig:gridw9-agents-goal-hardgs}
    \centering
    \includegraphics[width=\textwidth]{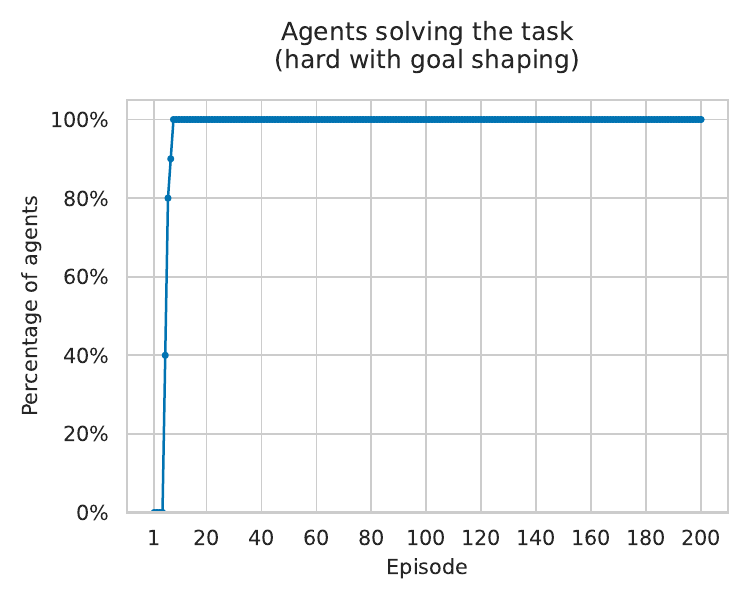}
  \end{subfigure}
  \begin{subfigure}{0.45\textwidth}\label{fig:gridw9-agents-goal-soft}
    \centering
    \includegraphics[width=\textwidth]{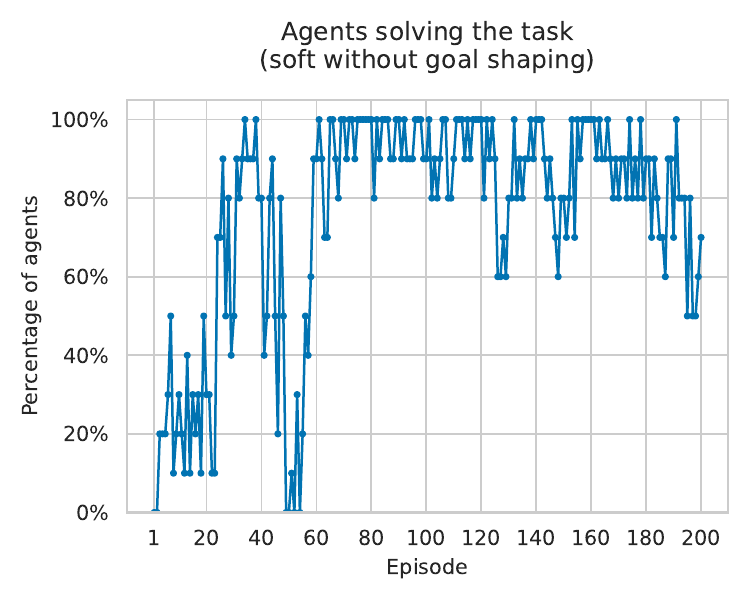}
  \end{subfigure}
  \begin{subfigure}{0.45\textwidth}\label{fig:gridw9-agents-goal-hard}
    \centering
    \includegraphics[width=\textwidth]{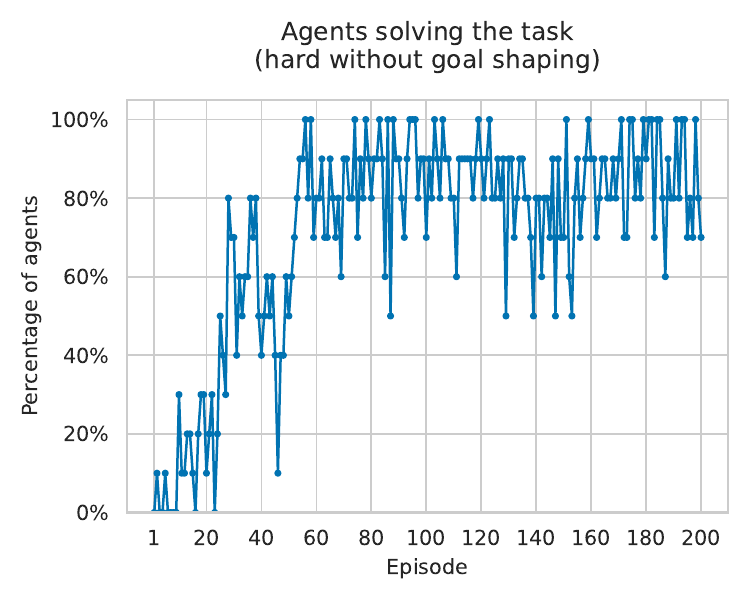}
  \end{subfigure}
  \caption{Percentage of agents reaching the goal state in each episode in the 5-step grid world (10 agents for each subplot).}\label{fig:gridw9-agents-goal}
\end{figure}

We start by comparing the percentage of agent solving the task across episodes in \cref{fig:gridw9-agents-goal}.
Goal-shaped agents learn more quickly and perform better overall than agents without goal shaping, which is to be expected since the former know exactly what intermediate goal to aim for at each time step, guiding them to reach their final goal (bottom right corner).

With goal-shaping, having hard goals is also a further advantage as all agents are able to find the path to the final goal in a handful of episodes and they stick to it for the rest of the experiment, i.e., a \(100\%\) success rate is consistently achieved from episode 8 onwards (see top-right plot in~\cref{fig:gridw9-agents-goal}).
In contrast, agents with soft goals take longer to find their way to the final goal and the percentage of successful agents falls below \(100\%\) in several episodes throughout the experiment, even after the main learning phase is over, i.e., from episode 17 onwards (compare top-left and top-right plots in~\cref{fig:gridw9-agents-goal}).

\begin{figure}[H]
  \centering
  \begin{subfigure}{0.45\textwidth}\label{fig:gridw9-aif-paths-softgs-sv}
    \centering
    \includegraphics[width=\textwidth]{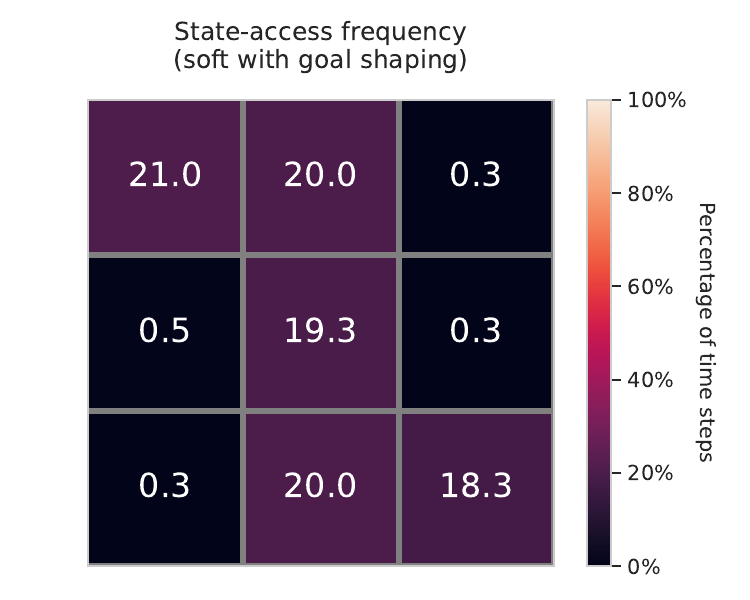}
  \end{subfigure}
  \begin{subfigure}{0.45\textwidth}\label{fig:gridw9-aif-paths-hardgs-sv}
    \centering
    \includegraphics[width=\textwidth]{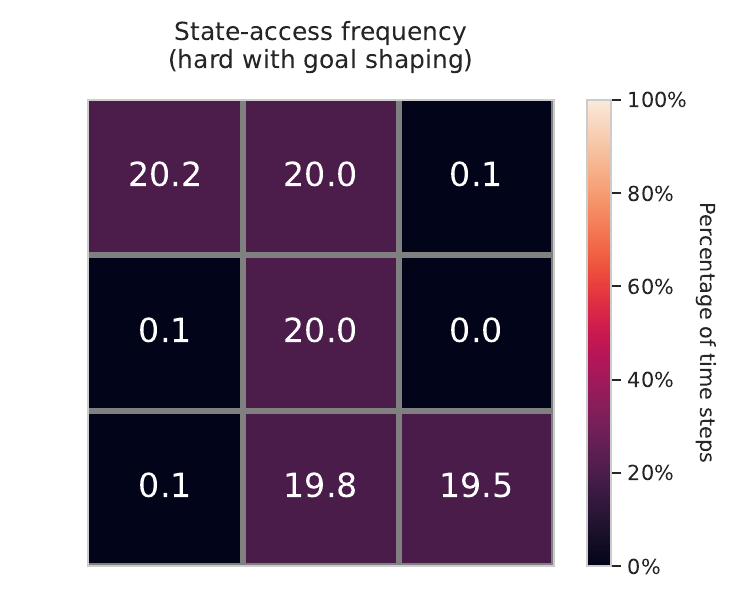}
  \end{subfigure}
  \begin{subfigure}{0.45\textwidth}\label{fig:gridw9-aif-paths-soft-sv}
    \centering
    \includegraphics[width=\textwidth]{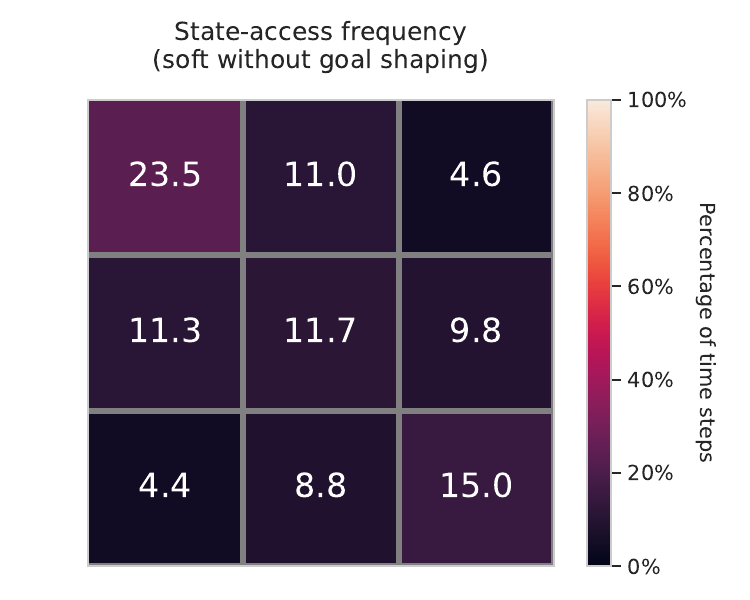}
  \end{subfigure}
  \begin{subfigure}{0.45\textwidth}\label{fig:gridw9-aif-paths-hard-sv}
    \centering
    \includegraphics[width=\textwidth]{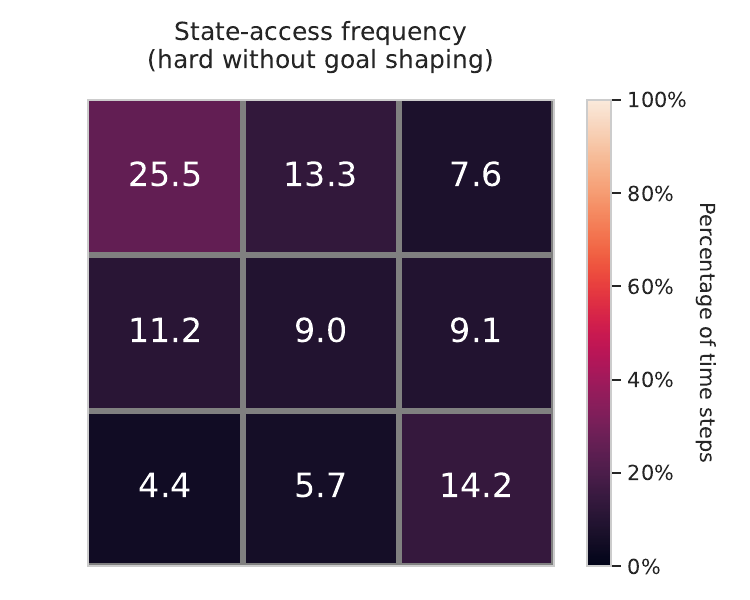}
  \end{subfigure}
  \caption{State-access frequency in the experiment (showing average of 10 agents).}\label{fig:gridw9-sv}
\end{figure}

Without goal-shaping, both kinds of agent take longer to learn and succeed: after an initial phase of performance improvements, there is a dramatic drop in the percentage of successful agents at around episode 50, followed by a quick recovery and with at least \(60\%\) of the agents being able to reach the final goal in most episodes until the end of the experiment (see bottom-left and bottom-right plots in~\cref{fig:gridw9-agents-goal}).
While the performance drop at around episode 50 is less pronounced for agents with hard goals, overall they appear to reach the final goal fewer times than agents with soft goals.
The percentages of time the final goal was reached during the experiment, \(15\%\) in soft-goal agents and \(14.2\%\) in hard-goal agents, confirm the above observation: without goal-shaping, agents with soft goal are marginally more successful than agents with hard goals (compare heat maps in~\cref{fig:gridw9-sv}).

These results indicate that agents with different kinds of preferences can successfully learn relevant aspects of the transition dynamics and solve the task.
Agents that are given more information, by means of intermediate goals (goal shaping) leading to the final goal have an advantage and can complete the task a higher number of times.
At the same time, soft goals enable better performance when goal shaping is not available but hard goals provides a further advantage when combined with goal shaping.

For a more in-depth understanding of how these differences in performance are brought about by the choice of preference distribution, we analyse perceptual inference and planning/decision-making in the simulated agents by considering related metrics, chiefly, policy-conditioned free energies and expected free energies, respectively.

\paragraph{Perceptual inference over trajectories}\label{ssec:perceptual-inf}
The successful performance of each type of agent hinges upon an appropriate inference of its current state from the received observations, corresponding to the perceptual stage in active inference.
This is achieved via free energy minimisation.
In episodic setups, this minimisation happens at different time steps.
Here we look at how free energy is minimised, across episodes, at the last step of each episode, as done in~\cite{Torresan2025a}.
At the last step, an agent has to evaluate the entire sequence of observations associated to a full episode so to infer the most probable sequence of states that have been visited, conditional on having executed a certain policy.
This thus reveals how agents infer their past and current states.
Policies that correspond to state trajectories that agree with the received observations will be associated with low values of (policy-conditioned) free energy.  

\begin{figure}[H]
  \centering
  \begin{subfigure}{0.45\textwidth}\label{fig:gridw9-aif-paths-softgs-policies-fe-step5}
    \centering
    \includegraphics[width=\textwidth]{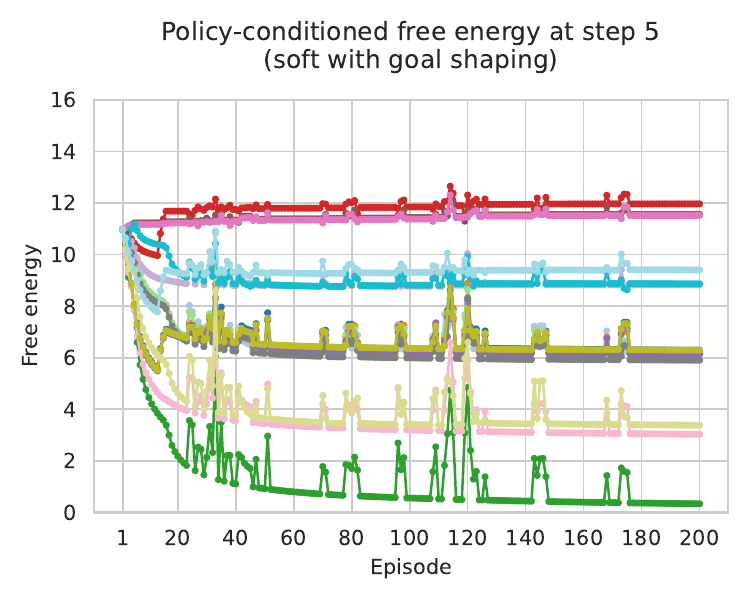}
  \end{subfigure}
  \begin{subfigure}{0.45\textwidth}\label{fig:gridw9-aif-paths-hardgs-policies-fe-step5}
    \centering
    \includegraphics[width=\textwidth]{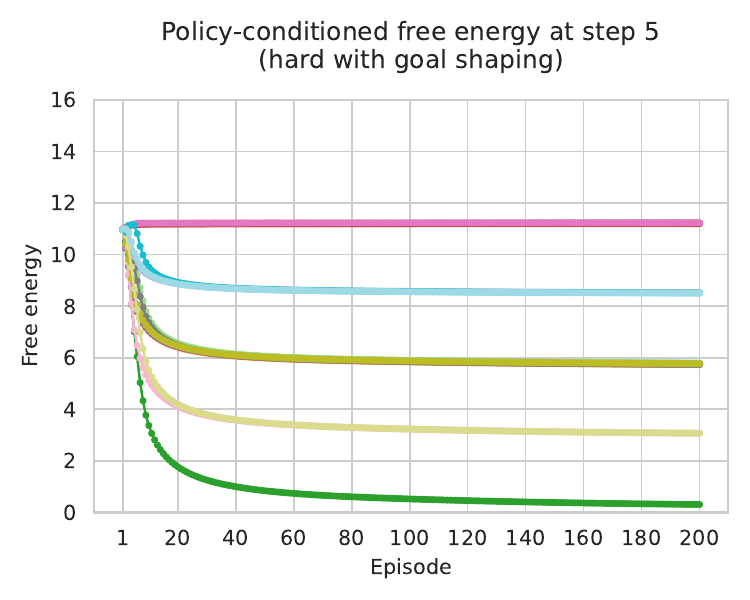}
  \end{subfigure}
  \begin{subfigure}{0.45\textwidth}\label{fig:gridw9-aif-paths-soft-policies-fe-step5}
    \centering
    \includegraphics[width=\textwidth]{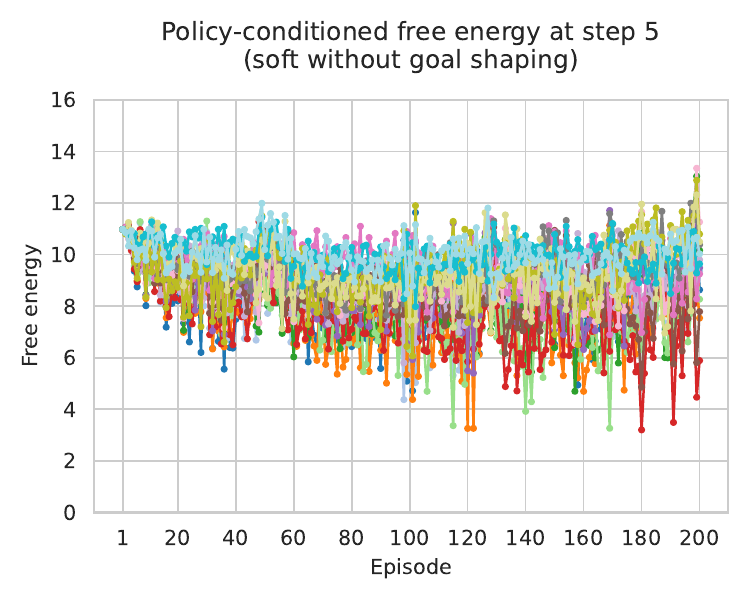}
  \end{subfigure}
  \begin{subfigure}{0.45\textwidth}\label{fig:gridw9-aif-paths-hard-policies-fe-step5}
    \centering
    \includegraphics[width=\textwidth]{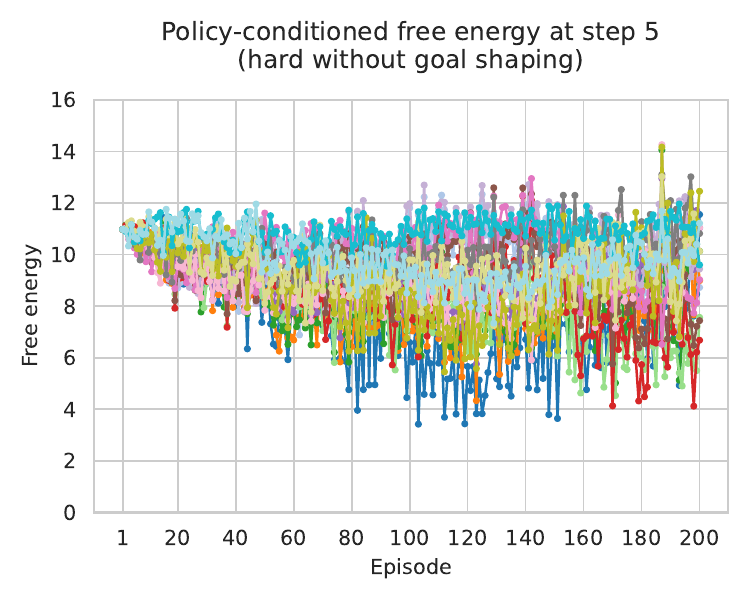}
  \end{subfigure}
  \begin{subfigure}{0.65\textwidth}
    \centering
    \includegraphics[width=\textwidth]{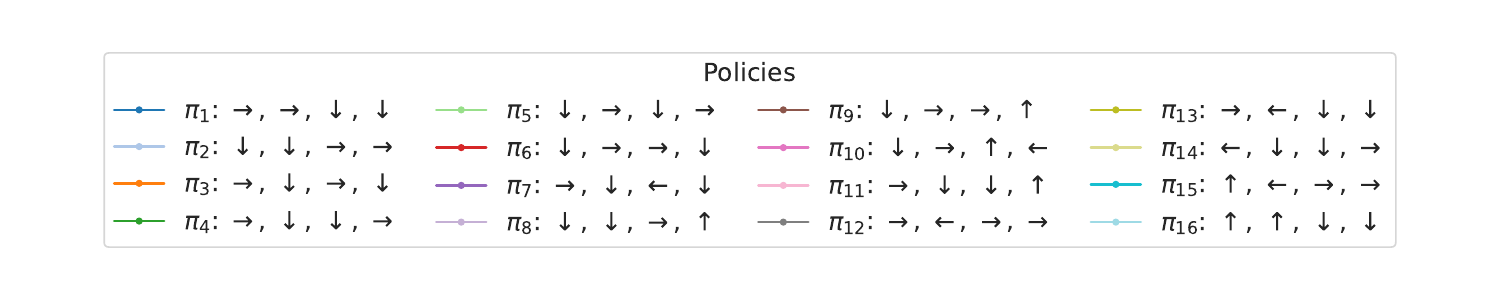}
  \end{subfigure}

  \caption{Policy-conditioned free energies at step 5 across episodes (showing average of 10 agents).}\label{fig:gridw9-policies-fe-step5}
\end{figure}

Plots of the policy-conditioned free energies at step 5, in~\cref{fig:gridw9-policies-fe-step5}, reveal which policy best accounts for the observations collected in each episode.
For all the figures involving policy-conditioned free energies, we selected 16 policies, among the 256, including the 6 task-solving policies that lead to the final goal.
These plots show that the presence vs.\ the absence of goal shaping affects what policy yields the lowest policy-conditioned free energy, i.e., what policy the agent most likely executed because it is inferred to be more consistent with the observations, thereby minimizing the associated policy-conditioned free energy. 

In general, we can see that goal-shaped agents learned to pick \(\policy_{4}\), one of the task-solving ones, as the preferred policy and stick with it throughout the experiment (see top-left and top-right plots in~\cref{fig:gridw9-policies-fe-step5}).
This yields a consistent minimisation of the corresponding policy-conditioned free energy across episodes: agents at the end of each episode correctly infer that \(\policy_{4}\) was the most likely executed policy.
In contrast, for agents without goal shaping there is no policy that consistently minimises the policy-conditioned free energy, see bottom-left and bottom-right plots in~\cref{fig:gridw9-policies-fe-step5}.
This is due to the fact that these agents are not constrained by their preference distribution to follow a particular trajectory in state-space but only to reach the final goal as soon as possible.
Therefore, they attempt different task-solving or task-failing policies, and no clear demarcation between these is visible in the plots.
This is because in different episodes the observation sequence is correctly inferred to be more consistent with different, task-solving or task-failing, policies.

\begin{figure}[h!]
  \centering
  \begin{subfigure}{0.45\textwidth}\label{fig:gridw9-aif-paths-softgs-policies-fe-step}
    \centering
    \includegraphics[width=\textwidth]{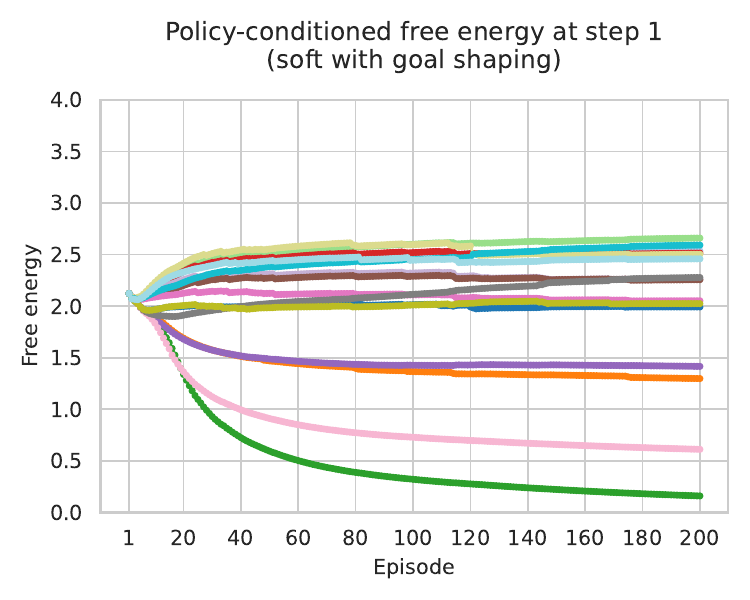}
  \end{subfigure}
  \begin{subfigure}{0.45\textwidth}\label{fig:gridw9-aif-paths-hardgs-policies-fe-step1}
    \centering
    \includegraphics[width=\textwidth]{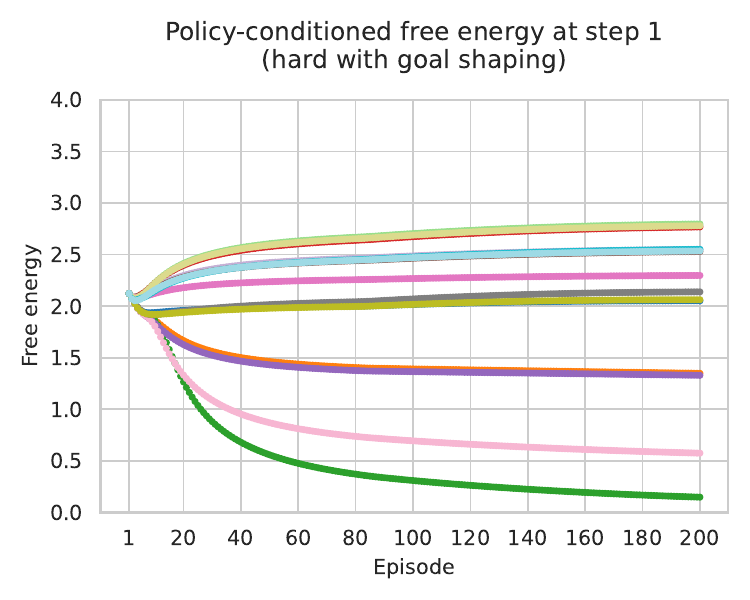}
  \end{subfigure}
  \begin{subfigure}{0.45\textwidth}\label{fig:gridw9-aif-paths-soft-policies-fe-step1}
    \centering
    \includegraphics[width=\textwidth]{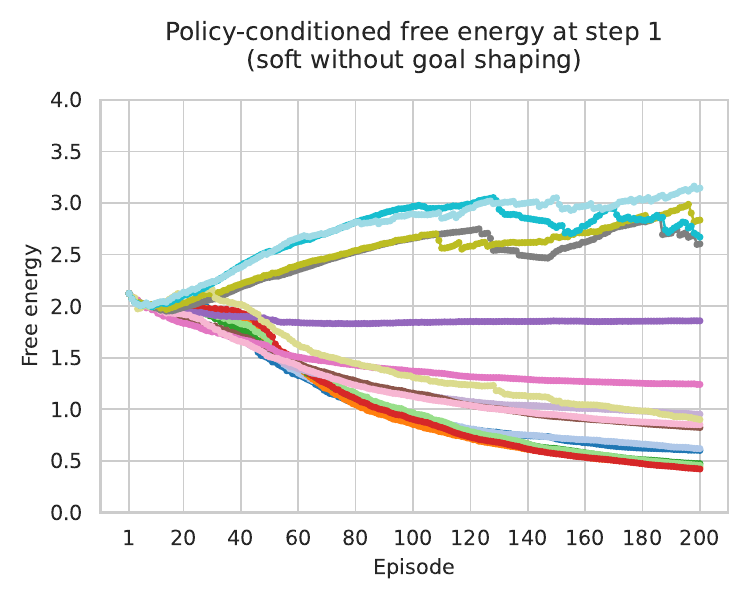}
  \end{subfigure}
  \begin{subfigure}{0.45\textwidth}\label{fig:gridw9-aif-paths-hard-policies-fe-step1}
    \centering
    \includegraphics[width=\textwidth]{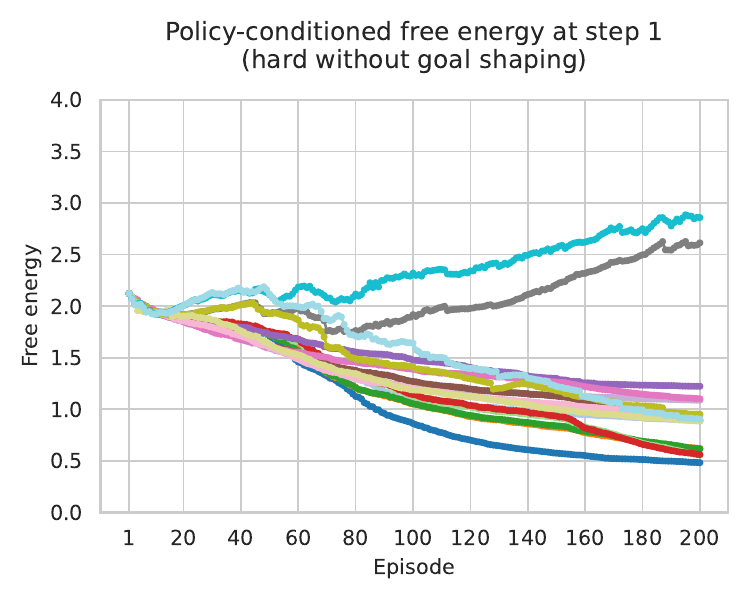}
  \end{subfigure}
  \begin{subfigure}{0.65\textwidth}
    \centering
    \includegraphics[width=\textwidth]{gridw9_aif_policies_legend}
  \end{subfigure}

  \caption{Policy-conditioned free energies at step 1 across episodes (showing average of 10 agents).}\label{fig:gridw9-policies-fe-step1}
\end{figure}

\paragraph{Policy probabilities}\label{ssec:policy-probs}
To examine how the choice of preference distribution affects planning and decision-making in each episode, we compare the respective contributions of the policy-conditioned free energies and expected free energies to the policy probabilities at the beginning of each episode, i.e., at step 1.
This step was chosen because it is the most indicative of how well the agent has learned the transition model, since at this step the agent performs perceptual inference with respect to future state, before any observation is received (with the exception of that from the starting state), and plans for the entire episode, see also~\cite{Torresan2025a}.
In general, we found that policy-conditioned free energies and expected free energies do not always agree, i.e., by jointly increasing the probability of a policy if it is one that leads to the goal state, revealing a mismatch between perception and planning that depends on the choice of preference distribution.

\begin{figure}[ht!]
  \centering
  \begin{subfigure}{0.45\textwidth}\label{fig:gridw9-aif-paths-softgs-efe-step0}
    \centering
    \includegraphics[width=\textwidth]{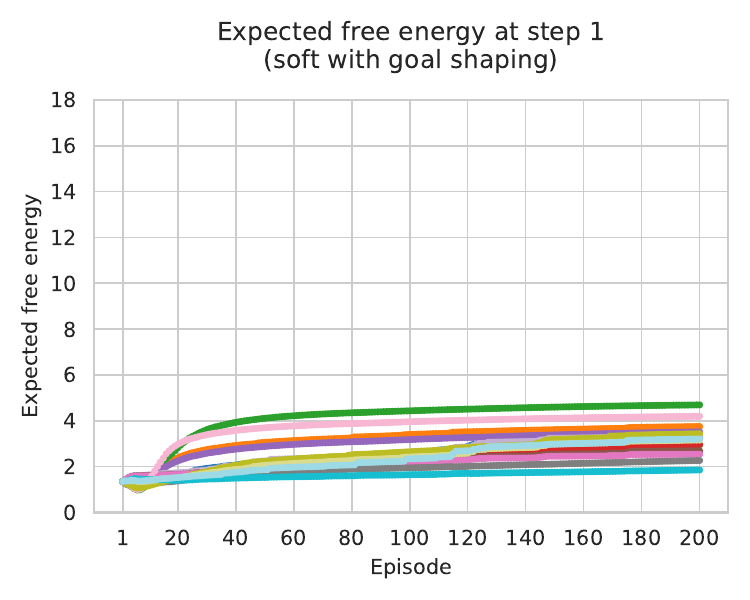}
  \end{subfigure}
  \begin{subfigure}{0.45\textwidth}\label{fig:gridw9-aif-paths-hardgs-efe-step0}
    \centering
    \includegraphics[width=\textwidth]{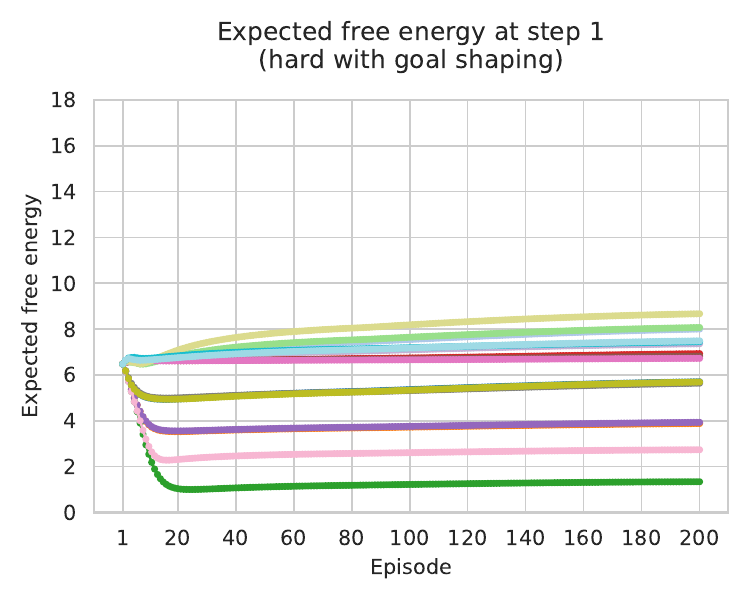}
  \end{subfigure}
  \begin{subfigure}{0.45\textwidth}\label{fig:gridw9-aif-paths-soft-efe-step0}
    \centering
    \includegraphics[width=\textwidth]{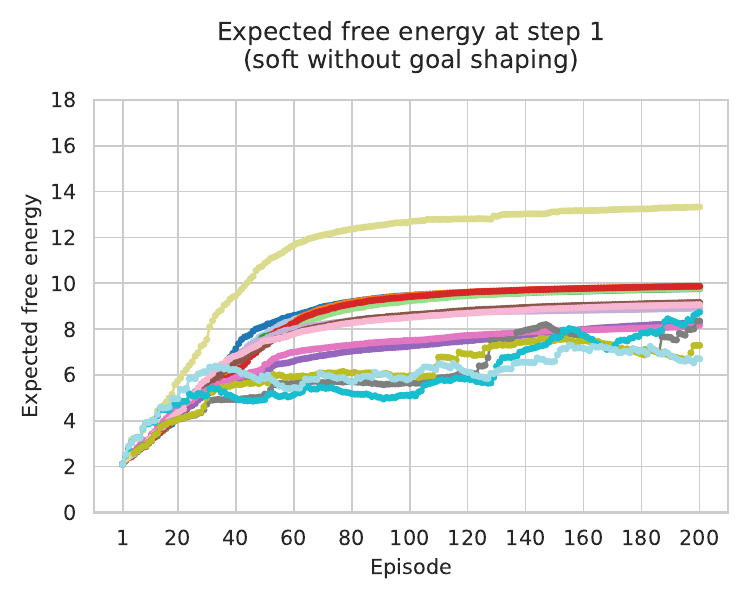}
  \end{subfigure}
  \begin{subfigure}{0.45\textwidth}\label{fig:gridw9-aif-paths-hard-efe-step0}
    \centering
    \includegraphics[width=\textwidth]{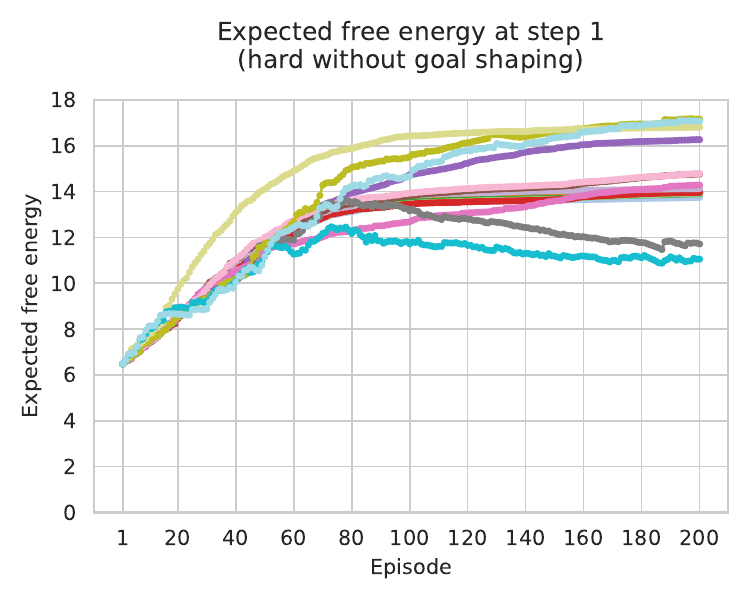}
  \end{subfigure}
  \begin{subfigure}{0.65\textwidth}
    \centering
    \includegraphics[width=\textwidth]{gridw9_aif_policies_legend}
  \end{subfigure}
  \caption{Expected free energy for each policy across episodes (showing average of 10 agents).
  Notice that we only draw 16 expected free energies, representative of the possible 256.}\label{fig:gridw9-efe-step0}
\end{figure}

\begin{figure}[h!]
  \centering
  \begin{subfigure}{0.45\textwidth}\label{fig:gridw9-aif-paths-softgs-first-pi-probs-offset0}
    \centering
    \includegraphics[width=\textwidth]{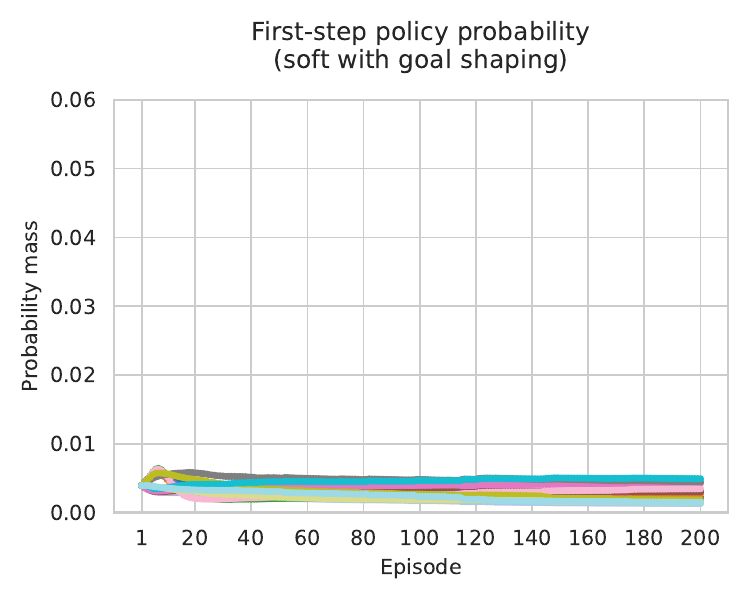}
  \end{subfigure}
  \begin{subfigure}{0.45\textwidth}\label{fig:gridw9-aif-paths-hardgs-first-pi-probs-offset0}
    \centering
    \includegraphics[width=\textwidth]{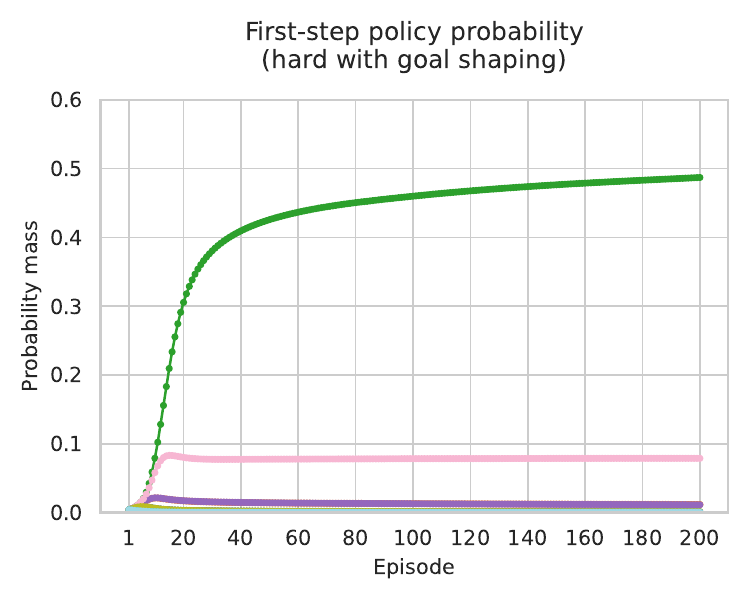}
  \end{subfigure}
  \centering
  \begin{subfigure}{0.45\textwidth}\label{fig:gridw9-aif-paths-soft-first-pi-probs-offset0}
    \centering
    \includegraphics[width=\textwidth]{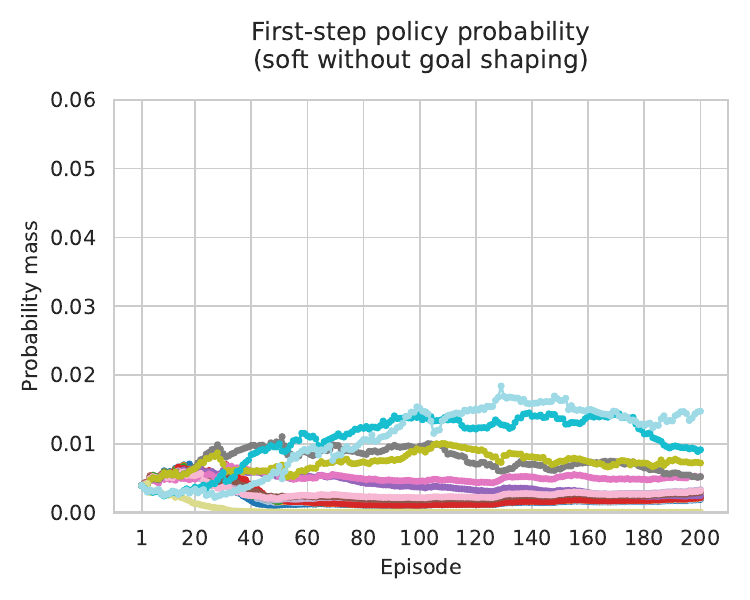}
  \end{subfigure}
  \begin{subfigure}{0.45\textwidth}\label{fig:gridw9-aif-paths-hard-first-pi-probs-offset0}
    \centering
    \includegraphics[width=\textwidth]{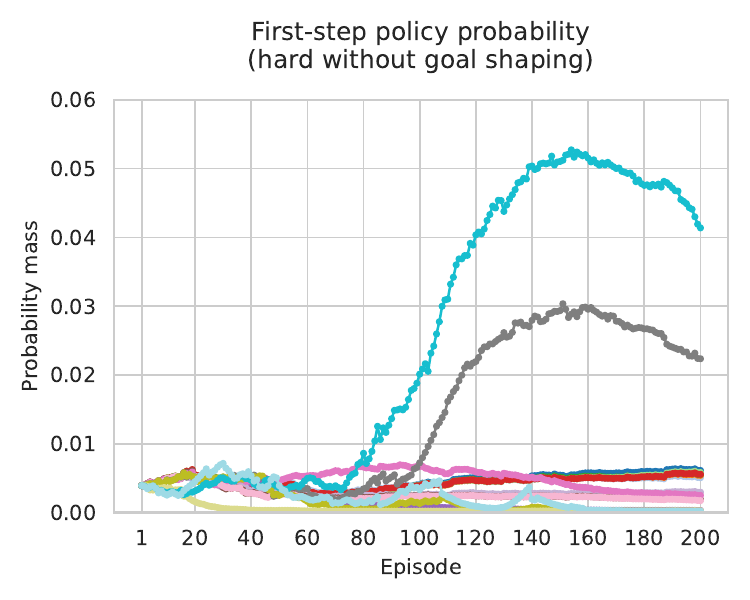}
  \end{subfigure}
  \begin{subfigure}{0.65\textwidth}
    \centering
    \includegraphics[width=\textwidth]{gridw9_aif_policies_legend}
  \end{subfigure}
  \caption{Policies probabilities at the first step of each episode (showing average of 10 agents).
  Notice we only draw 16 representative policies out of the possible 256.}\label{fig:gridw9-first-pi-probs}
\end{figure}

Policy probabilities are computed as a softmax of the sum of negative policy-conditioned free energy and negative expected free energy.
Therefore, the policy probabilities at step 1 in~\cref{fig:gridw9-first-pi-probs} are directly shaped by the contributions of policy-conditioned free energy and expected free energy at step 1, illustrated for each agent by~\cref{fig:gridw9-policies-fe-step1} and~\cref{fig:gridw9-efe-step0}.

In particular, we observe that in agents with soft goals and goal shaping, the policy-conditioned free energy assigned to the task-solving policy \(\policy_{4}\) is low, see the top-left plot in~\cref{fig:gridw9-policies-fe-step1}, and this makes it more probable.
On the other hand, the associated expected free energy is high, see the top-left plot in~\cref{fig:gridw9-efe-step0}, and this makes it less probable.
Overall, this means that the resulting policy probability is not significantly different from that of other policies, see top-left plot in~\cref{fig:gridw9-first-pi-probs}.

In agents without goal shaping the opposite happens for some task-failing policies, which are characterised by high policy-conditioned free energy values, see the bottom plots in~\cref{fig:gridw9-policies-fe-step1}.
This brings their probability down.
At the same time, their expected free energy is remarkably low, see the bottom plots in~\cref{fig:gridw9-efe-step0}, pushing the policy probability up.
This means that these task-failing policies become more probable than some task-solving policies, see bottom plots in~\cref{fig:gridw9-first-pi-probs} (see also~\cref{sssecx:gridw9-efe-breakdown} for a breakdown of these expected free energies showing how their risk component is characterised by a significant decrease over the course of the experiment).
Crucially, the policy-conditioned free energies and expected free energies agree for all policies only in agents with goal shaping and hard goals. 
In turn this means that one of the task-solving policies, \(\policy_{4}\), is correctly identified as the most likely, i.e., that is the policy for which both quantities are the lowest (see top-right plot in~\cref{fig:gridw9-first-pi-probs}, and compare top-right plots in~\cref{fig:gridw9-policies-fe-step1} and~\cref{fig:gridw9-efe-step0}).

\paragraph{Action selection and action probabilities}\label{ssec:action-selection}
Despite not inferring any of the task-solving policies as more likely than task-failing ones, some classes of agents appear to anyway reach the final goal state, albeit with varying degrees of success.
We refer in particular to agents without goal-shaping and agents with soft goals and goal-shaping, see~\cref{fig:gridw9-first-pi-probs}.
This seemingly unintuitive finding can be explained by the fact that agents do not pick the action suggested by the most probable policy at a certain time step but
the one with the highest marginal probability, see~\cref{eq:bayes-m-avg-policy}.
This action-selection mechanism, referred to as Bayesian model average in~\parencite{DaCosta2020b}, prevents agents from performing the potentially wrong action of a task-failing policy that was mistakenly inferred as more probable than others (perhaps only marginally more probable). 

\begin{figure}[ht!]
  \centering
  \begin{subfigure}{0.45\textwidth}\label{fig:gridw9-aif-au-softgs-s1-action-probs-avg}
    \centering
    \includegraphics[width=\textwidth]{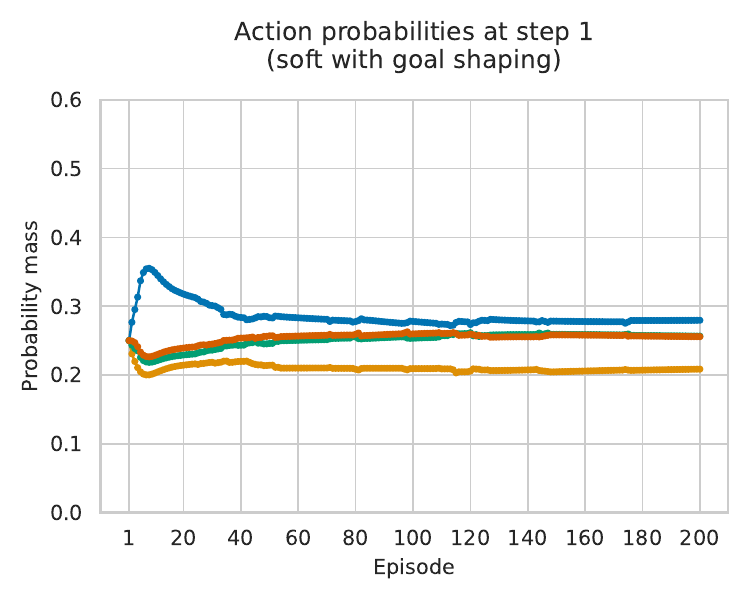}
  \end{subfigure}
  \begin{subfigure}{0.45\textwidth}\label{fig:gridw9-aif-au-hardgs-s1-action-probs-avg}
    \centering
    \includegraphics[width=\textwidth]{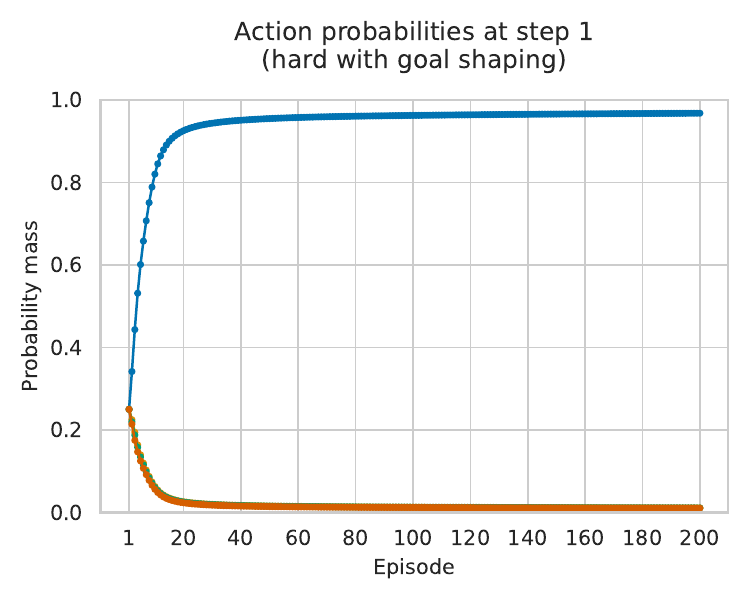}
  \end{subfigure}
  \centering
  \begin{subfigure}{0.45\textwidth}\label{fig:gridw9-aif-au-soft-s1-action-probs-avg}
    \centering
    \includegraphics[width=\textwidth]{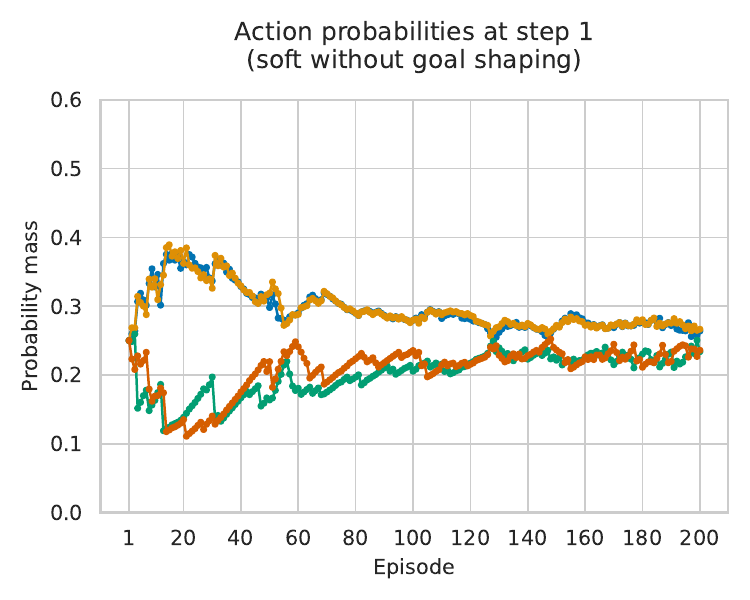}
  \end{subfigure}
  \begin{subfigure}{0.45\textwidth}\label{fig:gridw9-aif-au-hard-s1-action-probs-avg}
    \centering
    \includegraphics[width=\textwidth]{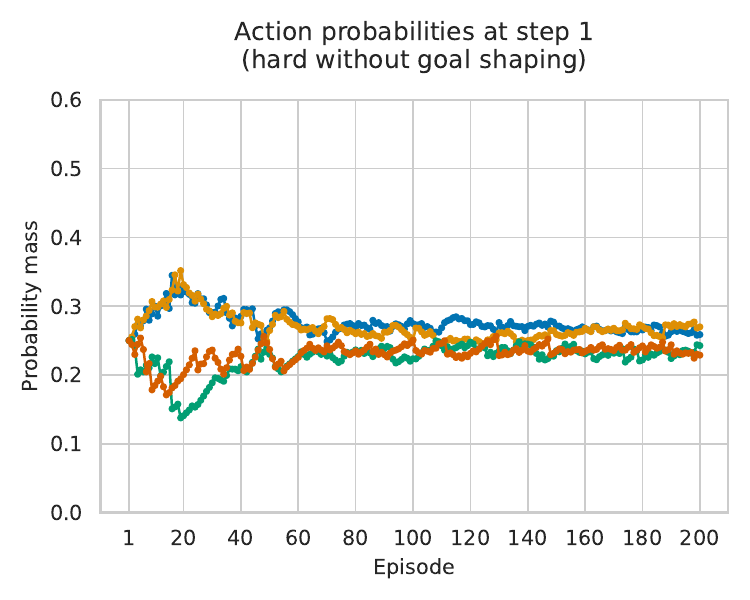}
  \end{subfigure}
  \begin{subfigure}{0.65\textwidth}
    \centering
    \includegraphics[width=\textwidth]{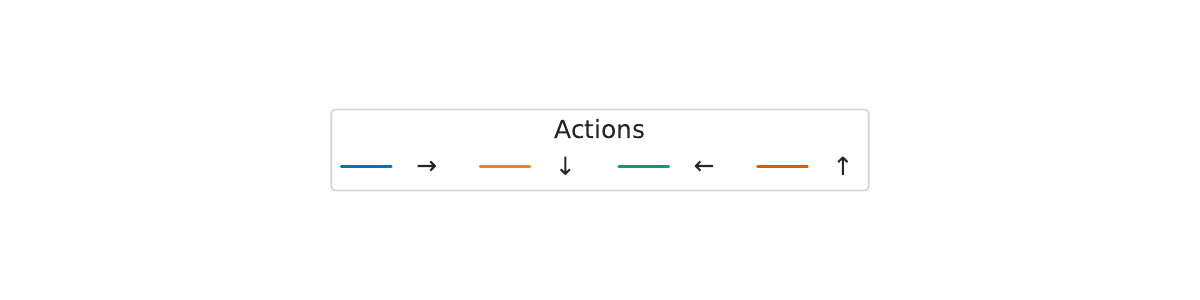}
  \end{subfigure}
  \caption{Action probabilities at step 1 of each episode (average of 10 agent).}\label{fig:gridw9-s1-action-probs-avg}
\end{figure}

Thanks to this, and as we see in~\cref{fig:gridw9-s1-action-probs-avg}, all agents assigned a sufficiently high probability mass to the six task-solving policies so that actions \(\downarrow\) or \(\rightarrow\) became more likely to be picked.
Note, as a reminder, that all agents start from the top left corner of a $3 \times 3$ grid and need to read the goal at the bottom right corner.
This means that all the task-solving policies are a combination of two \(\downarrow\) and two \(\rightarrow\) actions, in any order. 
In this way, most of the time, even agents that didn't infer any of the task-solving policies as more probable (agents without goal-shaping and agents with soft goals and goal-shaping) ended up performing one of the correct sequences of actions to reach the goal state.
This finding highlights the importance that the action-selection mechanism can have depending on the preference distribution given to the agent, e.g., one that may lead to a task-failing policy being inferred as more probable.

More in detail, \cref{fig:gridw9-s1-action-probs-avg} shows action probabilities at step 1, obtained by marginalising probabilities for each action, see~\cref{eq:bayes-m-avg-policy}.
Consistently with the fact that goal-shaped agents perform best, in general these agents assign the majority of probability mass at step 1 to action \(\rightarrow\), i.e., the first correct action of their preferred path to the final goal state (see top-left and top-right plots in~\cref{fig:gridw9-s1-action-probs-avg}). 
In agents without goal shaping something similar occurs.
However, since these agents do not have a preferred path to the goal state, the probability mass at step 1 is almost evenly assigned to the two correct actions \(\rightarrow\) and \(\downarrow\) of the available paths to the final goal state (see bottom-left and bottom-right plots in~\cref{fig:gridw9-s1-action-probs-avg}). 
In both kinds of agents, action probabilities at other time steps also reflect this preference for actions that form one of the correct action sequences to reach the final goal (see plots for the other steps in~\cref{sssecx:avg-action-probs}).

\begin{figure}[H]
  \centering
  \begin{subfigure}{0.45\textwidth}\label{fig:gridw9-aif-paths-softgs-s1-action-probs}
    \centering
    \includegraphics[width=\textwidth]{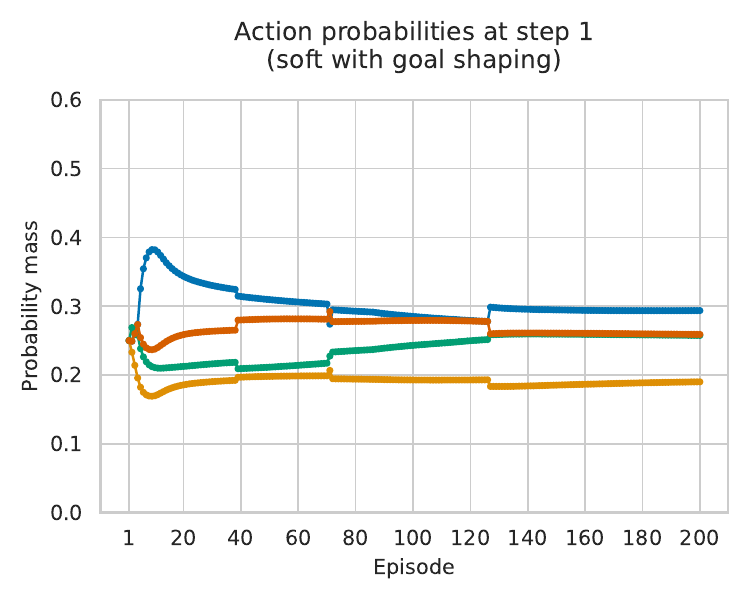}
  \end{subfigure}
  \begin{subfigure}{0.45\textwidth}\label{fig:gridw9-aif-paths-hardgs-s1-action-probs}
    \centering
    \includegraphics[width=\textwidth]{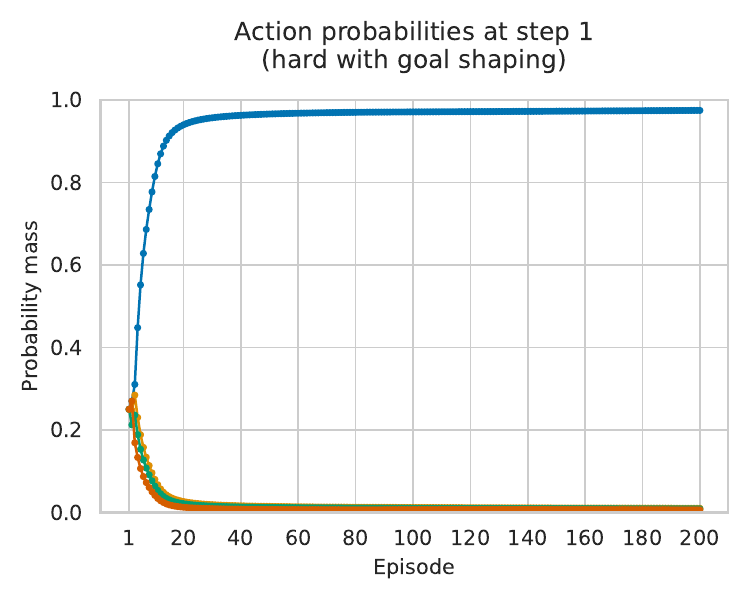}
  \end{subfigure}
  \centering
  \begin{subfigure}{0.45\textwidth}\label{fig:gridw9-aif-paths-soft-s1-action-probs}
    \centering
    \includegraphics[width=\textwidth]{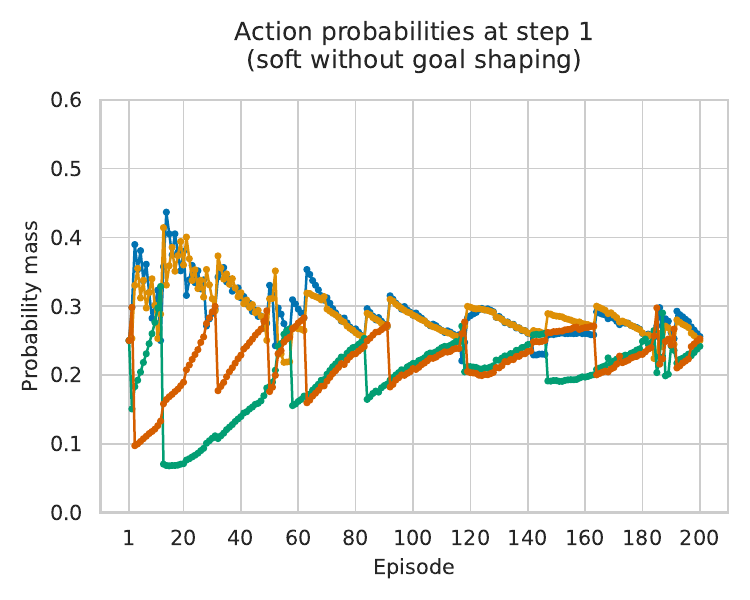}
  \end{subfigure}
  \begin{subfigure}{0.45\textwidth}\label{fig:gridw9-aif-paths-hard-s1-action-probs}
    \centering
    \includegraphics[width=\textwidth]{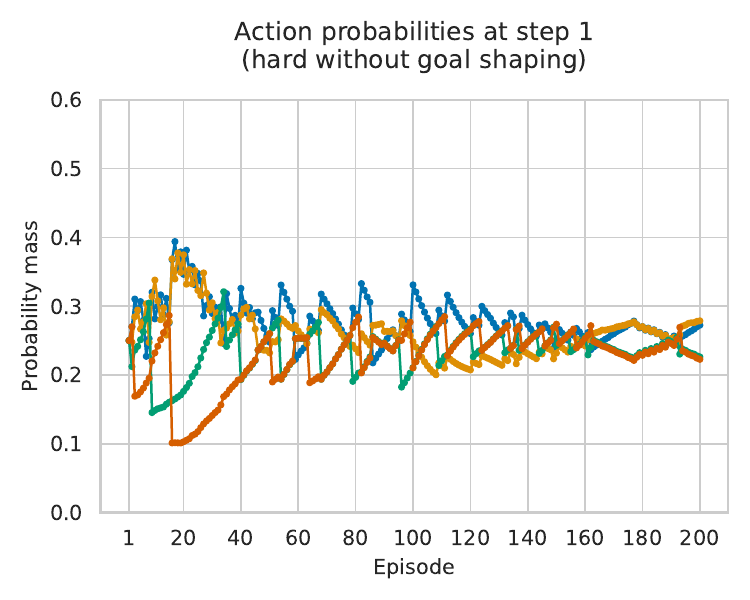}
  \end{subfigure}
  \begin{subfigure}{0.65\textwidth}
    \centering
    \includegraphics[width=\textwidth]{gridw9_actions_legend}
  \end{subfigure}
  \caption{Action probabilities at step 1 of each episode (showing results for one agent).}\label{fig:gridw9-s1-action-probs}
\end{figure}

\begin{figure}[H]
  \centering
  \begin{subfigure}{0.45\textwidth}\label{fig:gridw9-aif-paths-softgs-s2-action-probs}
    \centering
    \includegraphics[width=\textwidth]{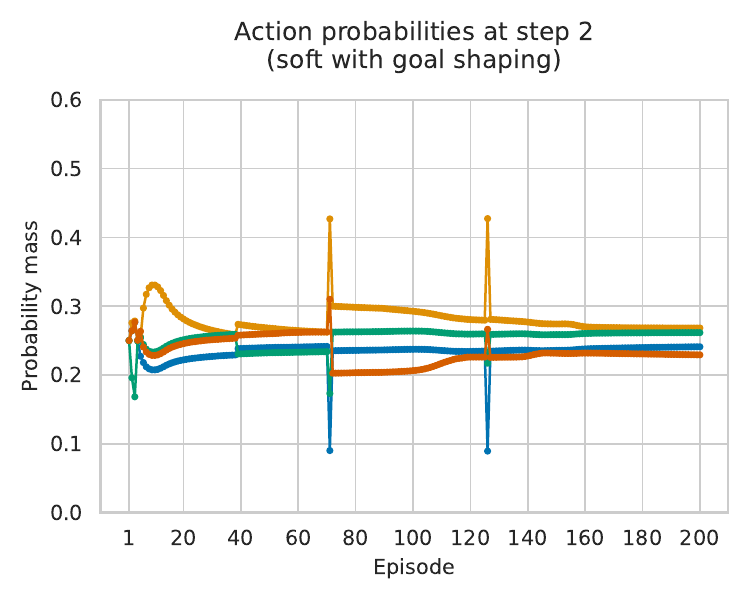}
  \end{subfigure}
  \begin{subfigure}{0.45\textwidth}\label{fig:gridw9-aif-paths-hardgs-s2-action-probs}
    \centering
    \includegraphics[width=\textwidth]{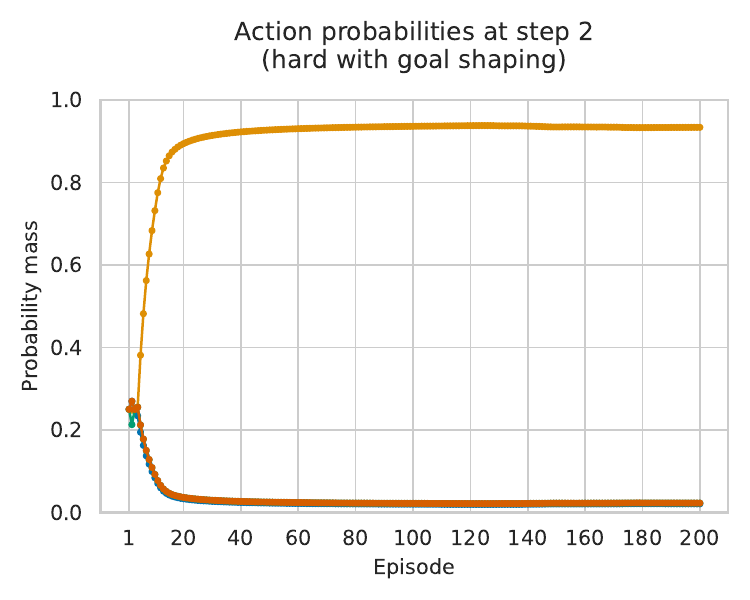}
  \end{subfigure}
  \centering
  \begin{subfigure}{0.45\textwidth}\label{fig:gridw9-aif-paths-soft-s2-action-probs}
    \centering
    \includegraphics[width=\textwidth]{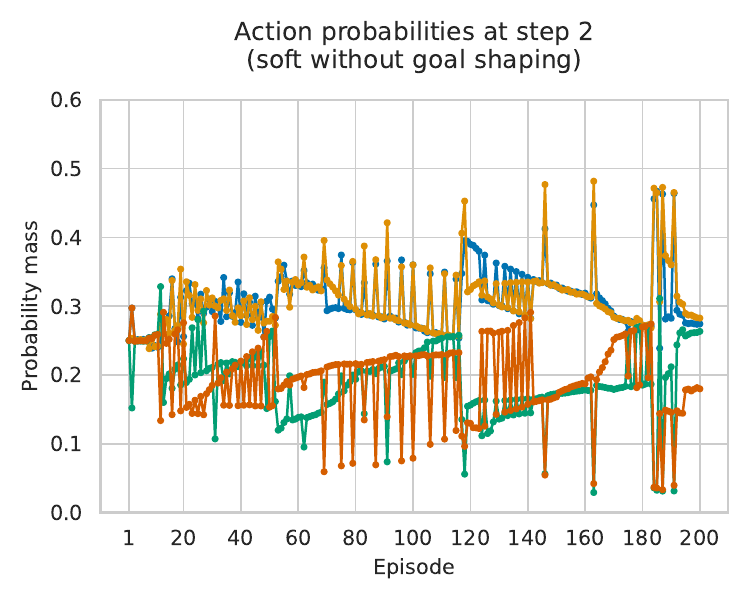}
  \end{subfigure}
  \begin{subfigure}{0.45\textwidth}\label{fig:gridw9-aif-paths-hard-s2-action-probs}
    \centering
    \includegraphics[width=\textwidth]{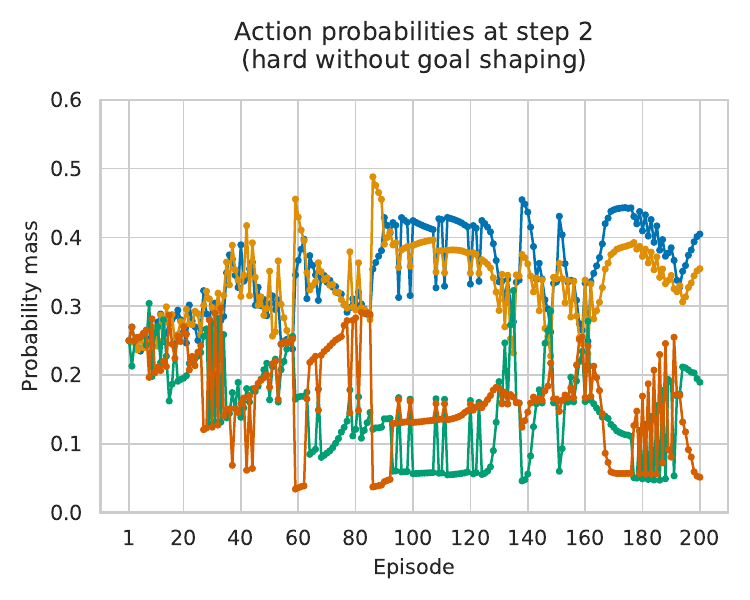}
  \end{subfigure}
  \begin{subfigure}{0.65\textwidth}
    \centering
    \includegraphics[width=\textwidth]{gridw9_actions_legend}
  \end{subfigure}
  \caption{Action probabilities at step 2 of each episode (showing results for one agent).}\label{fig:gridw9-s2-action-probs}
\end{figure}

Since Bayesian model average favours the action backed by most probable policies, these results are not surprising when one of the task-solving policies is significantly more probable than others, i.e., in agents with a hard goal and goal shaping.
But, in agents where this concentration of probability mass over policies does not occur, these findings indicates that the agents are able to spread a sufficiently high degree of probability mass among the task-solving policies.
Most of the time the action selection mechanism can then build the right sequence of action to the goal state.
When this does not occur, it is because agents have overall assigned more probability mass to policies suggesting incorrect actions (for the reasons explained in~\cref{ssec:discus-two}), which are then executed and cause the performance drops seen in~\cref{fig:gridw9-agents-goal}.
These action-selection mistakes become apparent when examining the individiual action probabilities for specific agents, as shown by the plots in~\cref{sssecx:single-action-probs}.
For the two agents without goal shaping, we observe a cyclical increase in the probability of the incorrect actions (most clearly at step 1 and 2, see~\cref{fig:gridw9-s1-action-probs} and~\cref{fig:gridw9-s2-action-probs}), eventually leading to some episodes where these incorrect actions are selected over the correct ones, albeit only by a small margin. 

\section{Discussion}\label{sec:discussion}

\subsection{Learning and performance elicited by different preference distributions}\label{ssec:discus-one}
Simulations in a \(3 \times 3\) grid world with the four preferences' combinations showed that agents with goal shaping and hard goals perform best, at the expense of reducing uncertainty about the ground-truth transition dynamics, which is to be expected as they are entirely guided by a collection of preference distributions that specify a distinct trajectory through various intermediate goals (maze positions) to the final goal (bottom-right tile).

In more complex environments, e.g., when the specification of a desired trajectory is unfeasible, agents will have to figure out by themselves the best way of solving a certain task or reaching a certain goal.
Not surprisingly, simulations of this kind of scenario using the same \(3 \times 3\) grid world showed that agents without goal shaping make more mistakes but eventually discover all the six possible trajectories to the goal.
This suggests that they might be more robust to perturbations, e.g., to one or more paths to the goal becoming obstructed, because they have spent more time learning about the environment's transition dynamics (see~\cref{fig:gridw9-kl-transitions}, and compare ground-truth transition maps in~\cref{sssecx:gridw9-gtruth-trans-maps} with learned ones in~\cref{sssecx:gridw9-learned-trans-maps}).

Without goal shaping, agents with soft goals turned out to be marginally more capable of reaching the final goal state, which can be explained by the fact that properly specified soft preferences (e.g., using the Manhattan distance from the final goal state) allow the agent to explore the environment while still having a (soft) preference for reaching a subset of states that are closer to the final goal state.

\subsection{The Risk of unfamiliar policies}\label{ssec:discus-two}
In describing how policy-conditioned free energies and expected free energies combine to determine policy probabilities at step 1, we noted that agents without goal shaping score task-failing policies higher than task-solving ones at some point, especially thanks to low risk values (see also~\cref{sssecx:gridw9-efe-breakdown}).
On the face of it, this seems puzzling because those policies are task-failing, i.e., they do not lead the agent to its goal state, and yet they are scored favourably in terms of risk.

However, since these policies haven't been sufficiently explored during the experiment, active inference agents have unresolved uncertainty as to where they lead, indicated by high-entropy variational distributions, \(\varprob{\statevar_{2}|\policy_{k}}, \dots, \varprob{\statevar_{\ntime}|\policy_{k}}\), for every \(k \in [1, \numpolicies]\), updated at the perceptual inference stage (in step 1 of each episode).
These variational beliefs produce large values of policy-conditioned free energies (see~\cref{fig:gridw9-policies-fe-step1}) and at the same time determine a drop of the policy's risks along the corresponding trajectory (see~\cref{fig:gridw9-efe-risk}).

To understand why this drop occurs, recall that risk is the KL divergence between two probability distributions, and that a low-entropy (peaked) distribution is “closer” to a high-entropy (flatter) distribution than to \emph{another} low-entropy distribution, i.e., a distribution with a peak somewhere else.
Thus, risk will be relatively low when computed between a low-entropy preference distribution, where probability mass is concentrated all on one goal or a few ones, and a high-entropy variational distribution, \(\varprob{\statevar_{t}|\policy_{k}}\), for a task-failing policy \(\policy_{k}\) that was rarely attempted (i.e., for which the agent is not certain about the resulting state trajectory). 
In contrast, risk will be relatively high between two different low-entropy distributions with different peaks, e.g., when a preference distribution with concentrated probability mass is compared with the variational distribution of a policy \(\policy_{k}\) that was executed more frequently (i.e., for which the agent is certain about the resulting state trajectory).

This situation does not occur in agents with hard goals and goal shaping because in this case we have a collection of preference distributions that traces a path through state-space and a unique policy that realises that trajectory, making the corresponding variational beliefs match perfectly the corresponding preference distributions.
Thus, the risk (KL divergence) between matching distribution will always be lower than that between a high-entropy and low-entropy distribution.
Because of this, the risk of the task-solving policy correctly turns out to be the lowest (see top-right plot in~\cref{fig:gridw9-efe-risk}).

Overall, when we consider agents without goal shaping (and, note, in an environment with deterministic state transitions), increasing the entropy of the preference distribution, i.e., using soft goals, can be seen as an effective method to limit this (incorrect) assignment of instrumental value to task-failing policies (see again~\cref{fig:gridw9-first-pi-probs}).

The phenomenon just described is somewhat mitigated by the fact that high-entropy variational beliefs for the trajectory of a policy make the corresponding policy-conditioned free energy soar, thereby penalising the policy when that quantity is combined with expected free energy to yield the policy probability via the softmax (see~\cref{sec:results}).
However, this does not prevent some task-failing policies from becoming more probable than task-solving ones (see again~\cref{fig:gridw9-first-pi-probs}).
As remarked in~\cref{ssec:action-selection}, these agents do not fail completely only because the action selection procedure does not involve picking the action from the most probable policy.

At the same time, we note that even if that were the case, the agents would learn the consequences of those task-failing policies, resolving their uncertainty about the transition dynamics of rarely tried actions' sequences (and for some agents this is what occurs, as indicated by the probabilities for these task-failing policies declining from episode 160 onwards, see especially the bottom-right plot in~\cref{fig:gridw9-first-pi-probs}).
In other words, in addition to the two novelty terms of the expected free energy, there is an implicit epistemic drive arising out of the risk term as well, that compels agents to execute less tried actions' sequences, when agents are trained without goal shaping.

\subsection{Related work}\label{ssec:discus-three}
One of the main distinctive features of active inference is the idea of integrating state estimation, planning, decision-making, and learning under the same objective, i.e., the minimisation of some kind of free energy, without the need to introduce \emph{ad-hoc} exploratory bonuses (as done in some reinforcement learning approaches, see~\parencite{Levine2018a}) to deal with the exploration-exploitation dilemma (but see~\parencite{Millidge2020d}) and without relying on notions of reward and value functions (which are foundational in reinforcement learning, see~\parencite{Sutton2018a}).

However, it does not follow that goal-directedness emerges from free-energy minimisation \emph{tout court}~\cite{Seth2015b, Baltieri2020b}.
The design of an active inference agent requires the specification of a preference distribution that creates a relevant learning signal for the agent, via expected free energy, so that some kind of adaptive behaviour is elicited in a certain environment.
The specification of this preference distribution, that indicates what states or observations are preferred by an active inference agent, effectively corresponds to the design of a reward function in reinforcement learning.
Because of this, similar theoretical considerations can be made, e.g., related to the origin or nature of this preference distribution, \emph{cf}.~\parencite{Juechems2019a}.
Indeed, the distinction between soft vs.\ hard goals corresponds to that between dense vs.\ sparse reward signals, and goal shaping can be understood similarly as reward shaping~\parencite{Randlov1998a,Eschmann2021a}.

In this respect, active inference and reinforcement learning appear to have more in common than what is often claimed~\cite{Friston2009b, Friston2011b, Sajid2019a}.
After all, the specification of a preference distribution is a way to let the agent knows what states or observations should be deemed to be rewarding, hence to be preferred.
This is not just a terminological issue, as in scaled-up applications of active inference it is common to use the reward from environments used in reinforcement learning as part of the observations the agent receives, thereby marking explicitly some observations as those to be preferred because associated with higher reward \parencite{Tschantz2019a,Tschantz2020c,Millidge2020e,VanDerHimst2020a}.
Furthermore, recent theoretical work has formally showed that minimising expected free energy implies reward maximisation in finite-horizon MDPs and POMDPs, with \emph{sophisticated inference}, a recursive kind of planning based on expected free energy~\parencite{DaCosta2023a}.
This work clarifies how goals can be interpreted as reward maximising states, by formally defining the preference distribution as a Boltzmann distribution with the reward assigned to a state as the random variable (see~\parencite[818]{DaCosta2023a} for the details).

Ultimately, agents need a learning signal that guides them towards achieving what they are designed to achieve, and whether we call it reward or preferences (for an agent) might matter less than \emph{how} that learning signal is exploited in clever ways.
In active inference, that learning signal arises out of variational free energy and expected free energy.
In particular, the latter would combine instrumental or extrinsic drives together with epistemic or intrinsic drives.
More generally, active inference invites us to take an agent-centric perspective revolving around the notion of free energy minimisation: the type of agent we are considering, its available actions and how it can interact with a certain environment, can inform us of the goal-directed behaviour to be expected, or provide some guidance on how that behaviour can be designed in a transparent way via the specification of the preference distribution.

\subsection{Limitations and future work}
The in-depth analysis of the impact of prior preferences in active inference agents was limited, in this work, to the case of preferences defined over states, in agents trained in a low-dimensional grid world.
The states in question can be considered as internal to the agent, i.e., indicating its current location in the environment in our experiment.
However, as remarked, preference distributions can also be defined over observations received from the environment and reward functions from reinforcement learning environments can be incorporated into them, i.e., reward is treated as an observable part of the environment (see~\parencite{Mazzaglia2022a} for a nice summary of the possibilities).
A study of the trade-offs and peculiarities of these alternative approaches, often used in implementations that rely on deep learning to train active inference agents in high-dimensional environments, must be left for future work.
In a similar vein, the question of how agent's preferences could be progressively learned in different environments, e.g., by introducing a Dirichlet prior for the parameters of the categorical preference distribution \parencite{Sajid2021a}, was not taken into consideration here and is an interesting avenue for future research.

\section{Concluding Remarks}\label{sec:ccnclusion}

In the present work, we investigated how different ways of fixing the preference distribution in action-unaware active inference agents affects their performance in a toy environment.
In particular, we focussed on the distinction between training agents with or without goal shaping and with soft of hard preference.
Goal shaping refers to the idea of using a collection of preference distributions to direct the agent towards pursuing a particular trajectory in state-space (e.g., a trajectory of intermediate goals to reach a main goal); without it, the agent needs to find by trial and error which trajectories converge onto its main goals, encoded by a single preference distribution.
The second dimension, whether to encode soft or hard goals into the preference distribution, refers to how much the probability mass is concentrated on a subset of states/observations.

\begin{figure}[H]
  \centering
  \begin{subfigure}{0.45\textwidth}\label{fig:gridw9-aif-paths-softgs-matrix-B-kl}
    \centering
    \includegraphics[width=\textwidth]{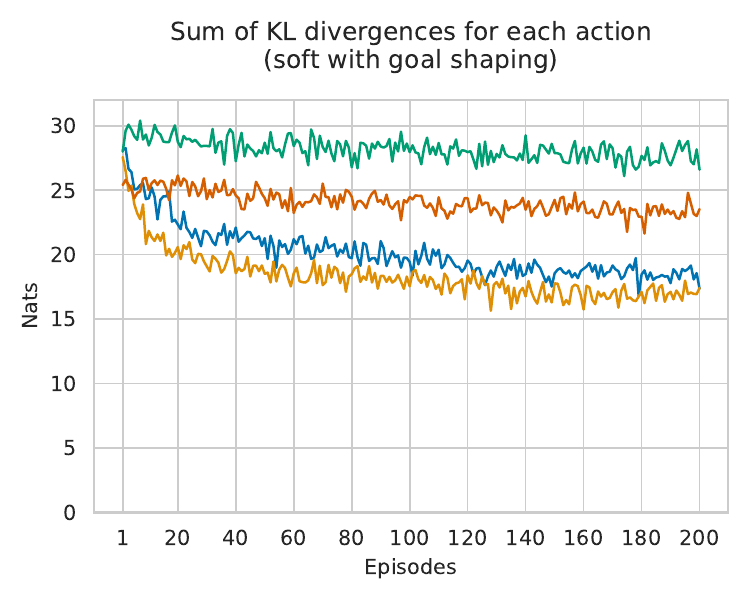}
  \end{subfigure}
  \begin{subfigure}{0.45\textwidth}\label{fig:gridw9-aif-paths-hardgs-matrix-B-kl}
    \centering
    \includegraphics[width=\textwidth]{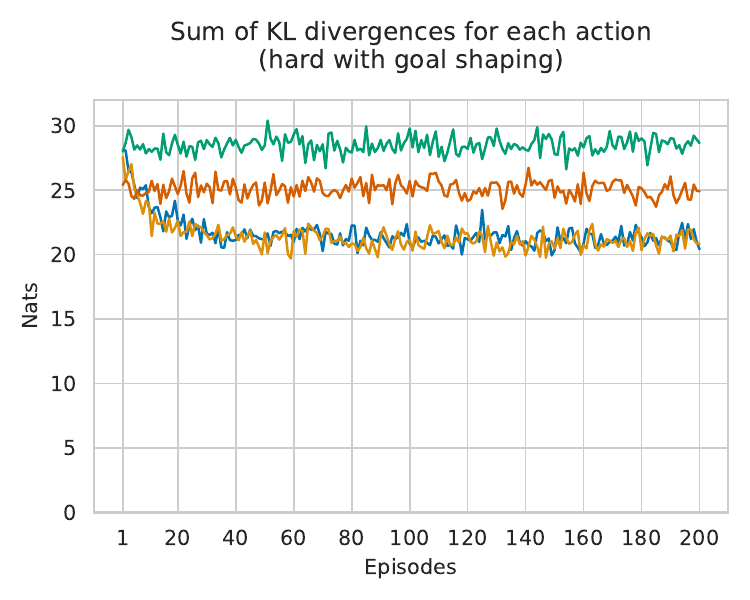}
  \end{subfigure}
  \centering
  \begin{subfigure}{0.45\textwidth}\label{fig:gridw9-aif-paths-soft-matrix-B-kl}
    \centering
    \includegraphics[width=\textwidth]{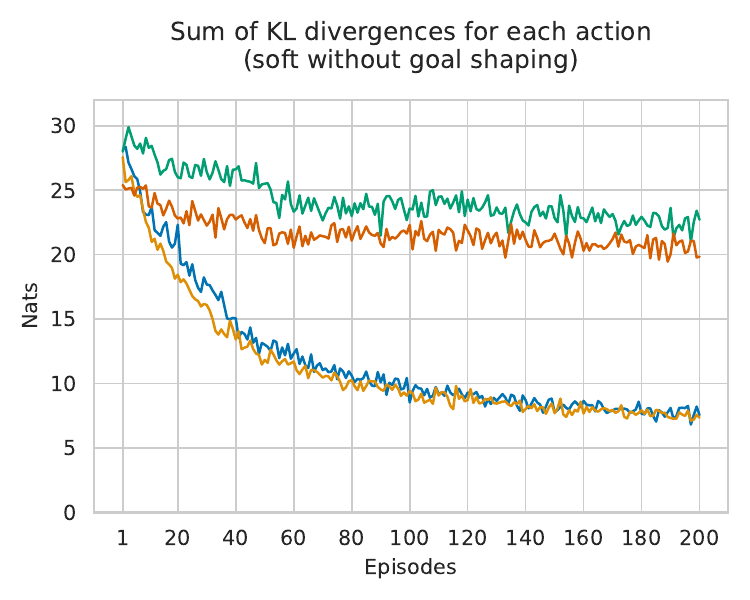}
  \end{subfigure}
  \begin{subfigure}{0.45\textwidth}\label{fig:gridw9-aif-paths-hard-matrix-B-kl}
    \centering
    \includegraphics[width=\textwidth]{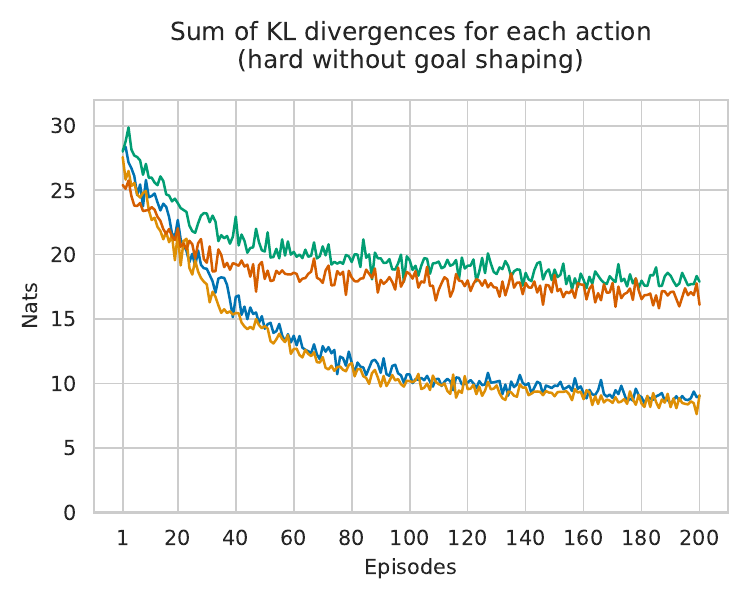}
  \end{subfigure}
  \begin{subfigure}{0.65\textwidth}
    \centering
    \includegraphics[width=\textwidth]{gridw9_actions_legend}
  \end{subfigure}
  \caption{Sums of KL divergences between learned and ground-truth transition probabilities (columns of \(\transmap\)-matrices), i.e., \(\sum_{\state_{t} \in \statespace}\kldiv [\tprob{\statevar_{t}} | P^{*}(\statevar_{t}| \state_{t}, \action_{t})]\), for each available action \(\action_{t}\) (showing average of 10 agents). The decrease of this total KL divergence for one action across episodes implies that the agent has learned the consequences of performing that action from multiple states in the environment.}\label{fig:gridw9-kl-transitions}
\end{figure}

In general, agents with goal-shaping perform better because they are provided with preference distributions that together indicate a sequence of preferred states, facilitating the selection of the single policy that will lead the agent through the corresponding path in state-space.
In contrast, agents without goal-shaping have to find that preferred path, relying only on a single preference distribution that indicates the ultimate state or goal they are supposed to achieve (in our case: being located in tile 9).
As a result, their performance suffers but at the same time they are more free to explore the environment, therefore they are able to find all the possible paths to the ultimate goal.
Thus, we can conclude that the absence of goal-shaping determines a more widespread learning of the environment's transition dynamics, i.e., of the transition probabilities for each available action (see again~\cref{fig:gridw9-kl-transitions} and learned transition maps in~\cref{sssecx:gridw9-learned-trans-maps}).

\phantomsection
\addcontentsline{toc}{section}{Acknowledgments} 
\section*{Acknowledgments}
This work was supported by JST, Moonshot R\&D, Grant Number JPMJMS2012.

\printbibliography%
\clearpage
\beginsupplement%

\section{Mathematical background}\label{secx:math-background}

\subsection{Notation}\label{ssecx-notation}

\renewcommand{\arraystretch}{1.0}
\begin{table}[H]\label{tab:summary-notation}
    \caption{Summary of notation.}
    \centering
    \resizebox{\textwidth}{!}{%
    \begin{tabular}{lc}
    \toprule
    Symbol & Meaning\\
    \midrule
    \(t, \tau, \ntime\) & integers, i.e., generic, current, and terminal time index, respectively\\
    \(1:t\), \(1:\ntime\) & sequences of times steps up to \(t\) and \(\ntime\), respectively\\
    \(\polhorizon\) & integer, length of a sequence of actions (i.e., the policy horizon), in general \(H \leq T\) \\
    \(\numpolicies\) & integer, the number of action sequences (policies) the agent considers \\
    \(X_{t}\) & random variable with support in \(\mathcal{X}\), and with \(t \in [1, \ntime]\) \\
    \(\seqv{X}{1}{T}, \seqv{x}{1}{T}\) & sequence of random variables with time index and related values \\
    \(\mathbf{X}_{:,j}\) & \(j\)th column of matrix \(\mathbf{X}\) or random vector associated with that column \\
    \(\prob{X_{t}}, \prob{x_{t}}\) & probability distribution of random variable \(X_{t}\) and probability that \(X_{t} = x_{t}\) (when defined) \\
    \(\entropy[X_{t}]\) & Shannon entropy of random variable \(X_{t}\) \\
    \(\statespace\) & finite set of cardinality \(\card{\statespace}\), i.e., the set of states \\
    \(\obsspace\) & finite set of cardinality \(\card{\obsspace}\), i.e., the set of observations \\
    \(\actionspace\) & finite set of cardinality \(\card{\actionspace}\), i.e., the set of actions \\
    \(\actionspace^{\polhorizon}\) & finite set of action tuples (\(\polhorizon\)-fold Cartesian product) \\
    \(\Pi\) & subset of action sequences, i.e.,  \(\Pi \subseteq\actionspace^{\polhorizon}\) \\
    \(\statevar_{t}, \obsvar_{t}, \actionvar_{t}, \policy\) & categorical random variables with support in \(\statespace, \obsspace, \actionspace, \policyspace\), respectively, i.e., \(\statevar_{t} \sim \text{Cat}(\mathbf{s}_{t}), \dots\) \\
    \(\state, \obs, \action_{t}, \policy_{k}\)  & elements in \(\statespace, \obsspace, \actionspace, \policyspace\), respectively, where \(k \in [1, \numpolicies]\) and \(\numpolicies \in [1, \card{\policyspace}]\) \\
    \(\state^{j}, \obs^{j}\) & particular realisations of random variables \(\statevar_{t}, \obsvar_{t}\), with \(j \in [1, \card{\statespace}]\) and \(j \in [1, \card{\obsspace}]\), respectively\\
    \(\stateparams_{t}, \obsparams_{t}, \policyparams\), & column vectors of parameters for state, observation, and policy random variables, respectively \\
    \(\stateparams_{t}[i], \obsparams_{t}[i], \policyparams[i]\) & \(i\)th element of the parameter vector for state, observation, and policy random variables, respectively \\
    \(\Transition\) & transition map/function \\
    \(\Emission\) & emission map/function \\
    \(\tprob{\state_{t}}\) & transition probability distribution (returned by \(\Transition\)) \\
    \(\eprob{\obs_{t}}\) & emission probability distribution (returned by \(\Emission\)) \\
    \(\pprob{\state_{t}}, \pprob{\obs_{t}}\) & stationary distributions over \(\statespace\) and \(\obsspace\), respectively \\
    \(\genmodel\) & generative model (collection of probability distributions) \\
    \(\obsmap\) & matrix in \(\mathbb{R}^{n \times m}\) storing parameters of \(P(\obsvar_{t} \vert \statevar_{t-1})\) (the same for any \(t\)) \\
    \(\transmap\) & tensor in \(\mathbb{R}^{\card{\actionspace} \times m \times m}\) storing parameters of \(P(\statevar_{t} \vert \statevar_{t-1})\) (the same for any \(t\)) \\
    \(\transmap^{a_{1}}, \dots, \transmap^{a_\mathtt{d}}\) & state-transition matrices in \(\mathbb{R}^{m \times m}\) for each available action, \(d = \card{\actionspace}\) \\
    \(\fe\) & free energy \\
    \(\fe_{a_{\tau-1}}\) & action-conditioned free energy in vanilla active inference \\
    \(\fe_{\policy_{k}}\) & policy-conditioned free energy in variational message passing \\
    \(\efe_{t}\) & single-step expected free energy \\
    \(\totefe\) & total expected free energy, i.e., sum of expected free energies for \(H\) time steps in the future \\
    \(\nabla_{\stateparams_{t}} \fe_{\policy_{k}}\) & gradient of policy-conditioned free energy with respect to vector of parameters \(\stateparams_{t}\) \\
    \(\nabla_{\policyparams} \fe\) & gradient of free energy with respect to vector of policy parameters \(\policyparams\) \\
    \(\boldsymbol{\fe}_{\policy}^{\intercal}\) & row-vector in \(\mathbb{R}^{1 \times \card{\policyspace}}\) of policy-conditioned free energies \\
    \(\boldsymbol{\totefe}^{\intercal}\) & row-vector in \(\mathbb{R}^{1 \times \card{\policyspace}}\) of total expected free energy \\
    \bottomrule
    \end{tabular}}
\end{table}

\section{Further information on experiments}\label{secx:further-info-exps}

\subsection{How to Reproduce the Results of the Experiment}\label{ssecx:how-reproduce-results}
The results reported in~\cref{sec:results} were obtained by using the active inference implementation available at~\url{https://github.com/FilConscious/cleanAIF} and with the following command line instructions.

Soft preferences with goal-shaping:

\begin{lstlisting}[breaklines=true,]
main_aif_au --exp_name aif_au_softgs --gym_id gridworld-v1 --env_layout gridw9 --num_runs 10 --num_episodes 200 --num_steps 5 --inf_steps 10 --action_selection kd -lB --num_policies 256 --pref_type states_manh --pref_loc all_diff
\end{lstlisting}

Hard preferences with goal-shaping:

\begin{lstlisting}[breaklines=true,]
main_aif_au --exp_name aif_au_hardgs --gym_id gridworld-v1 --env_layout gridw9 --num_runs 10 --num_episodes 200 --num_steps 5 --inf_steps 10 --action_selection kd -lB --num_policies 256 --pref_type states --pref_loc all_diff
\end{lstlisting}

Soft preferences without goal-shaping:

\begin{lstlisting}[breaklines=true,]
main_aif_au --exp_name aif_au_soft --gym_id gridworld-v1 --env_layout gridw9 --num_runs 10 --num_episodes 200 --num_steps 5 --inf_steps 10 --action_selection kd -lB --num_policies 256 --pref_type states_manh --pref_loc all_goal
\end{lstlisting}

Hard preferences without goal-shaping:

\begin{lstlisting}[breaklines=true,]
main_aif_au --exp_name aif_au_hard --gym_id gridworld-v1 --env_layout gridw9 --num_runs 10 --num_episodes 200 --num_steps 5 --inf_steps 10 --action_selection kd -lB --num_policies 256 --pref_type states --pref_loc all_goal
\end{lstlisting}

The plots were obtained using the following command line instructions:

\begin{lstlisting}[breaklines=true,]
vis_aif -gid gridworld-v1 -el gridw9 -nexp 1 -rdir episodic_e200_pol16_maxinf10_learnB_cr_Bparams_softgs -fpi 0 1 2 3 4 -i 4 -v 8 -ti 4 -tv 8 -vl 3 -hl 3 -xtes 20 -ph 4 -selrun 0 -npv 16 -sb 4 -ab 0 1 2 3
\end{lstlisting}

With these instructions, one can visualize more metrics than those reported in the main text.
We offer a selection next.

\subsection{Additional figures}\label{ssecx:additional-figures}



\subsubsection{State-acces frequency}\label{sssecx:gridw9-sv}

\subsubsection{Breakdown of expected free energy at step 1}\label{sssecx:gridw9-efe-breakdown}

\begin{figure}[H]
  \centering
  \begin{subfigure}{0.45\textwidth}
    \centering
    \includegraphics[width=\textwidth]{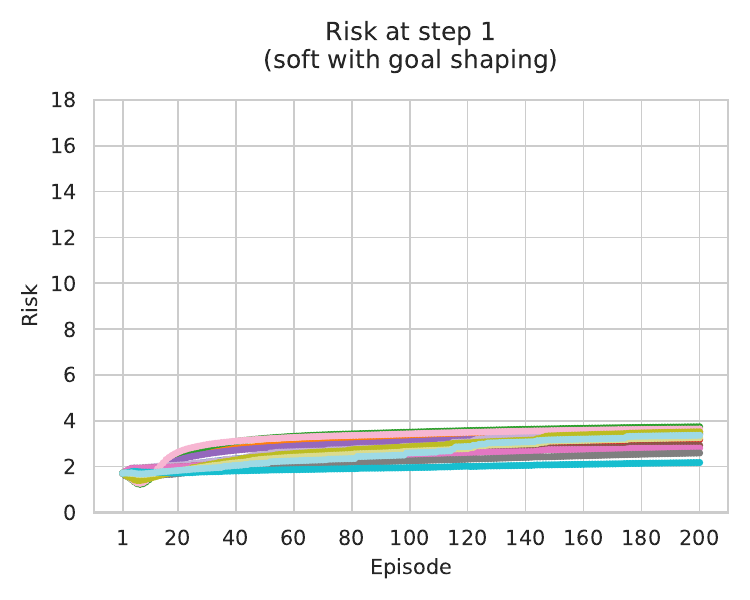}
  \end{subfigure}
  \begin{subfigure}{0.45\textwidth}
    \centering
    \includegraphics[width=\textwidth]{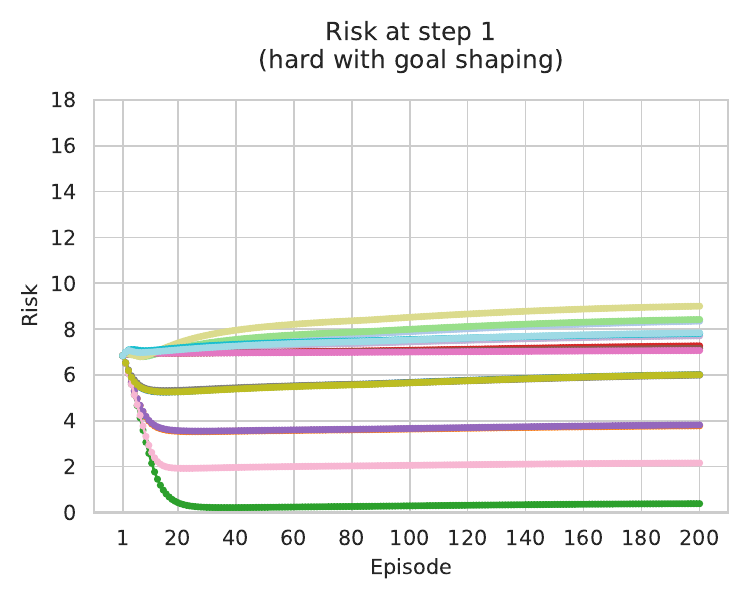}
  \end{subfigure}
  \begin{subfigure}{0.45\textwidth}
    \centering
    \includegraphics[width=\textwidth]{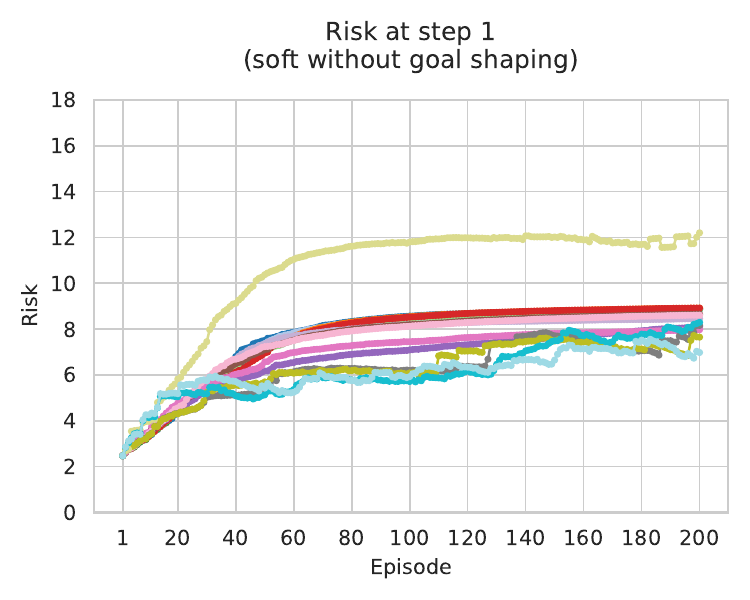}
  \end{subfigure}
  \begin{subfigure}{0.45\textwidth}
    \centering
    \includegraphics[width=\textwidth]{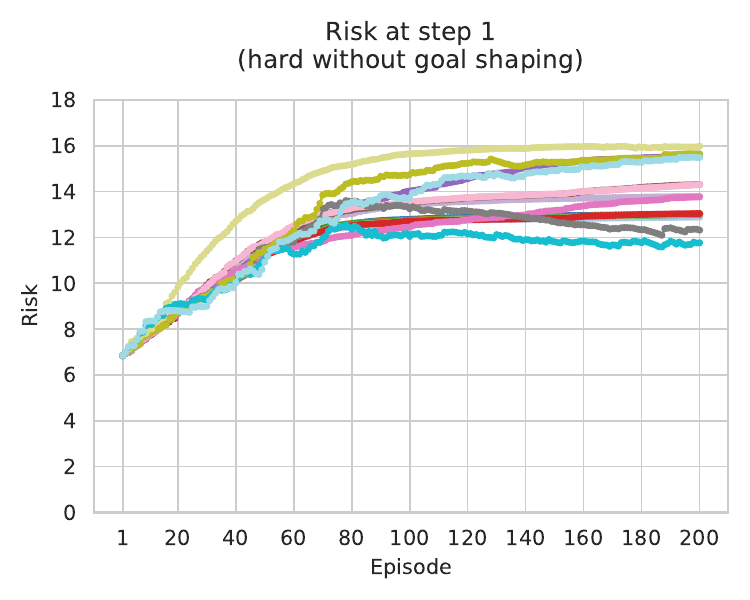}
  \end{subfigure}
  \begin{subfigure}{0.65\textwidth}
    \centering
    \includegraphics[width=\textwidth]{gridw9_aif_policies_legend}
  \end{subfigure}
  \caption{Risk (expected free energy term) for each policy across episodes (showing average of 10 agents).}\label{fig:gridw9-efe-risk}
\end{figure}
\begin{figure}[H]
  \centering
  \begin{subfigure}{0.45\textwidth}
    \centering
    \includegraphics[width=\textwidth]{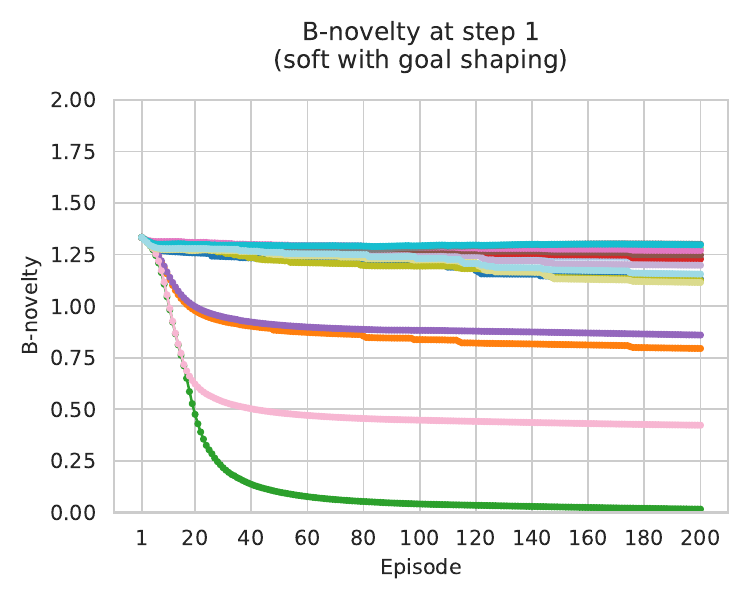}
  \end{subfigure}
  \begin{subfigure}{0.45\textwidth}
    \centering
    \includegraphics[width=\textwidth]{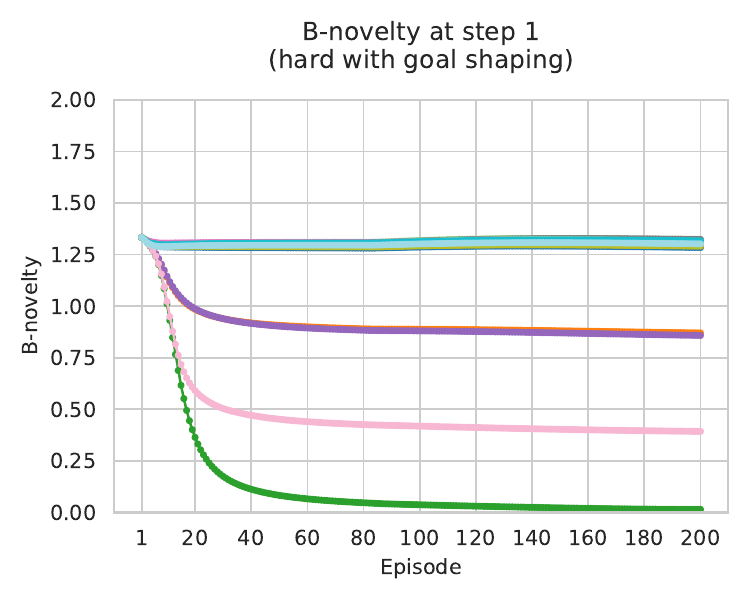}
  \end{subfigure}
  \begin{subfigure}{0.45\textwidth}
    \centering
    \includegraphics[width=\textwidth]{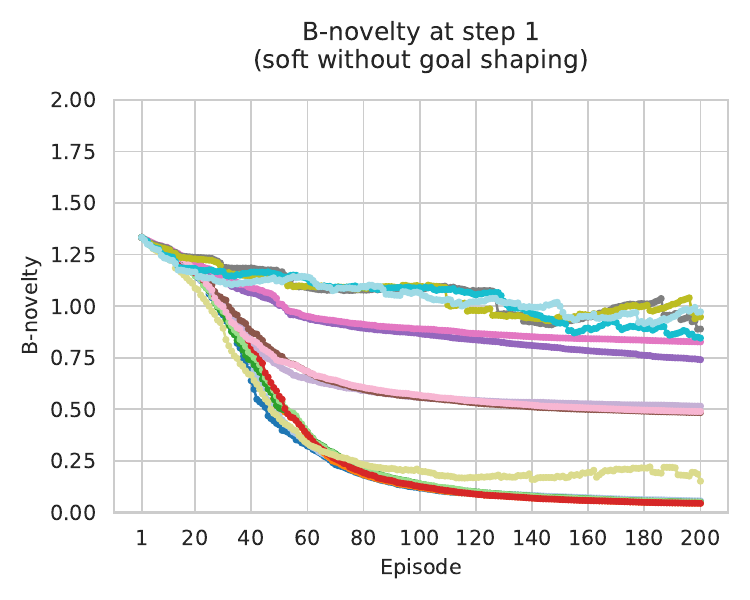}
  \end{subfigure}
  \begin{subfigure}{0.45\textwidth}
    \centering
    \includegraphics[width=\textwidth]{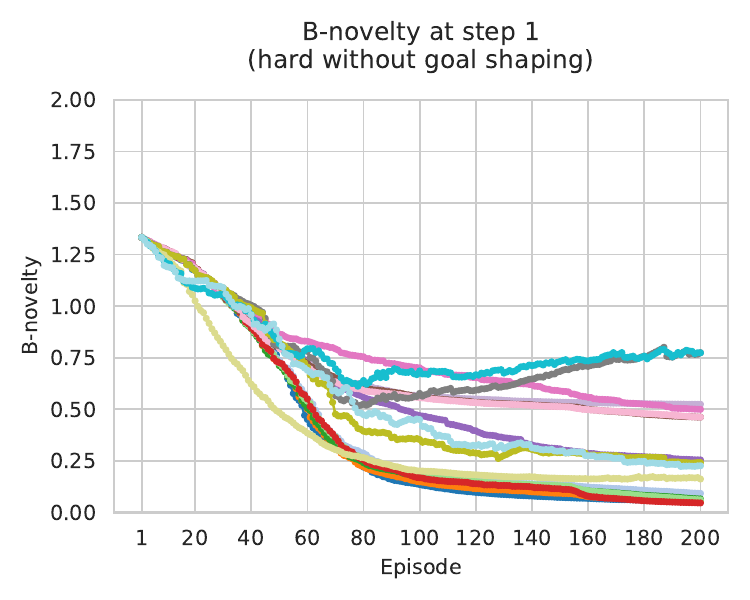}
  \end{subfigure}
  \begin{subfigure}{0.65\textwidth}
    \centering
    \includegraphics[width=\textwidth]{gridw9_aif_policies_legend}
  \end{subfigure}
  \caption{\(\transmap\)-novelty (expected free energy term) for each policy across episodes (showing average of 10 agents).}\label{fig:gridw9-efe-bnov}
\end{figure}

Risk is the largest component in expected free energy to the point of determining its trend (compare the expected free energy and risk figures,~\cref{fig:gridw9-efe-step0} and~\cref{fig:gridw9-efe-risk}, respectively, and see~\cref{fig:gridw9-efe-bnov} for \(\transmap\)-novelty).
In agents with soft goals and goal-shaping, all policies converge on risk values between 2 and 4, with no clear distinction betweem task-solving and task-failing policies (see top-left plot in~\cref{fig:gridw9-efe-risk}).
This is the case because in an environment characterised by deterministic transitions (and without any kind of teleportation), there is no policy that can attain a probability distribution over environment's states that matches the agent's prior (soft) preferences at each time step, thereby minimizing risk.
In contrast, in agents with hard goals, the policy minimizing risk is \(\policy_{4}\) because it brings the agent to visit all the intermediate steps (goals), forming the trajectory to the main goal (see top-right plot in~\cref{fig:gridw9-efe-risk}).
In agents without goal shaping, risk evolves similarly: it increases in the first 60 or so episodes before converging to a stationary value (see bottom-left and bottom-right plots in~\cref{fig:gridw9-efe-risk}).
All task-solving policies achieve the same risk value, but it is not the lowest one as a few task-failing policies appear to minimize expected free energy even further.
This is again due to how preferences are encoded and on how risk is computed (see~\cref{ssec:discus-two}).

As to \(\transmap\)-novelty, its evolution is more in line with expectations: the more a policy has been tried and executed, the more its \(\transmap\)-novelty will decrease.
In agents with goal shaping, that is precisely what happens to the preferred \(\policy_{4}\) (see top-left and top-right plots in~\cref{fig:gridw9-efe-bnov}).
In agents without goal shaping, the same reduction in \(\transmap\)-novelty occurs for all the task-solving policies as well as few task-failing (see bottom-left and bottom-right plots in~\cref{fig:gridw9-efe-bnov}).

\subsubsection{Average action probabilities at steps 2--4}\label{sssecx:avg-action-probs}

\begin{figure}[H]
  \centering
  \begin{subfigure}{0.45\textwidth}\label{fig:gridw9-aif-au-softgs-s2-action-probs-avg}
    \centering
    \includegraphics[width=\textwidth]{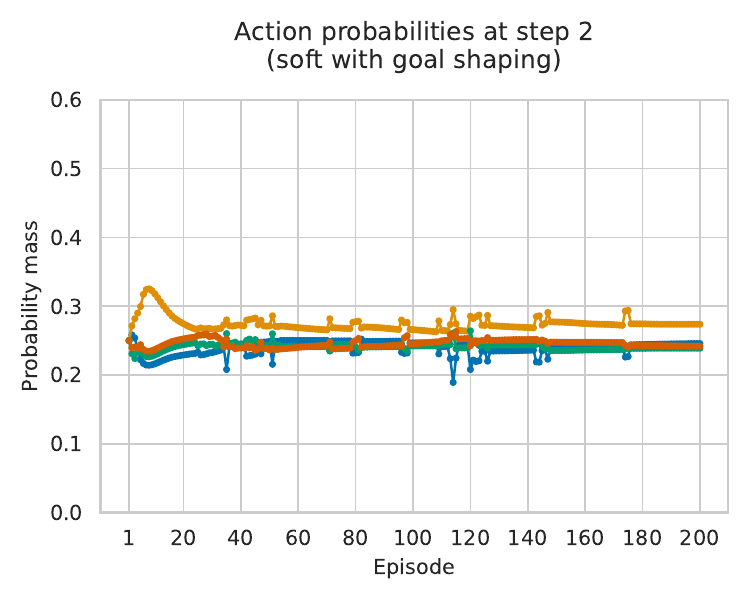}
  \end{subfigure}
  \begin{subfigure}{0.45\textwidth}\label{fig:gridw9-aif-au-hardgs-s2-action-probs-avg}
    \centering
    \includegraphics[width=\textwidth]{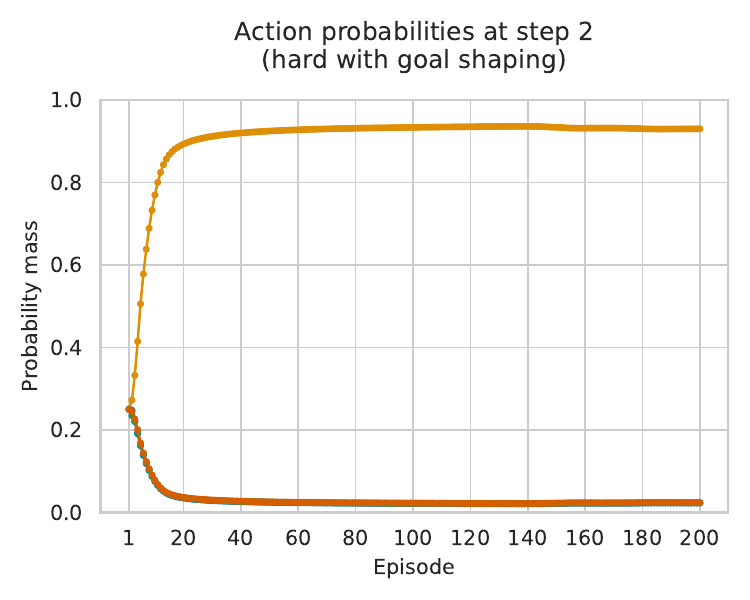}
  \end{subfigure}
  \centering
  \begin{subfigure}{0.45\textwidth}\label{fig:gridw9-aif-au-soft-s2-action-probs-avg}
    \centering
    \includegraphics[width=\textwidth]{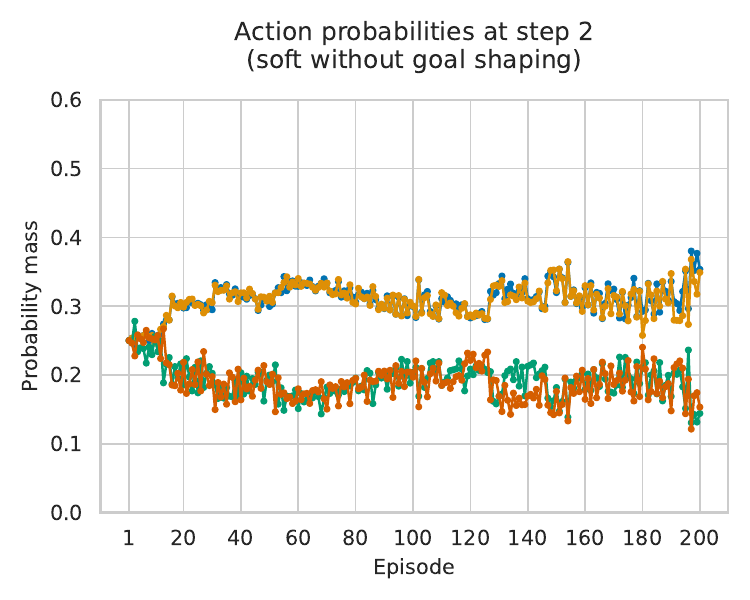}
  \end{subfigure}
  \begin{subfigure}{0.45\textwidth}\label{fig:gridw9-aif-au-hard-s2-action-probs-avg}
    \centering
    \includegraphics[width=\textwidth]{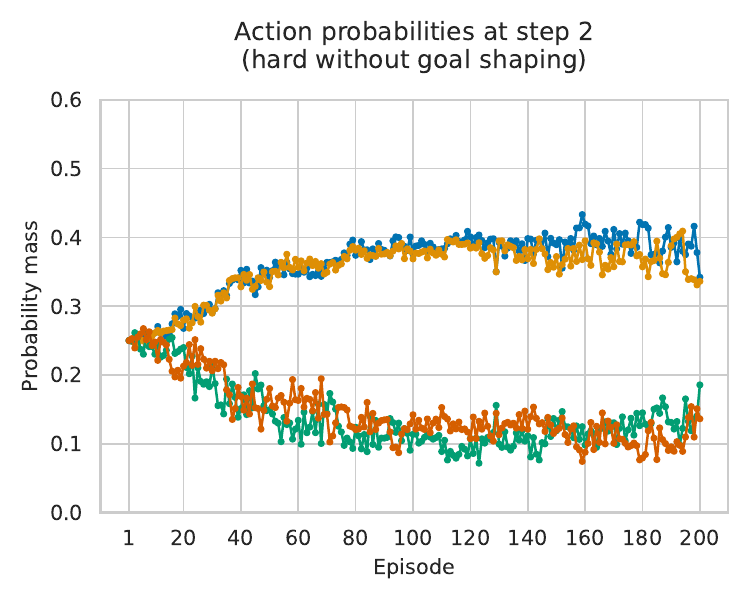}
  \end{subfigure}
  \begin{subfigure}{0.65\textwidth}
    \centering
    \includegraphics[width=\textwidth]{gridw9_actions_legend}
  \end{subfigure}
  \caption{Action probabilities at step 2 of each episode (showing average of 10 agents).}\label{fig:gridw9-s2-action-probs-avg}
\end{figure}

\begin{figure}[H]
  \centering
  \begin{subfigure}{0.45\textwidth}\label{fig:gridw9-aif-au-softgs-s3-action-probs-avg}
    \centering
    \includegraphics[width=\textwidth]{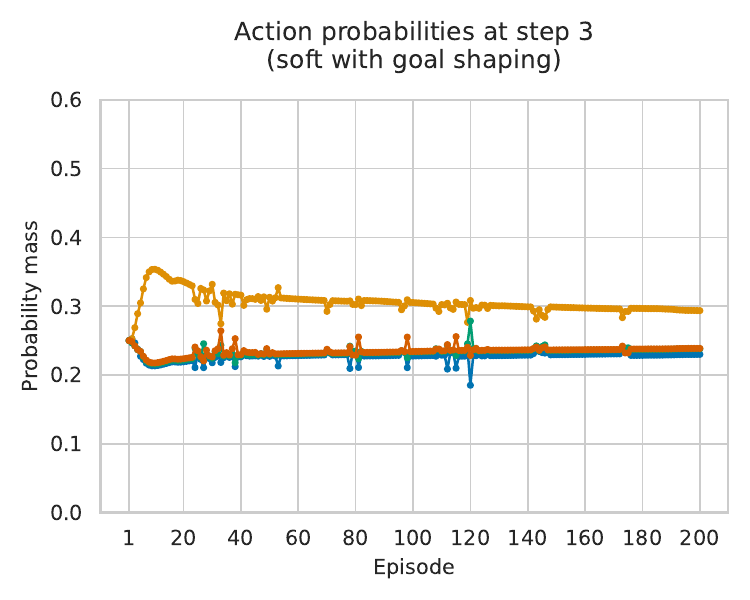}
  \end{subfigure}
  \begin{subfigure}{0.45\textwidth}\label{fig:gridw9-aif-au-hardgs-s3-action-probs-avg}
    \centering
    \includegraphics[width=\textwidth]{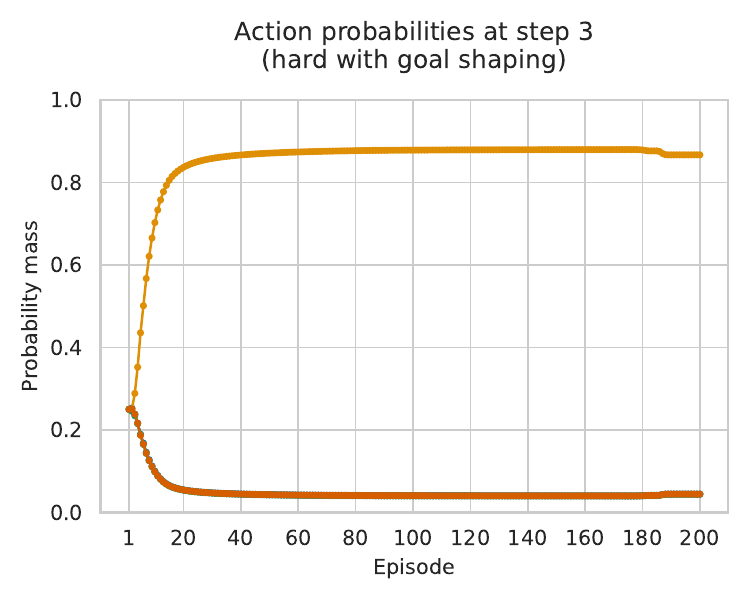}
  \end{subfigure}
  \centering
  \begin{subfigure}{0.45\textwidth}\label{fig:gridw9-aif-au-soft-s3-action-probs-avg}
    \centering
    \includegraphics[width=\textwidth]{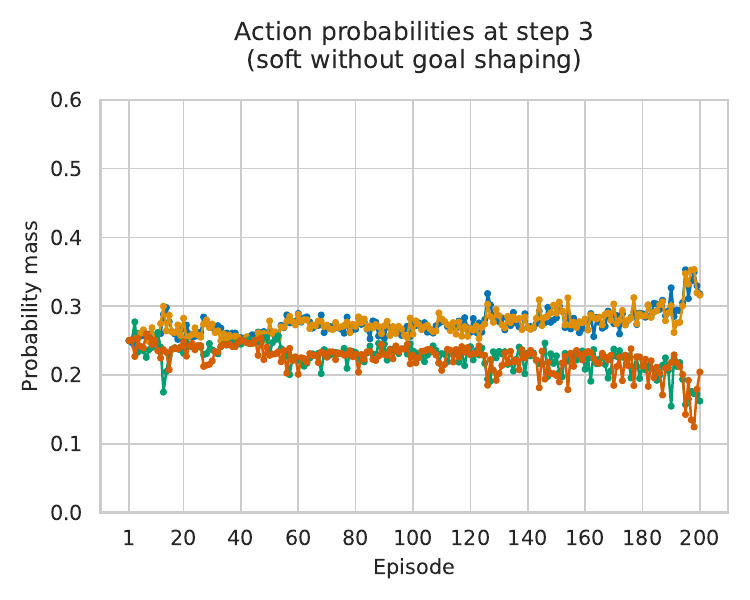}
  \end{subfigure}
  \begin{subfigure}{0.45\textwidth}\label{fig:gridw9-aif-au-hard-s3-action-probs-avg}
    \centering
    \includegraphics[width=\textwidth]{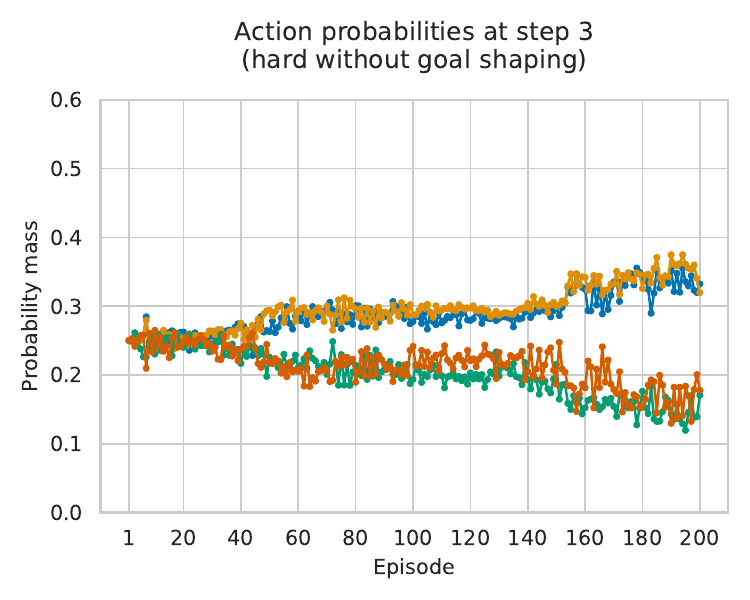}
  \end{subfigure}
  \begin{subfigure}{0.65\textwidth}
    \centering
    \includegraphics[width=\textwidth]{gridw9_actions_legend}
  \end{subfigure}
  \caption{Action probabilities at step 3 of each episode (showing average of 10 agents).}\label{fig:gridw9-s3-action-probs-avg}
\end{figure}

\begin{figure}[H]
  \centering
  \begin{subfigure}{0.45\textwidth}\label{fig:gridw9-aif-au-softgs-s4-action-probs-avg}
    \centering
    \includegraphics[width=\textwidth]{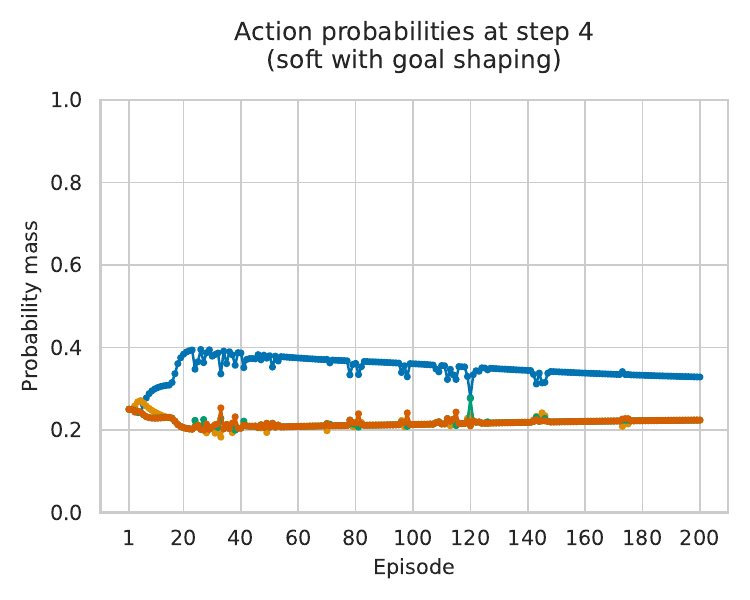}
  \end{subfigure}
  \begin{subfigure}{0.45\textwidth}\label{fig:gridw9-aif-au-hardgs-s4-action-probs-avg}
    \centering
    \includegraphics[width=\textwidth]{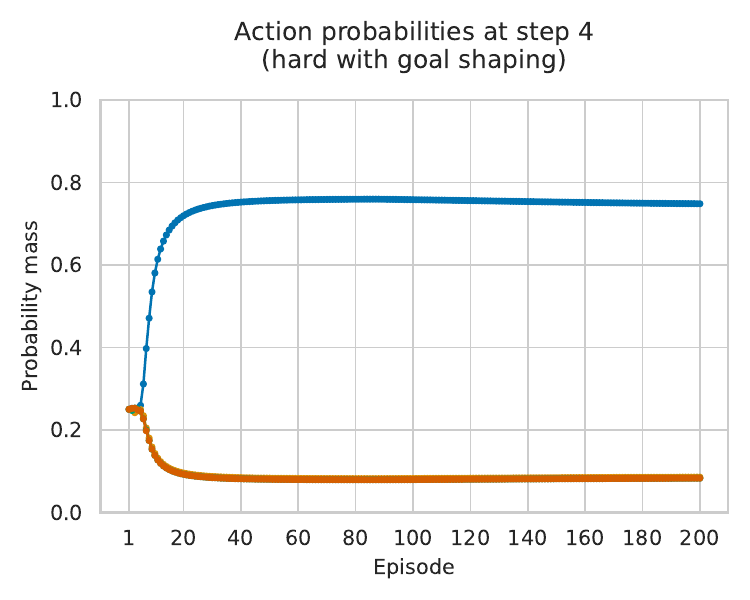}
  \end{subfigure}
  \centering
  \begin{subfigure}{0.45\textwidth}\label{fig:gridw9-aif-au-soft-s4-action-probs-avg}
    \centering
    \includegraphics[width=\textwidth]{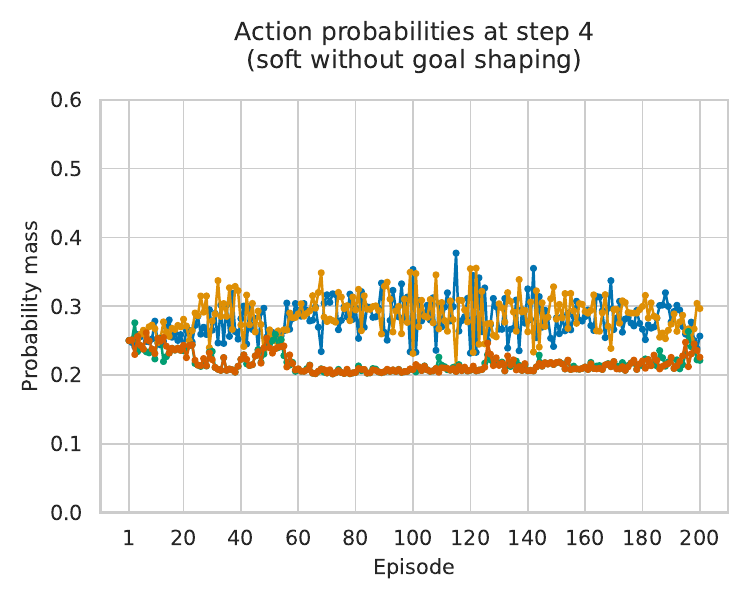}
  \end{subfigure}
  \begin{subfigure}{0.45\textwidth}\label{fig:gridw9-aif-au-hard-s4-action-probs-avg}
    \centering
    \includegraphics[width=\textwidth]{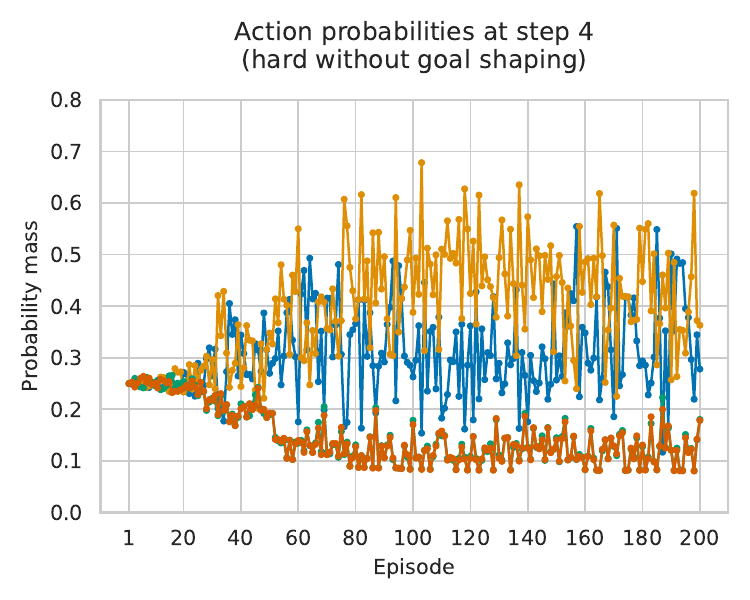}
  \end{subfigure}
  \begin{subfigure}{0.65\textwidth}
    \centering
    \includegraphics[width=\textwidth]{gridw9_actions_legend}
  \end{subfigure}
  \caption{Action probabilities at step 4 of each episode (showing average of 10 agent).}\label{fig:gridw9-s4-action-probs-avg}
\end{figure}

\subsubsection{Single-agent actions probabilities at step 1--4}\label{sssecx:single-action-probs}

\begin{figure}[H]
  \centering
  \begin{subfigure}{0.45\textwidth}\label{fig:gridw9-aif-paths-softgs-s3-action-probs}
    \centering
    \includegraphics[width=\textwidth]{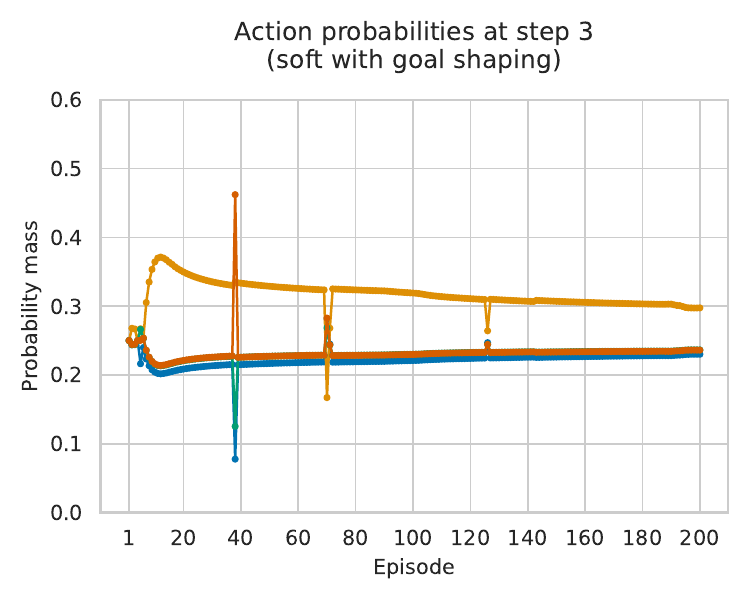}
  \end{subfigure}
  \begin{subfigure}{0.45\textwidth}\label{fig:gridw9-aif-paths-hardgs-s3-action-probs}
    \centering
    \includegraphics[width=\textwidth]{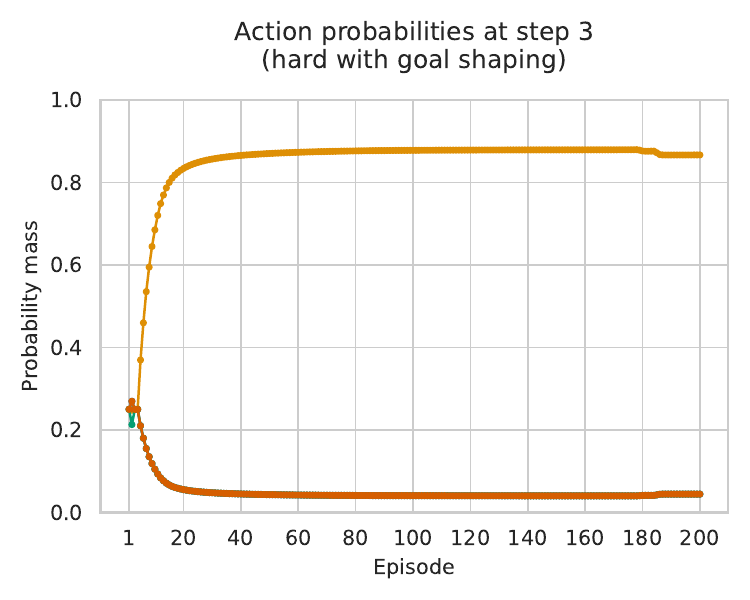}
  \end{subfigure}
  \centering
  \begin{subfigure}{0.45\textwidth}\label{fig:gridw9-aif-paths-soft-s3-action-probs}
    \centering
    \includegraphics[width=\textwidth]{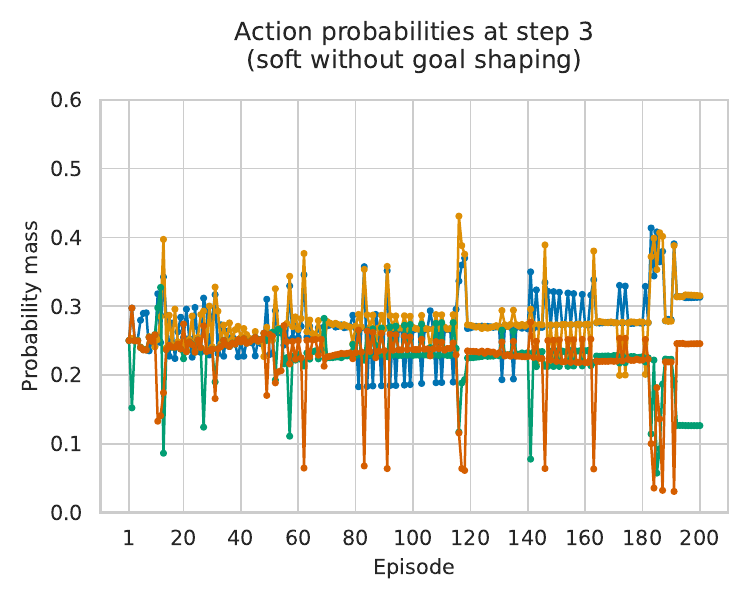}
  \end{subfigure}
  \begin{subfigure}{0.45\textwidth}\label{fig:gridw9-aif-paths-hard-s3-action-probs}
    \centering
    \includegraphics[width=\textwidth]{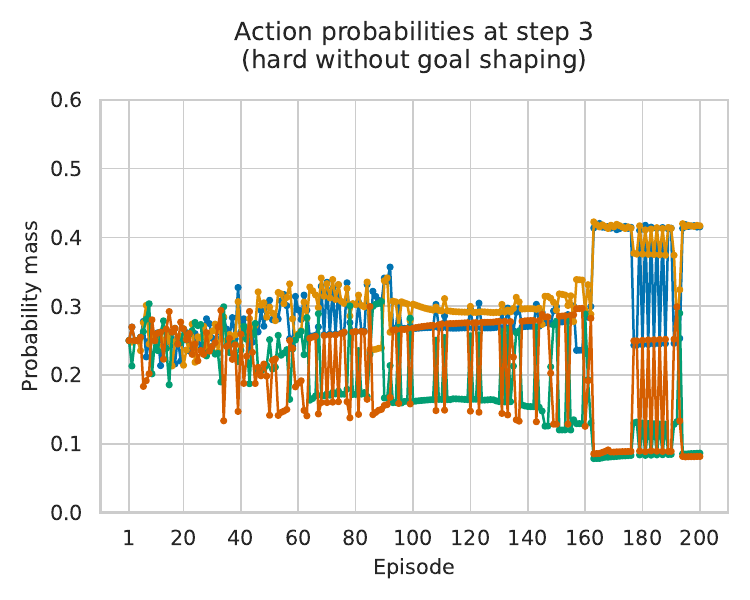}
  \end{subfigure}
  \begin{subfigure}{0.65\textwidth}
    \centering
    \includegraphics[width=\textwidth]{gridw9_actions_legend}
  \end{subfigure}
  \caption{Action probabilities at step 3 of each episode (showing results for one agent).}\label{fig:gridw9-s3-action-probs}
\end{figure}

\begin{figure}[H]
  \centering
  \begin{subfigure}{0.45\textwidth}\label{fig:gridw9-aif-paths-softgs-s4-action-probs}
    \centering
    \includegraphics[width=\textwidth]{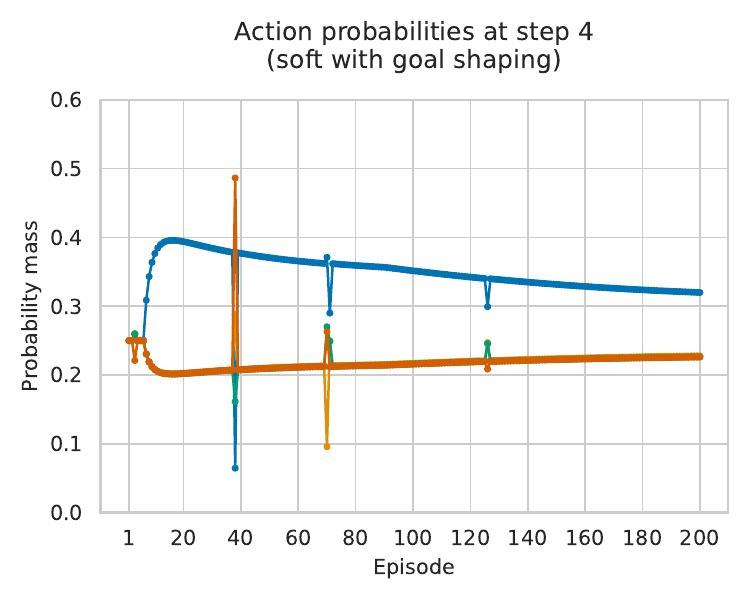}
  \end{subfigure}
  \begin{subfigure}{0.45\textwidth}\label{fig:gridw9-aif-paths-hardgs-s4-action-probs}
    \centering
    \includegraphics[width=\textwidth]{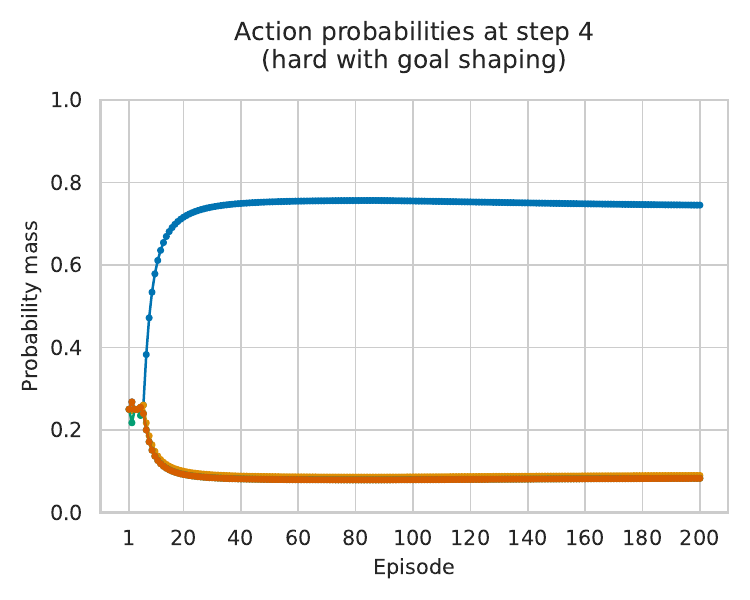}
  \end{subfigure}
  \centering
  \begin{subfigure}{0.45\textwidth}\label{fig:gridw9-aif-paths-soft-s4-action-probs}
    \centering
    \includegraphics[width=\textwidth]{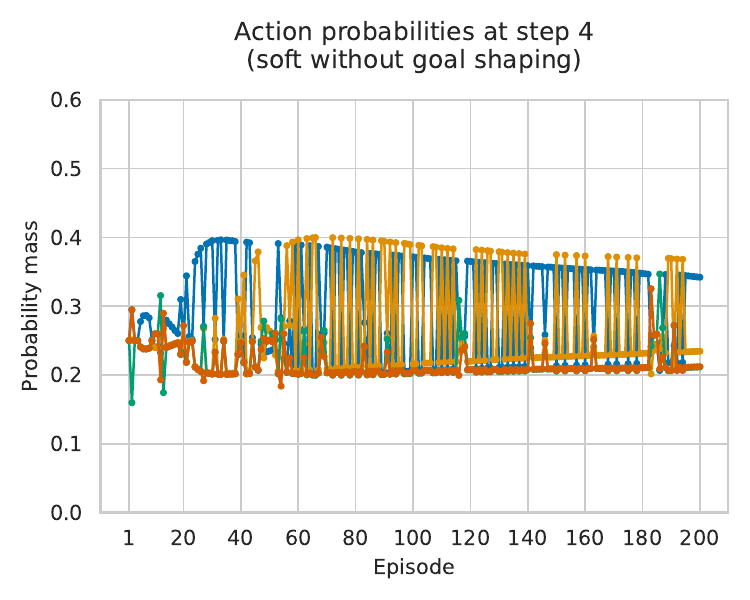}
  \end{subfigure}
  \begin{subfigure}{0.45\textwidth}\label{fig:gridw9-aif-paths-hard-s4-action-probs}
    \centering
    \includegraphics[width=\textwidth]{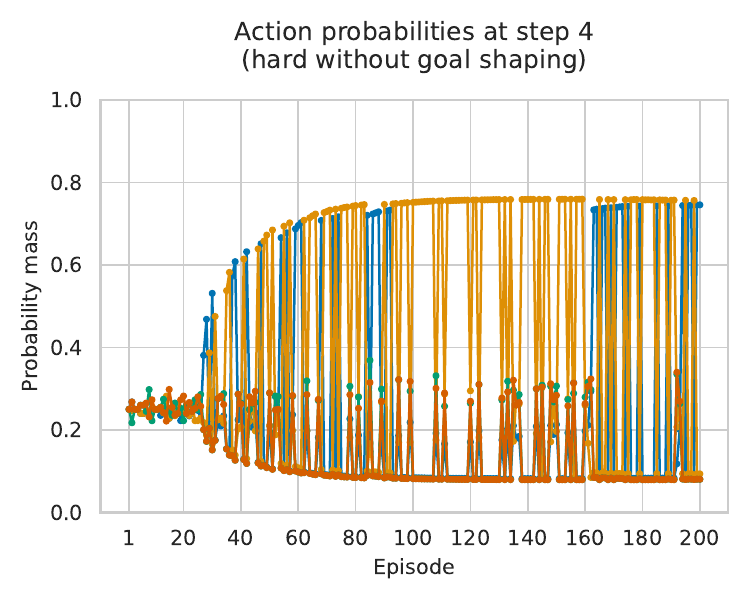}
  \end{subfigure}
  \begin{subfigure}{0.65\textwidth}
    \centering
    \includegraphics[width=\textwidth]{gridw9_actions_legend}
  \end{subfigure}
  \caption{Action probabilities at step 4 of each episode (showing results for one agent).}\label{fig:gridw9-s4-action-probs}
\end{figure}

\subsubsection{Comparison of action-dependent KL divergence from ground truth}\label{sssecx:gridw9-actions-kl}

\subsubsection{Ground truth transition maps}\label{sssecx:gridw9-gtruth-trans-maps}

\begin{figure}[H]
\centering
\begin{subfigure}{0.45\textwidth}
    \centering
    \includegraphics[width=\textwidth]{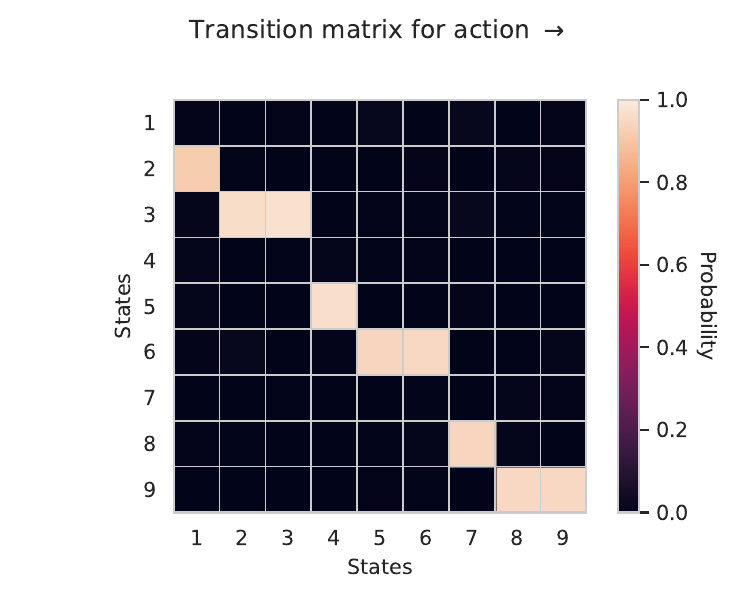}\label{fig:gridw9-gtruth-matrix-B-a0}
\end{subfigure}
\begin{subfigure}{0.45\textwidth}
    \centering
    \includegraphics[width=\textwidth]{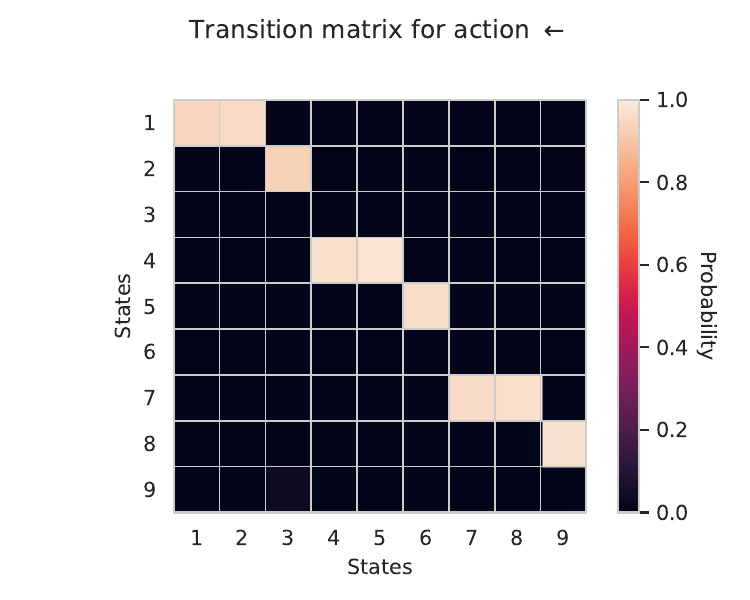}\label{fig:gridw9-gtruth-matrix-B-a2}
\end{subfigure}
\caption{Ground truth transition maps for action \(\rightarrow\) and \(\leftarrow\) in the grid world.}\label{fig:gridw9-gt-arl}
\end{figure}

\begin{figure}[H]
\centering
\begin{subfigure}{0.45\textwidth}
    \centering
    \includegraphics[width=\textwidth]{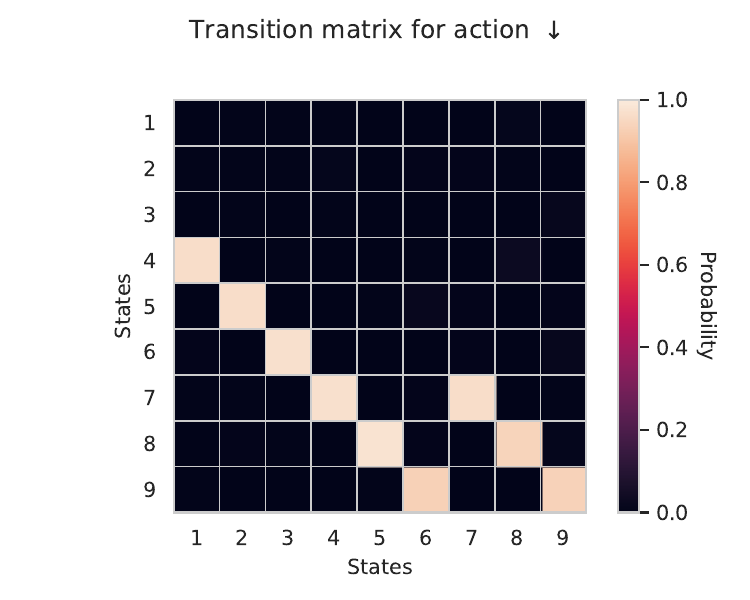}\label{fig:gridw9-gtruth-matrix-B-a1}
\end{subfigure}
\begin{subfigure}{0.45\textwidth}
    \centering
    \includegraphics[width=\textwidth]{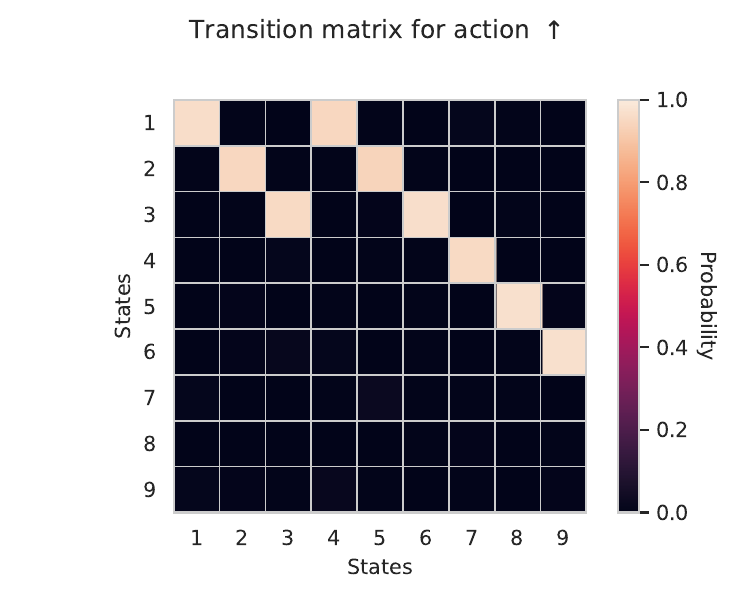}\label{fig:gridw9-gtruth-matrix-B-a3}
\end{subfigure}
\caption{Ground truth ransition maps for action \(\downarrow\) and \(\uparrow\) in the grid world.}\label{fig:gridw9-a13}
\end{figure}

\subsubsection{Learned transition maps}\label{sssecx:gridw9-learned-trans-maps}

\begin{figure}[H]
\centering
\begin{subfigure}{0.45\textwidth}
    \centering
    \includegraphics[width=\textwidth]{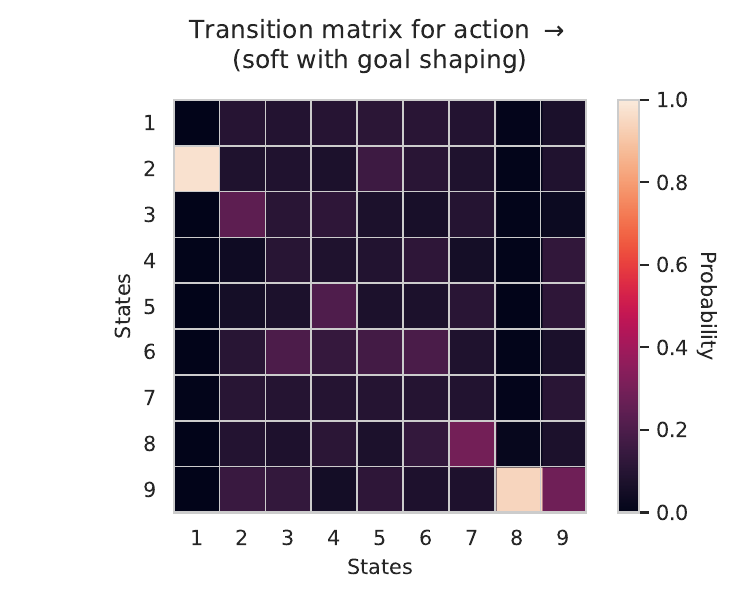}\label{fig:gridw9-aif-au-softgs-matrix-B-a0}
\end{subfigure}
\begin{subfigure}{0.45\textwidth}
    \centering
    \includegraphics[width=\textwidth]{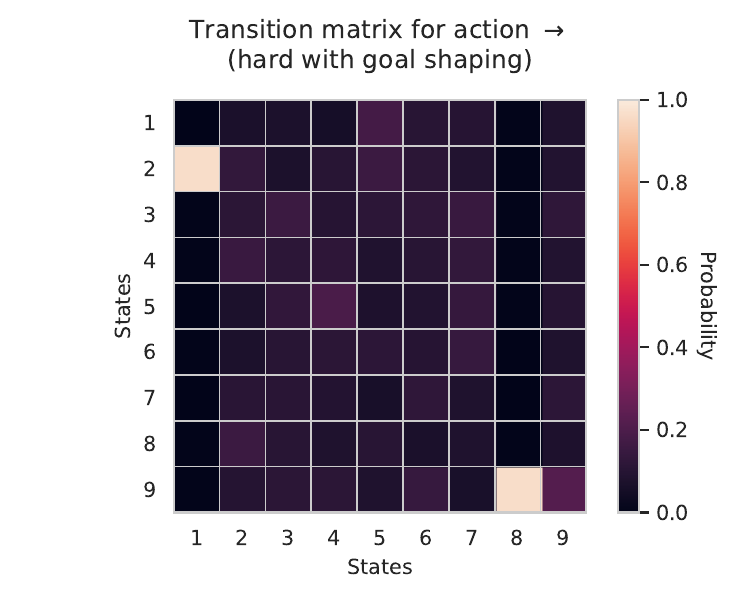}\label{fig:gridw9-aif-au-hardgs-matrix-B-a0}
\end{subfigure}
\begin{subfigure}{0.45\textwidth}
    \centering \includegraphics[width=\textwidth]{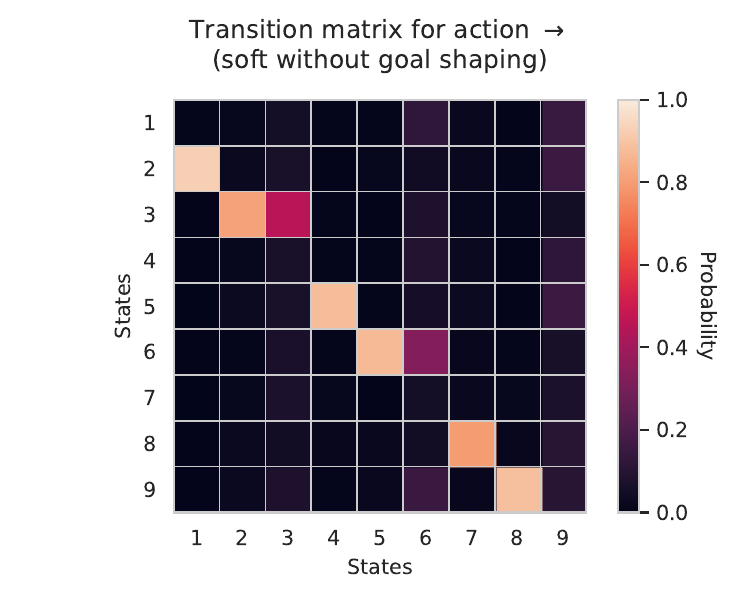}\label{fig:gridw9-aif-au-soft-matrix-B-a0}
\end{subfigure}
\begin{subfigure}{0.45\textwidth}
    \centering \includegraphics[width=\textwidth]{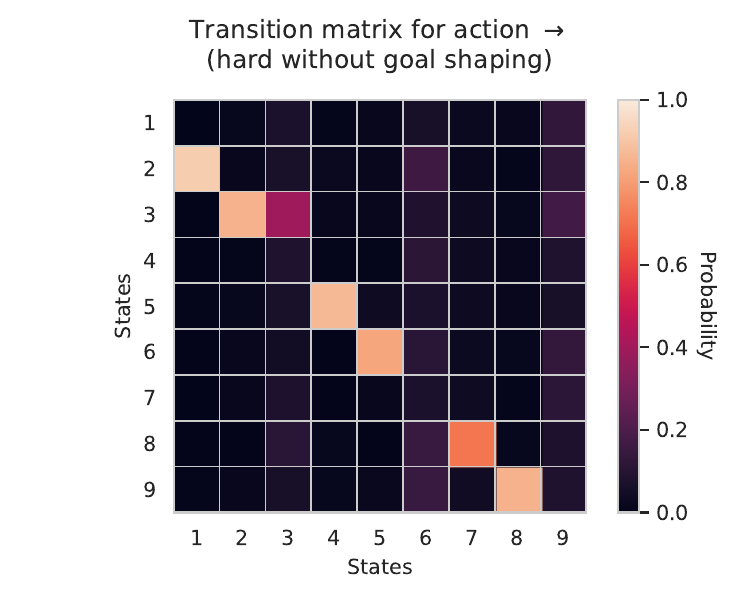}\label{fig:gridw9-aif-au-hard-matrix-B-a0}
\end{subfigure}
\caption{Transition maps for action \(\rightarrow\).}\label{fig:gridw9-a0r}
\end{figure}

\begin{figure}[H]
\centering
\begin{subfigure}{0.45\textwidth}
    \centering \includegraphics[width=\textwidth]{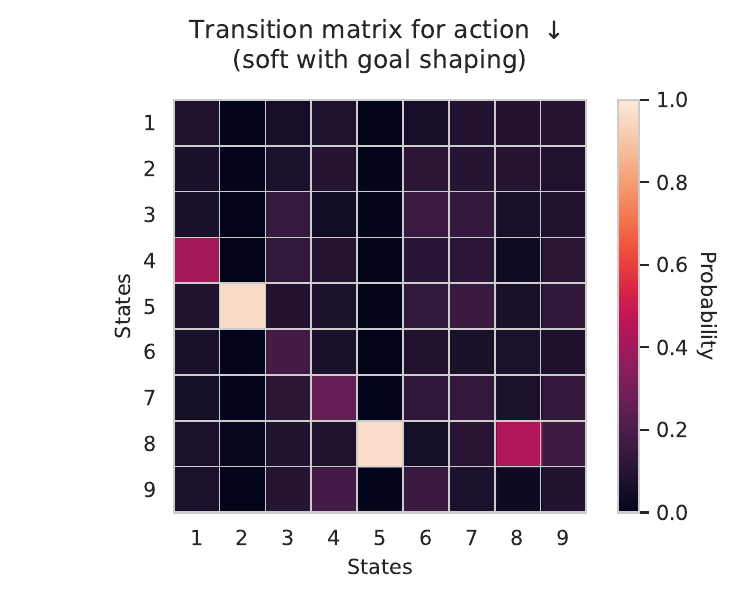}\label{fig:gridw9-aif-au-softgs-matrix-B-a1}
\end{subfigure}
\begin{subfigure}{0.45\textwidth}
    \centering \includegraphics[width=\textwidth]{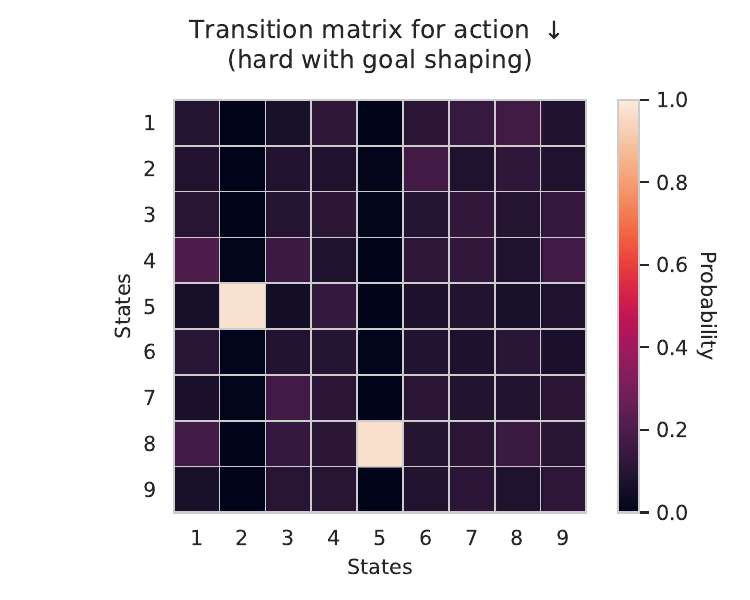}\label{fig:gridw9-aif-au-hardgs-matrix-B-a1}
\end{subfigure}
\begin{subfigure}{0.45\textwidth}
    \centering \includegraphics[width=\textwidth]{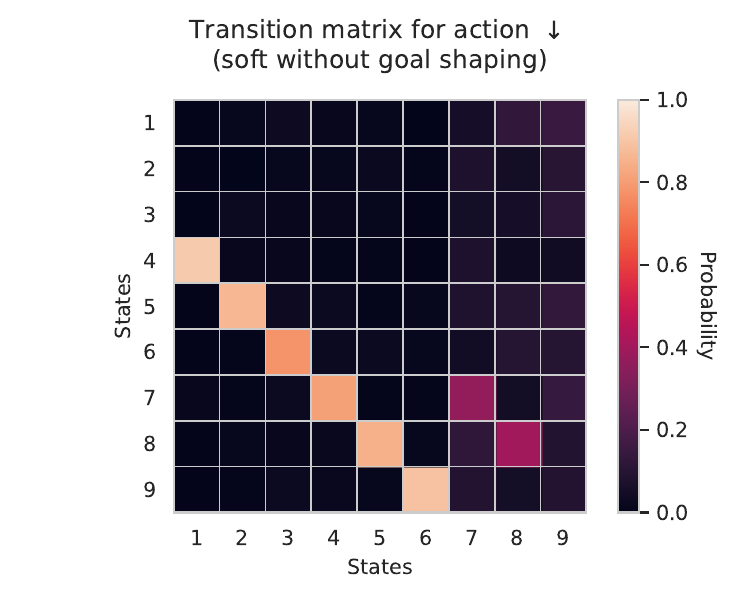}\label{fig:gridw9-aif-au-soft-matrix-B-a1}
\end{subfigure}
\begin{subfigure}{0.45\textwidth}
    \centering \includegraphics[width=\textwidth]{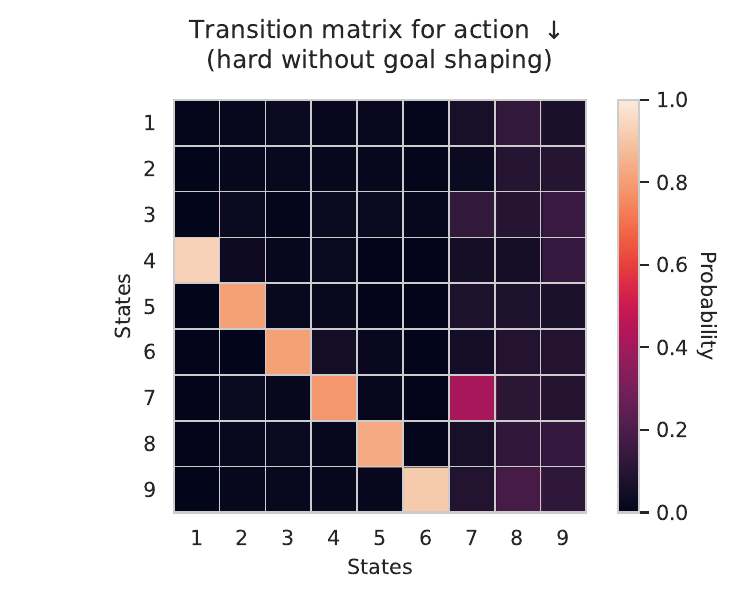}\label{fig:gridw9-aif-au-hard-matrix-B-a1}
\end{subfigure}
\caption{Transition maps for action \(\downarrow\).}\label{fig:gridw9-a1u}
\end{figure}

\begin{figure}[H]
\centering
\begin{subfigure}{0.45\textwidth}
    \centering \includegraphics[width=\textwidth]{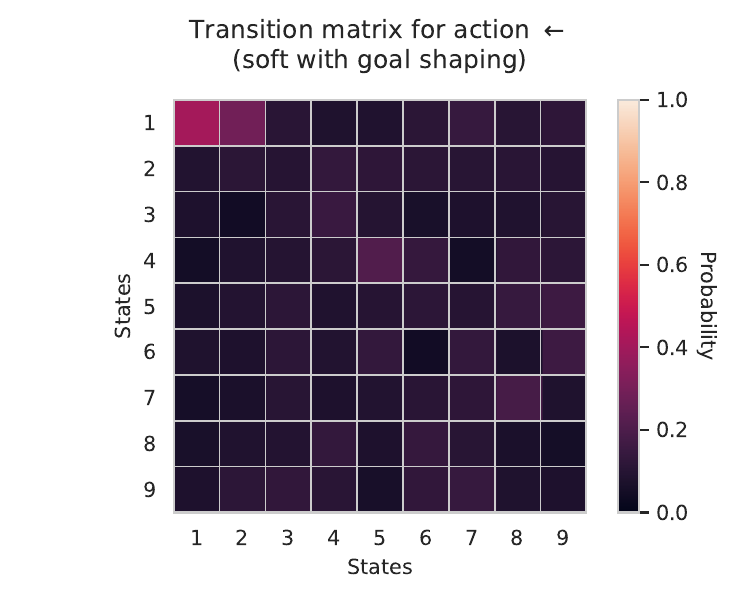}\label{fig:gridw9-aif-au-softgs-matrix-B-a2}
\end{subfigure}
\begin{subfigure}{0.45\textwidth}
    \centering \includegraphics[width=\textwidth]{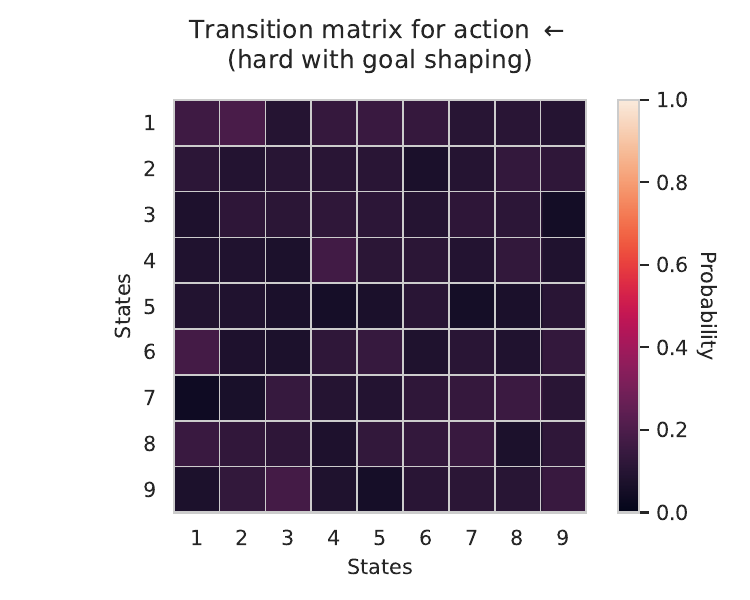}\label{fig:gridw9-aif-au-hardgs-matrix-B-a2}
\end{subfigure}
\begin{subfigure}{0.45\textwidth}
    \centering \includegraphics[width=\textwidth]{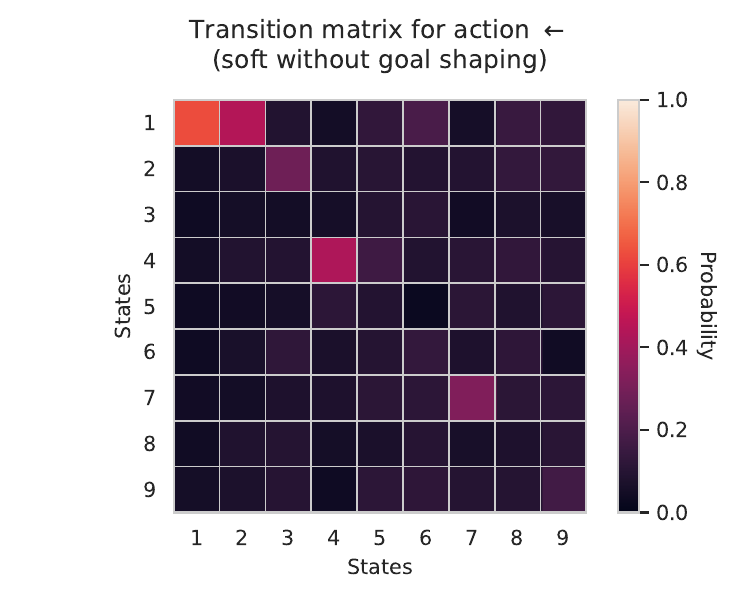}\label{fig:gridw9-aif-au-soft-matrix-B-a2}
\end{subfigure}
\begin{subfigure}{0.45\textwidth}
    \centering \includegraphics[width=\textwidth]{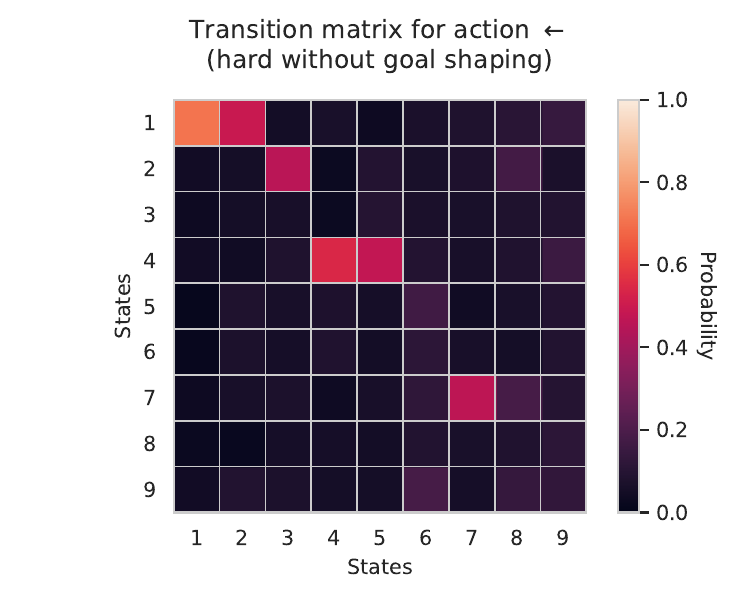}\label{fig:gridw9-aif-au-hard-matrix-B-a2}
\end{subfigure}
\caption{Transition maps for action \(\leftarrow\).}\label{fig:gridw9-a2l}
\end{figure}

\begin{figure}[H]
\centering
\begin{subfigure}{0.45\textwidth}
    \centering \includegraphics[width=\textwidth]{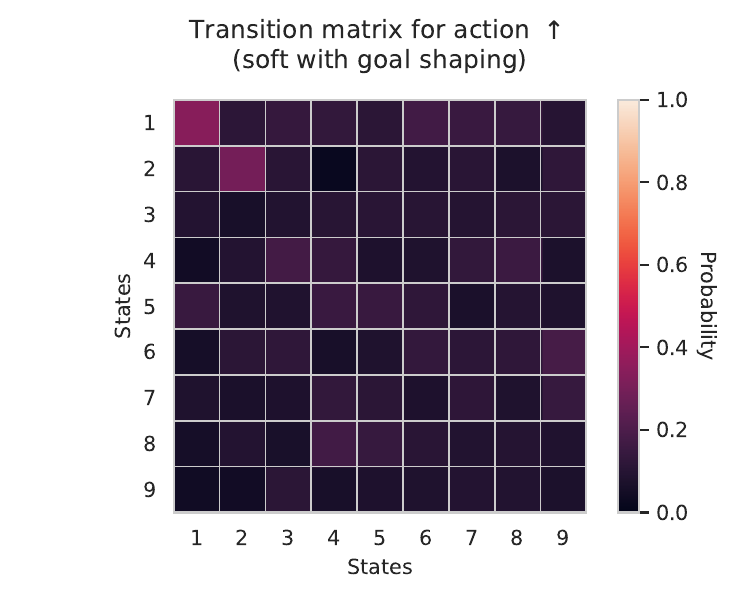}\label{fig:gridw9-aif-au-softgs-matrix-B-a3}
\end{subfigure}
\begin{subfigure}{0.45\textwidth}
    \centering \includegraphics[width=\textwidth]{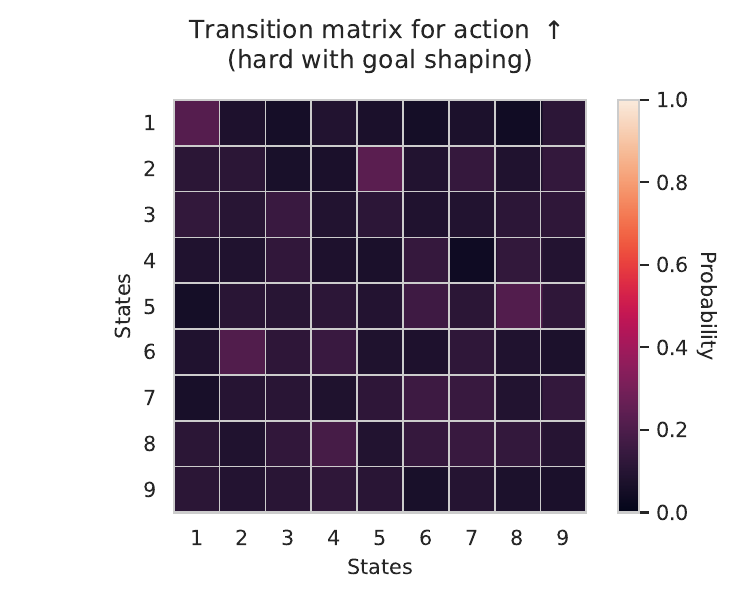}\label{fig:gridw9-aif-au-hardgs-matrix-B-a3}
\end{subfigure}
\begin{subfigure}{0.45\textwidth}
    \centering \includegraphics[width=\textwidth]{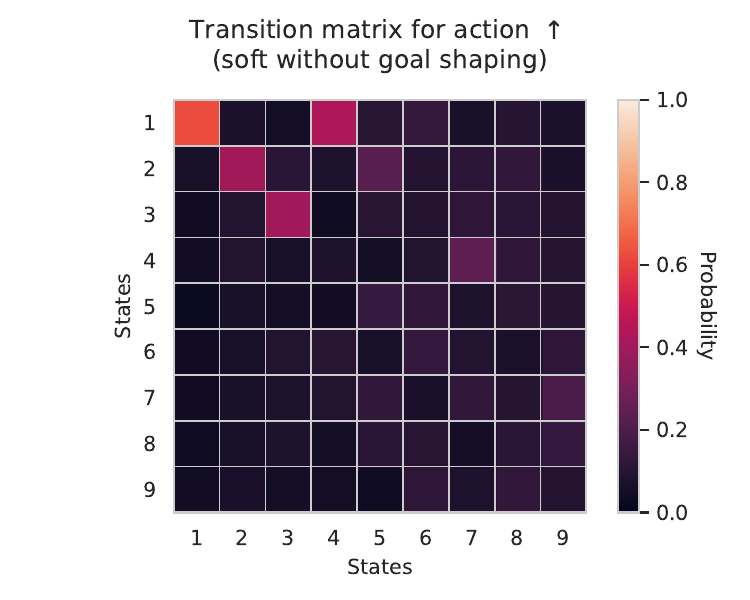}\label{fig:gridw9-aif-au-soft-matrix-B-a3}
\end{subfigure}
\begin{subfigure}{0.45\textwidth}
    \centering \includegraphics[width=\textwidth]{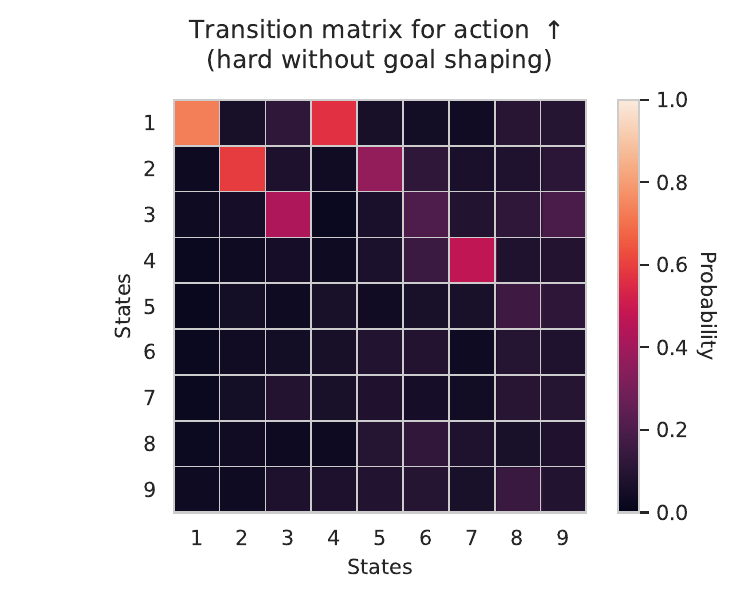}\label{fig:gridw9-aif-au-hard-matrix-B-a3}
\end{subfigure}
\caption{Transition maps for action \(\uparrow\).}\label{fig:gridw9-a3d}
\end{figure}

\end{document}